\documentclass[lettersize,journal]{IEEEtran}
\usepackage{amsmath,amsfonts}
\usepackage{algorithmic}
\usepackage{algorithm}
\usepackage{array}
\usepackage[caption=false,font=normalsize,labelfont=sf,textfont=sf]{subfig}
\usepackage{textcomp}
\usepackage{stfloats}
\usepackage{url}
\usepackage{verbatim}
\usepackage{graphicx}
\usepackage{cite}
\usepackage{booktabs}
\usepackage{graphics}
\begin{document}

\title{GBMST: An Efficient Minimum Spanning Tree Clustering Based on Granular-Ball Computing}
    \author{Jiang~Xie,
	Shuyin~Xia*,
	Guoyin~Wang,
	Xinbo~Gao,
	\thanks{Jiang Xie, Shuyin Xia, Guoyin Wang and Xinbo Gao are with the Chongqing Key Laboratory of Computational Intelligence, Chongqing University of Telecommunications and Posts, 400065, Chongqing, China. E-mail: xiejiang@cqupt.edu.cn, xiasy@cqupt.edu.cn(corresponding author), wanggy@cqupt.edu.cn, gaoxb@cqupt.edu.cn.}}


\IEEEpubid{}

\maketitle

\begin{abstract}
 Most of the existing clustering methods are based on a single granularity of information, such as the distance and density of each data. This most fine-grained based approach is usually inefficient and susceptible to noise. Therefore, we propose a clustering algorithm that combines multi-granularity Granular-Ball and minimum spanning tree (MST). We construct coarse-grained granular-balls, and then use granular-balls and MST to implement the clustering method based on "large-scale priority", which can greatly avoid the influence of outliers and accelerate the construction process of MST. Experimental results on several data sets demonstrate the power of the algorithm. All codes have been released at https://github.com/xjnine/GBMST.
\end{abstract}


\section{Introduction}
\IEEEPARstart{A}{s} an unsupervised learning, the purpose of the clustering algorithm is to divide the data into target clusters, the data of the same cluster are similar, but the data of other clusters are different. As a basic machine learning method, clustering has a wide range of applications and research in the fields of image processing and data mining\cite{Chenping2015Discriminative}. As a classic partitional method, K-means\cite{J.M1967Some} is simple and efficient, but it has good clustering effect only for spherical clusters and is greatly affected by the initial clustering center. Chameleon\cite{JF2009Chameleon} is one of the hierarchical clustering. It can automatically and adaptively merge clusters, which is suitable for arbitrary shape clusters. However, it needs to set more parameters, and the worst time complexity is $O(n^{2})$. DBSCAN\cite{1996AEster} based on density, which is suitable for arbitrary shape clusters and can automatically determine the number of clusters. However, it cannot handle data with large density changes, and the density needs to be determined in advance; Spectral\cite{Ng2001On} clustering can cluster high-dimensional data by introducing the concept of graph cutting.

Recently, graph-based clustering has received increasing attention. Chen\cite{Chen2016Decentralized} proposed a graph-based approach to represent data with k-nearest neighbor relations. Based on the same graph theory idea as spectral clustering, Zahn Zahn\cite{Zahn1971Graph} proposed a clustering method using MST. The core problem of graph theory algorithm is to find a graph suitable for data sets. In general, the method of graph theory selects the MST more because the weight of the connected edges in the MST is the smallest, which is consistent with the basic idea of clustering. In the graph theory algorithm, all data points in the data to be clustered are regarded as a fully connected undirected graph $G = (V, E)$. Then, the distance between any two points (generally taking the Euclidean distance) is defined as the weight of the edge, and the MST is constructed for the undirected graph G using Prim or Kruskal algorithm.

For MST-based clustering, it is completed if inconsistent edges between clusters can be found and removed accurately in MST. Many data to be studied can be represented as graphs. The set of nodes in graphs represents data points, and the edges connecting nodes represent the relationship between data points. 
Generally, the clustering algorithm based on MST includes the following three steps: 1) constructing a MST; 2) removing inconsistent edges to obtain a set of connected clusters between nodes; 3) repeating step 2 until the termination condition is satisfied. 

Since Zahn\cite{Zahn1971Graph} first proposed MST-based clustering algorithm, many scholars have focused on how to determine inconsistent edges. Under the ideal condition that the distance between clusters is large and there is no outlier, the inconsistent edge is the longest edge. Due to the existence of outliers and noise points, the edge information in MST will be seriously disturbed. The longest cut edge is not the correct one. Xu et al\cite{Ying0Clustering}. expressed multidimensional gene expression data using MST. They define three objective functions. Minimizing the weight of the subtree by removing the $K-1$ longest edges is the first objective function. Minimizing the total distance between data points and cluster centers is the second objective function. Obtaining the global optimal solution by combining different data points around the representative data point is the third objective function. Moving inconsistent edges according to the edge length is easily affected by outliers. Laszlo\cite{Laszlo2005Minimum} proposed to solve this problem by limiting the size of the smallest cluster. Based on Euclidean distance, a cascaded MST algorithm was proposed by Grygorash et al\cite{Grygorash2006Minimum}. By inputting the number of clusters $K$, the method uses the sum of the mean and variance of the edge weights of MST as the threshold, deletes edges with weights greater than the threshold, and repeats this process until $K$ clusters are generated. In addition to inconsistent edges, how to define the node density also has an important impact on the clustering results. However, these methods often require manual adjustment of parameters to achieve better results. 

Traditional MST clustering algorithms usually only use edge information to complete clustering. Due to the limitation of information, these algorithms are very susceptible to outliers. Therefore, many MST clustering methods based on local density have been proposed. These methods\cite{Zhong2011Minimum} typically use the number of adjacent nodes and the degree of a point to define the density of nodes. Combining density and MST, Chowdhury\cite{Chowdhury1997Minimal} assumes that the density of index points is the lowest compared to its surrounding neighbors, and then finds the area composed of these index points to complete the clustering. Zhong et al.\cite{Zhong2010A} detected distance-separated and density-separated clusters through two rounds of MST construction. In recent years, many MST clustering methods combined with multivariate Gaussian\cite{Vathy2006Hybrid} and information theory\cite{Andreas2012Information} have also been widely studied.

Most of the existing clustering methods are based on a single granularity of information, such as the distance and density of each data. Therefore, the MST clustering algorithm is greatly affected by outliers, and the process of establishing MST is slow. Inspired by Chen's "large-scale priority"\cite{Linchen1982T} published in the journal Science in 1982, we use granular-ball to represent the data in a coarse-grained way. We propose a clustering algorithm that combines multi-granularity Granular-Ball and  MST (GBMST). GBMST first divides the data set by the granular-ball and establishes the fully connected graph of the granular-ball. With the idea of MST algorithm, each hyper ball is regarded as a sample point, and the MST is generated by Prim algorithm. Then input the parameter K of the generated cluster, and the maximum weight edges are pruned each time to generate the corresponding number of clusters. GBMST not only reduces the influence of outliers on cluster formation, but also accelerates the process of MST establishment by using the hyper ball as a sample point generation.

The main contributions of this paper are as follows:

1) Self-adaption: The generation of coarse-grained granular-balls is based on the data distributed measurement, which is completed through adaptive iteration. Therefore, the construction of minimum spanning tree based on granulation is adaptive.

2) Efficiency: Because the number of granular-balls is far less than the number of data, the efficiency of constructing the minimum spanning tree based on granular-balls is higher, and the efficiency of the proposed clustering algorithm is also improved.

3) Robustness: Since each granular-ball covers many points and only contains two data, namely center and radius, a small number of noise points can be smoothed, so that the granular-ball will not be affected.

We have arranged the rest of this paper as follows. In Section II, we review related work based on three classes of classical clustering algorithms, MST clustering algorithms and granular computing. Section III introduces the proposed clustering algorithm GBMST. In Section IV, we present and analyze the relevant experimental results. In Section V, we give a summary of this paper. 
\begin{figure*}[!h]
	\centering
	\hfil
	\subfloat[]{\includegraphics[width=1.5in]{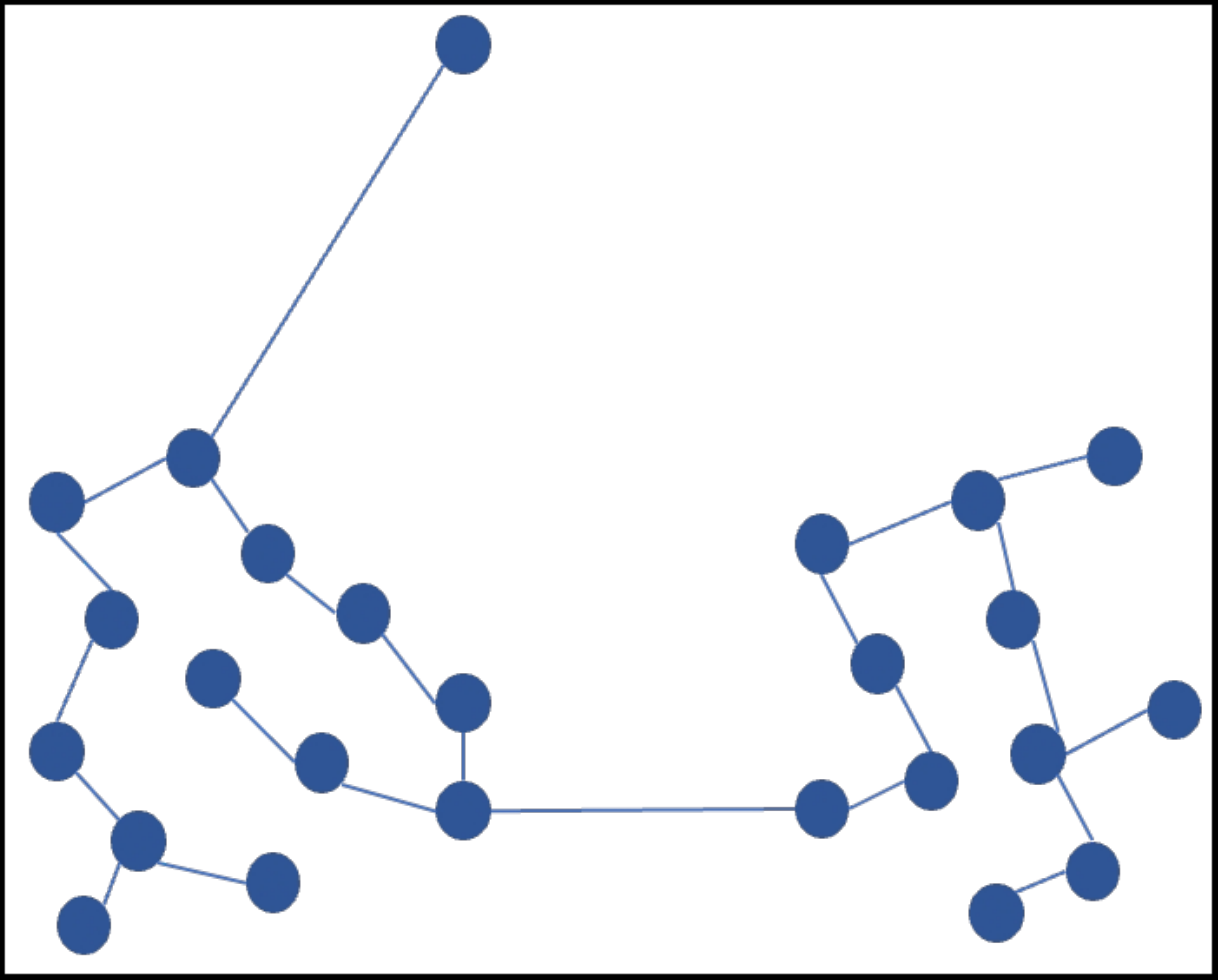}%
		\label{fig_1_1}}
	\hfil
	\subfloat[]{\includegraphics[width=1.5in]{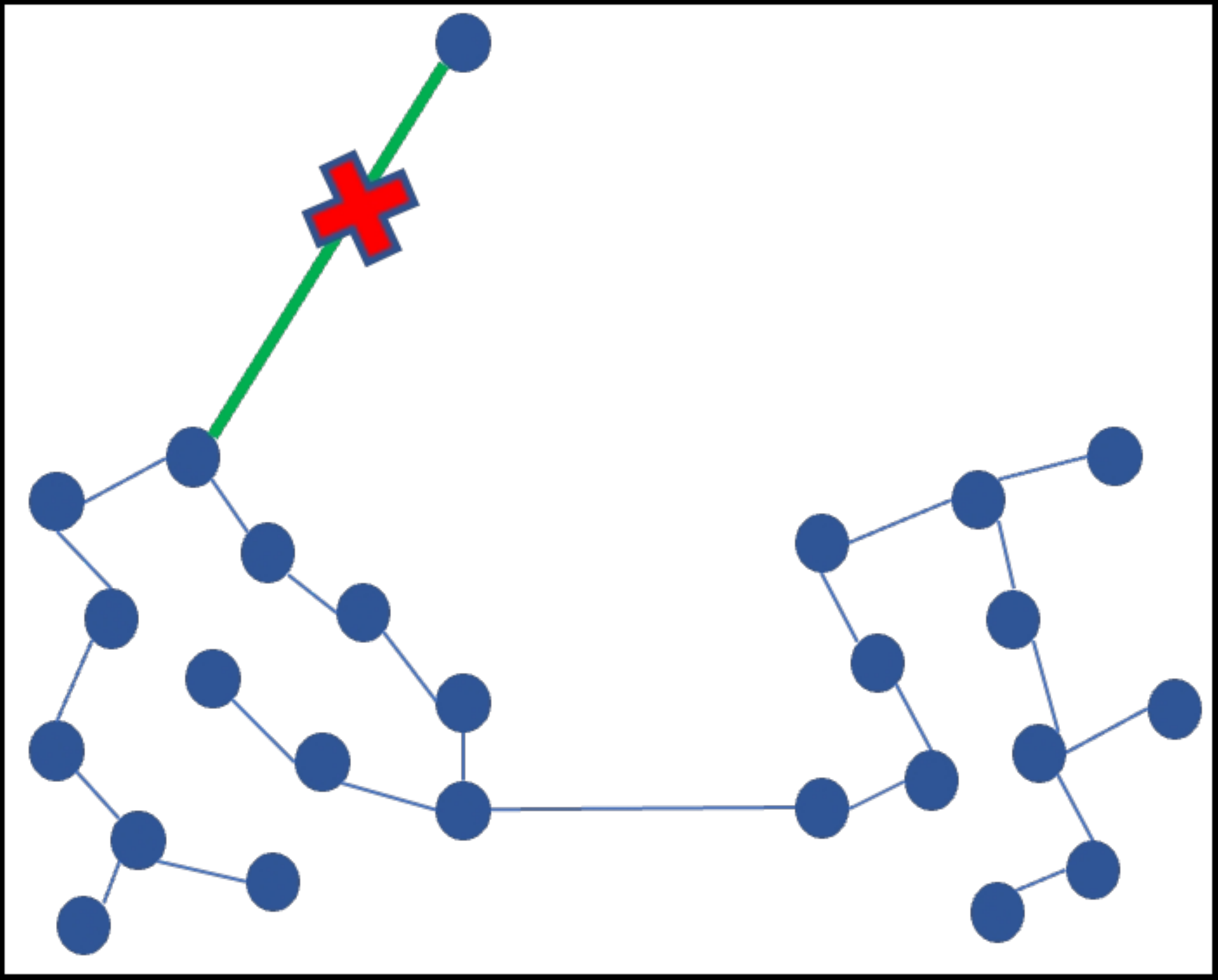}%
		\label{fig_1_2}}
	\hfil
	\subfloat[]{\includegraphics[width=1.5in]{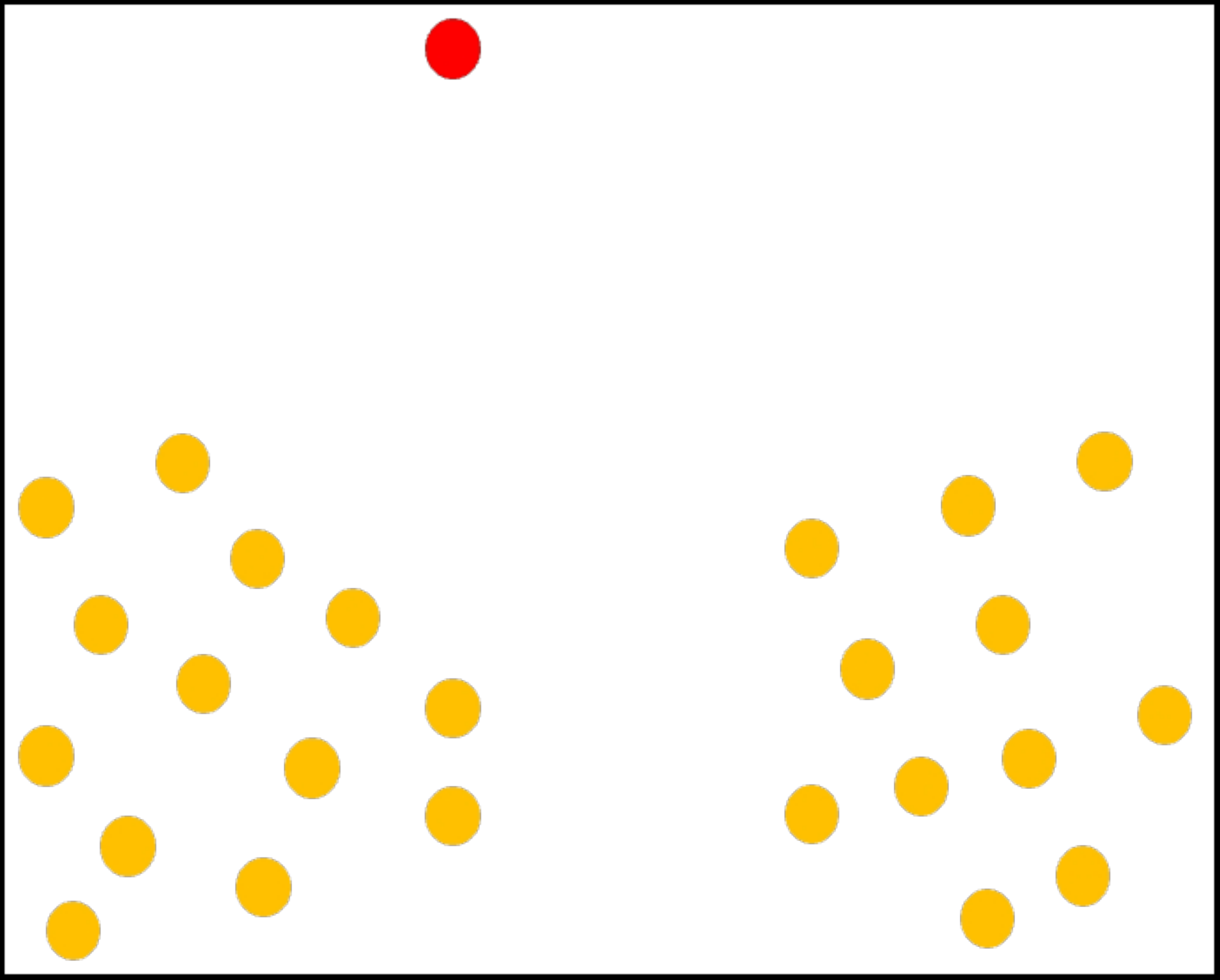}%
		\label{fig_1_3}}
	\hfil
	\subfloat[]{\includegraphics[width=1.5in]{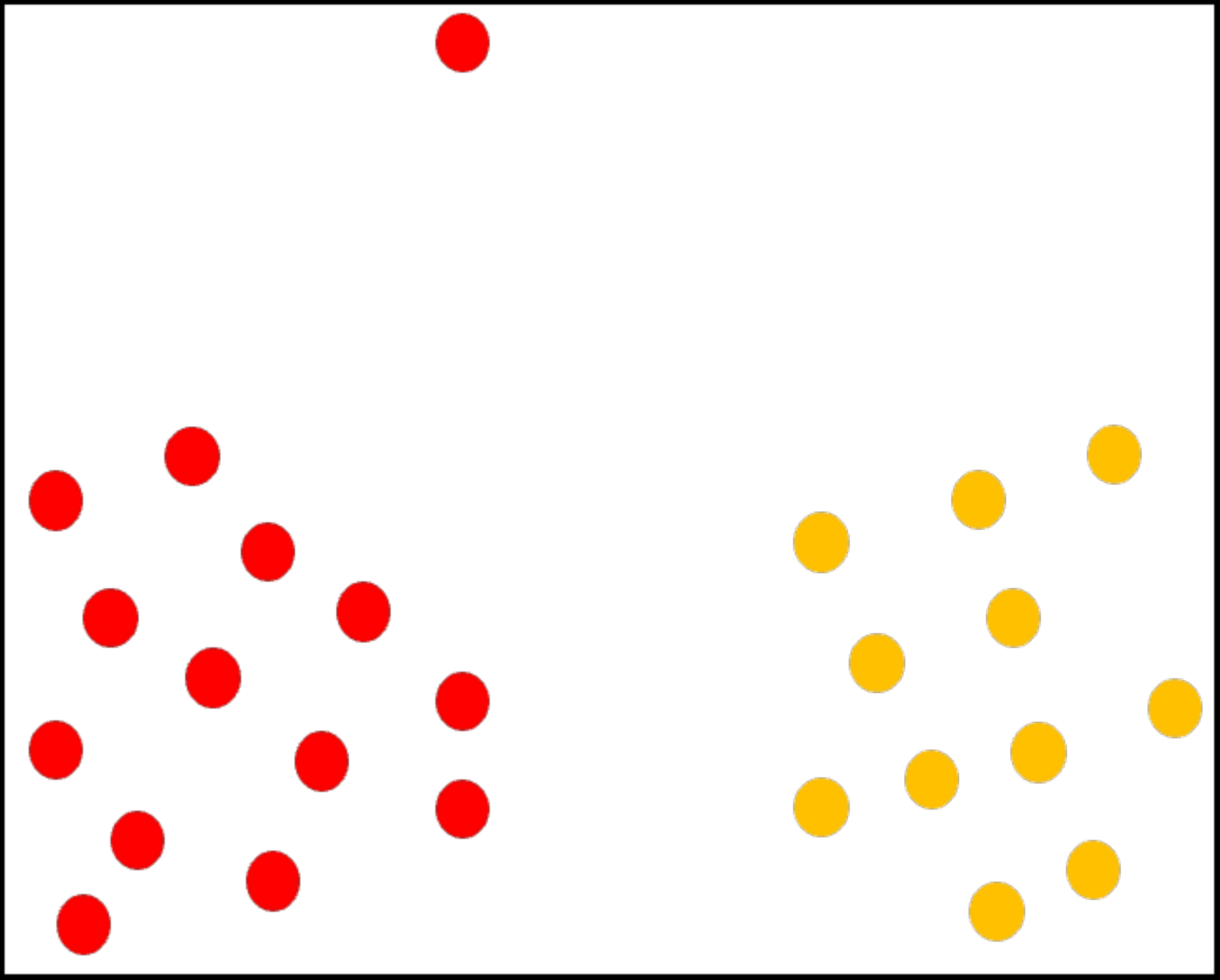}%
		\label{fig_1_4}}
	\caption{The traditional MST-Based clustering algorithm. (a) Minimum spanning tree of the original data. (b) The green edge is cut edge. (c) The clustering results. (d) The correct clustering results.}
	\label{fig_1}
\end{figure*}
\section{RELATED WORK}
\subsection{Classical Clustering Algorithms}
K-means\cite{J.M1967Some} is the most widely used partition algorithm, and its main drawback is that the clustering results are subject to the cluster shape, and better clustering results can be obtained only on spherical clusters. When the boundary of the cluster is irregular, the clustering result is poor. Because the algorithm randomly selects the initial clustering center and uses the mean value of the data points in the class as the representative point while ignoring the density distribution in the class, the clustering result cannot obtain the global optimum, but starts to converge at the local optimum. To overcome these shortcomings, many researchers focus on how to select the initial cluster centers. For example, Gonzalez\cite{Gonzalez1985Clustering} proposed the maximum and minimum distance algorithm. The core idea of the algorithm is to take the data points as far as possible. However, this method still needs to randomly select the first initial clustering center, which will lead to unstable clustering results. Similarly, David Arthur et al\cite{Arthur2007K} proposed the K-means++ algorithm, which has the same problem as the maximum and minimum distance algorithm. In addition to these methods based on the distance between data points, there is a clustering center initialization method CCIA based on data compression principle proposed by Khan et al.\cite{Khan2013Cluster}, which is not suitable for the initial clustering center selection of high dimensional data sets. Stephen J. Redmond et al.\cite{Redmond2007A} proposed a clustering method based on KD-Tree, which estimates the density of data at different locations to select the initial clustering center. This method uses the mean value of all data points in the data box to replace the data points in the data box, which cannot correctly express the distribution of data points in the box; if an attribute of all data points in a data box is the same, the data box volume is 0, resulting in infinite density and meaningless results. Density-based clustering algorithms\cite{1996AEster,Chen2016Decentralized} identify different high-density regions formed by low-density segmentations through the definition of density. Density-based applied spatial clustering with noise (DBSCAN)\cite{1996AEster}, as the most classic density-based clustering technique, can effectively detect clusters of complex shapes. DBSCAN defines density as the number of data points that lie within the scan radius and the radius is fixed, so the density is dynamically changing, making DBSCAN unsuitable for datasets with different densities. Therefore, the existing problems of DBSCAN have also been extensively studied\cite{Chen2016Decentralized,Ankerst1999OPTICS,Rodriguez2014Clustering}.The DP\cite{Rodriguez2014Clustering} algorithm assumes that the density of the cluster center is much higher than that of the surrounding neighbors, and it is farther away from other high-density points. As a new density-based clustering method, DP only needs to use density and distance to get the cluster centers, and then assign other points to the cluster centers to complete the clustering. However, this assumption cannot handle data with complex shapes and large variations in density.

\subsection{MST Clustering Algroithms}
The key challenge of MST clustering algorithms is how to solve the two problems of inconsistent edges and node density. Chowdhury\cite{Chowdhury1997Minimal} assumes that sparse regions must lie between the boundaries of any two clusters, and through these sparse regions, inconsistent edges can be located. Based on neighborhood density difference estimation in MST-based, Luo et al.\cite{Ting2010A} proposed a clustering algorithm. Wang et al.\cite{Wang2013Enhancing} eliminated outliers by obtaining the local density factor of each data point, and then constructed the MST method for clustering.

Real datasets may have different densities, imbalanced clusters, and arbitrary shapes\cite{XiP2015A}, and it is difficult to obtain a method that can accurately identify inconsistent edges with a single MST structure. Therefore, many methods based on MST and other fields are cross-combined to improve the clustering results\cite{He2011CHSMST}. SAM\cite{Zhong2011Minimum,Ma2021A} combines hierarchical clustering and obtains the initial points through the information of MST, and then uses K-mean to obtain the final clustering result. KP\cite{Huang2019A,Mishra2019A} is a hybrid clustering algorithm. KP uses natural neighborhoods to adaptively obtain the K value and natural density of each object. In this search process, MST is built from natural cores. This method is suitable for complex manifold patterns and data sets with large variation in density. 

In order to reduce the interference of noise and outlier, noise and outlier can be excluded before constructing the minimal generation\cite{Wang2013Enhancing}. Its working principle is to calculate the density factor and threshold in the process of building the MST, and use the data below the threshold as noise, thereby reducing the interference to the cluster, and after the MST is obtained, the longest edge is continuously removed until the target cluster is obtained. Recently, a clustering method LDP\_MST\cite{Cheng2019Clustering} was proposed to construct a MST by replacing all data with local density cores. Based on the density core, LDP\_MST has strong robustness to noise.
\subsection{Granular-Ball Computing}
Given a data set $D = {p_i(i=1, 2, ..., n)}$, where $n$ is the number of samples on $D$. Granular balls $GB_1, GB_2, \dots , GB_m$ are used to cover and represent the data set $D$. Suppose the number of samples in the $j^{th}$ granular-ball $GB_j$ is expressed as $|GB_j|$, then its coverage degree can be expressed as $ {\textstyle \sum_{j=1}^{m}}\left ( \left | GB_j \right |  \right ) /n $. The basic model of granular-ball coverage can be expressed as
\begin{equation}\label{eqGB}
	\setlength{\abovedisplayskip}{6pt}
	\setlength{\belowdisplayskip}{3pt}
	\begin{split}
		min \ \ \lambda _1 \ast n /{\sum_{j=1}^{m}}\left ( \left | GB_j \right |  \right ) /n + \lambda _2 \ast m,  \\
		s.t. \ \ quality(GB_j) \ge T, \ \ \ \ \ \ \ \ \ \ \ \ \ \ 
	\end{split}
\end{equation}
where $\lambda _1$ and $\lambda _2$ are the corresponding weight coefficients, and $m$ is the number of granular balls. When other factors remain unchanged, the higher the coverage, the less the sample information is lost, and the more the number of granular-balls, the the characterization is more accurate. Therefore, the minimum number of granular-balls should be considered to obtain the maximum coverage degree when generating granular-balls. By adjusting the parameters $\lambda _1$ and $\lambda _2$, the optimal granular-ball generation results can be obtained to minimize the value of the whole equation. In most cases, the two items in the objective function do not affect each other and do not need trade off, so $\lambda _1$ and $\lambda _2$ are set to 1 by default. Granular-ball computing can fit arbitrarily distributed data \cite{xia2019granular,xia2022gbsvm}.
\section{GBMST}
Like most existing machine learning methods, existing MST-based clustering algorithms use the finest granularity, treating each data sample as a node and using only edge information to divide clusters. As a result, the construction of MST is easily disturbed by noise, boundary points and outlier, and the construction efficiency is also low. As shown in Fig.1, Due to the influence of outliers, the information of inconsistent edges that need to be cut cannot be obtained correctly, the traditional MST-Based clustering algorithm cuts off the wrong edges, resulting in incorrect clustering results. The correct clustering result is shown in Fig 1d. In view of this, our paper presents a granular-ball clustering clustering algorithm based on MST. (GBMST). Taking the dataset shown in Fig.2 as an example, GBMST are mainly divided into three stages. In stage 1, we generate granular-ball based on the similarity of the data distribution of the original dataset\cite{Gonzalez1985Clustering}. In stage 2, after filtering out some outliers and outlier granular-balls, we treat each ball as a sample point, and generate a MST according to the prim algorithm. In stage 3, the MST is pruned according to the parameter $k$, and then cut the longest $k-1$ edges, to generate $k$ connected branches, each connected branch represents a cluster, and then the noises and outlier granular-balls are grouped into clusters of granular-ball closest to them.
\begin{figure*}[!h]
	\centering
	\subfloat[]{\includegraphics[width=1.75in]{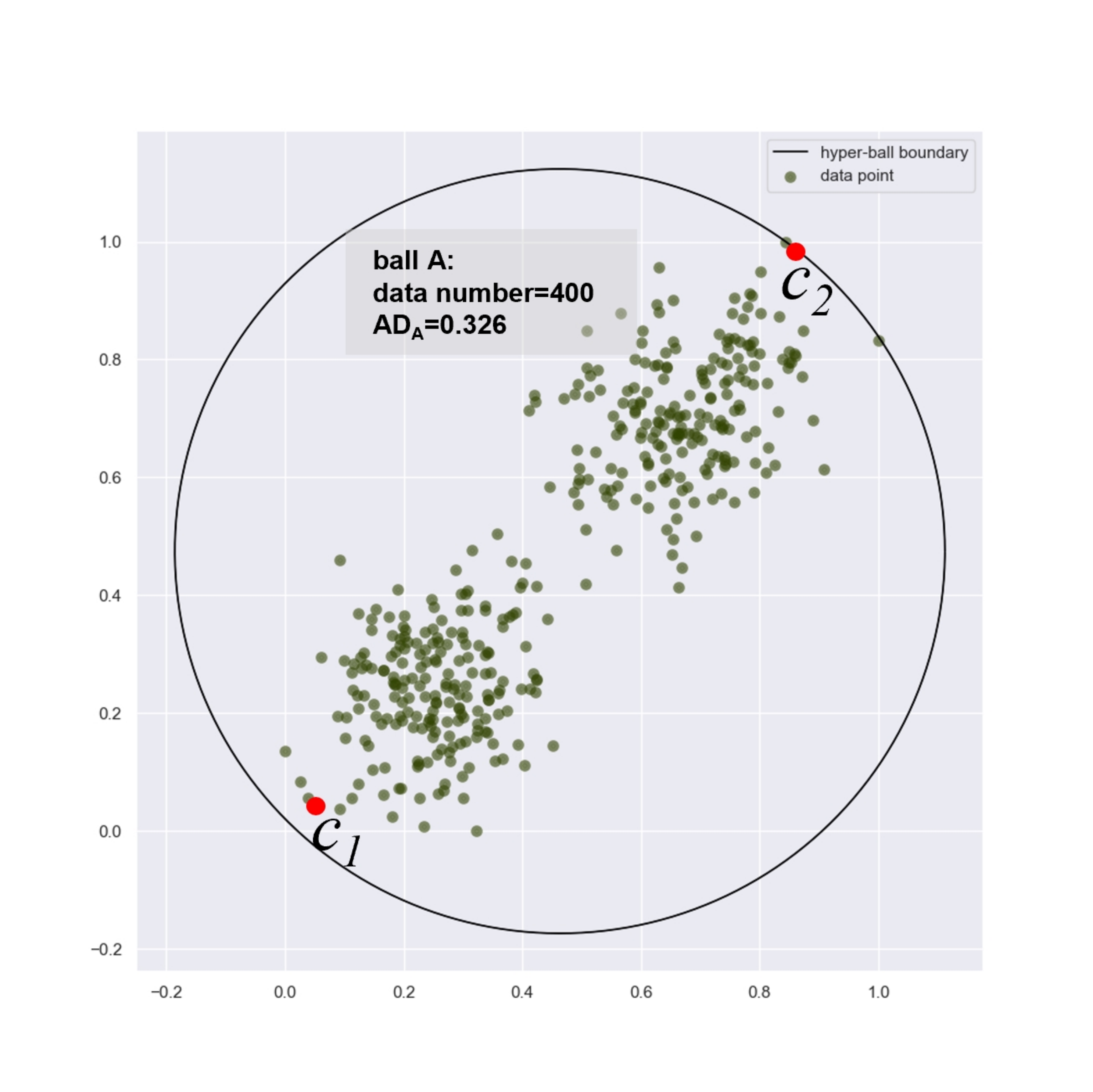}%
		\label{fig_2_1}}
	\subfloat[]{\includegraphics[width=1.75in]{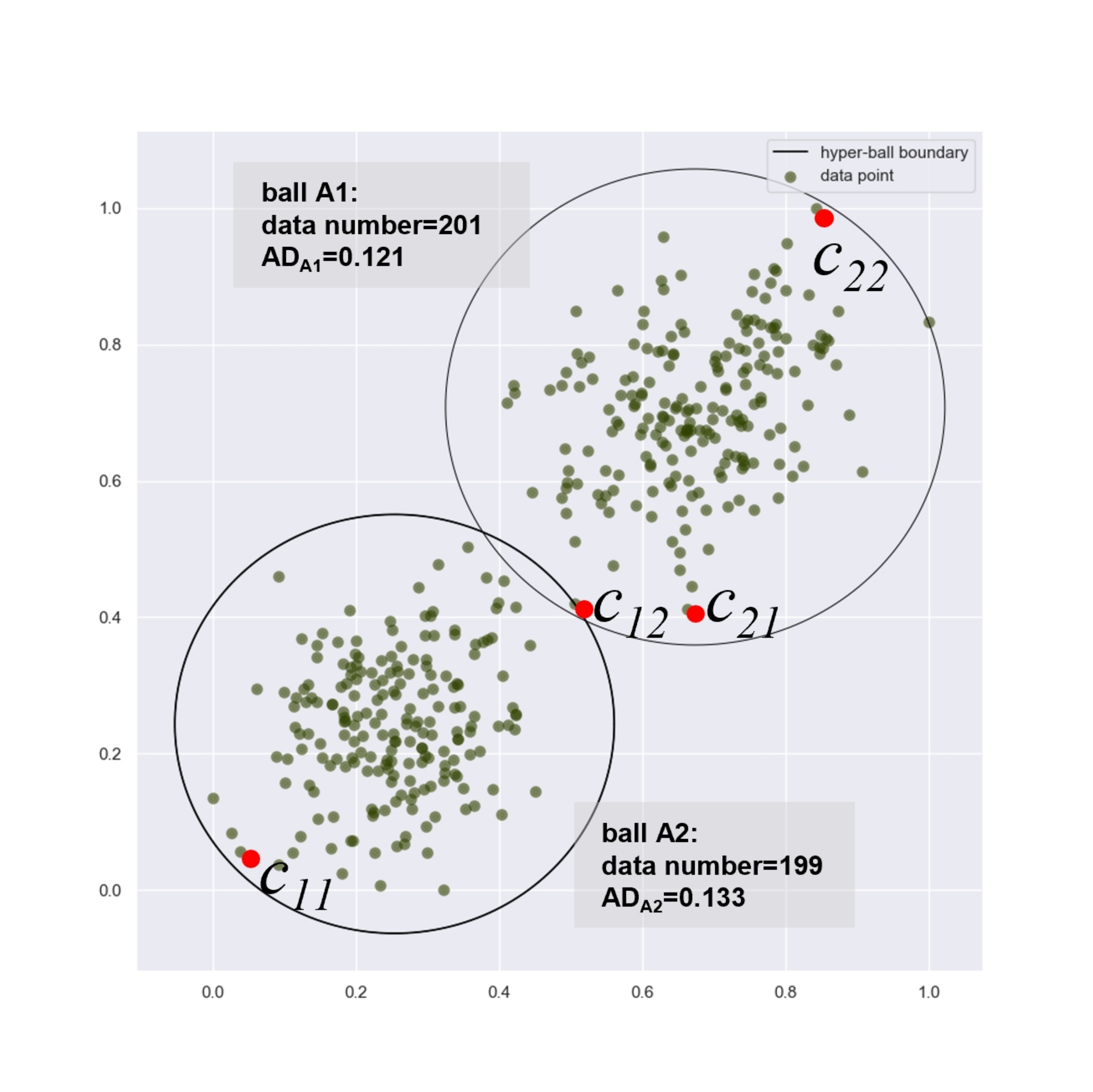}%
		\label{fig_2_2}}
	\subfloat[]{\includegraphics[width=1.75in]{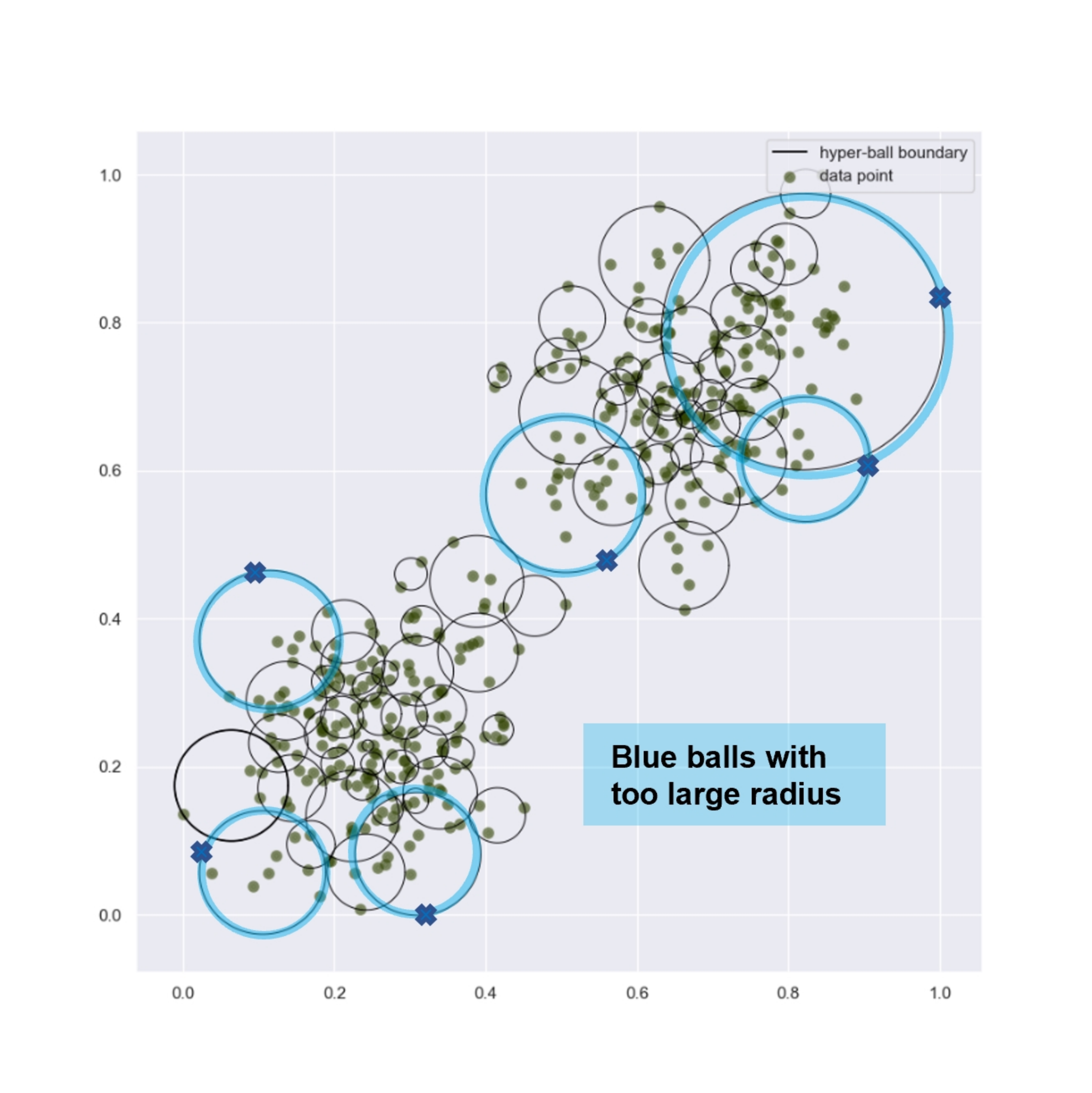}%
		\label{fig_2_3}}
	\subfloat[]{\includegraphics[width=1.75in]{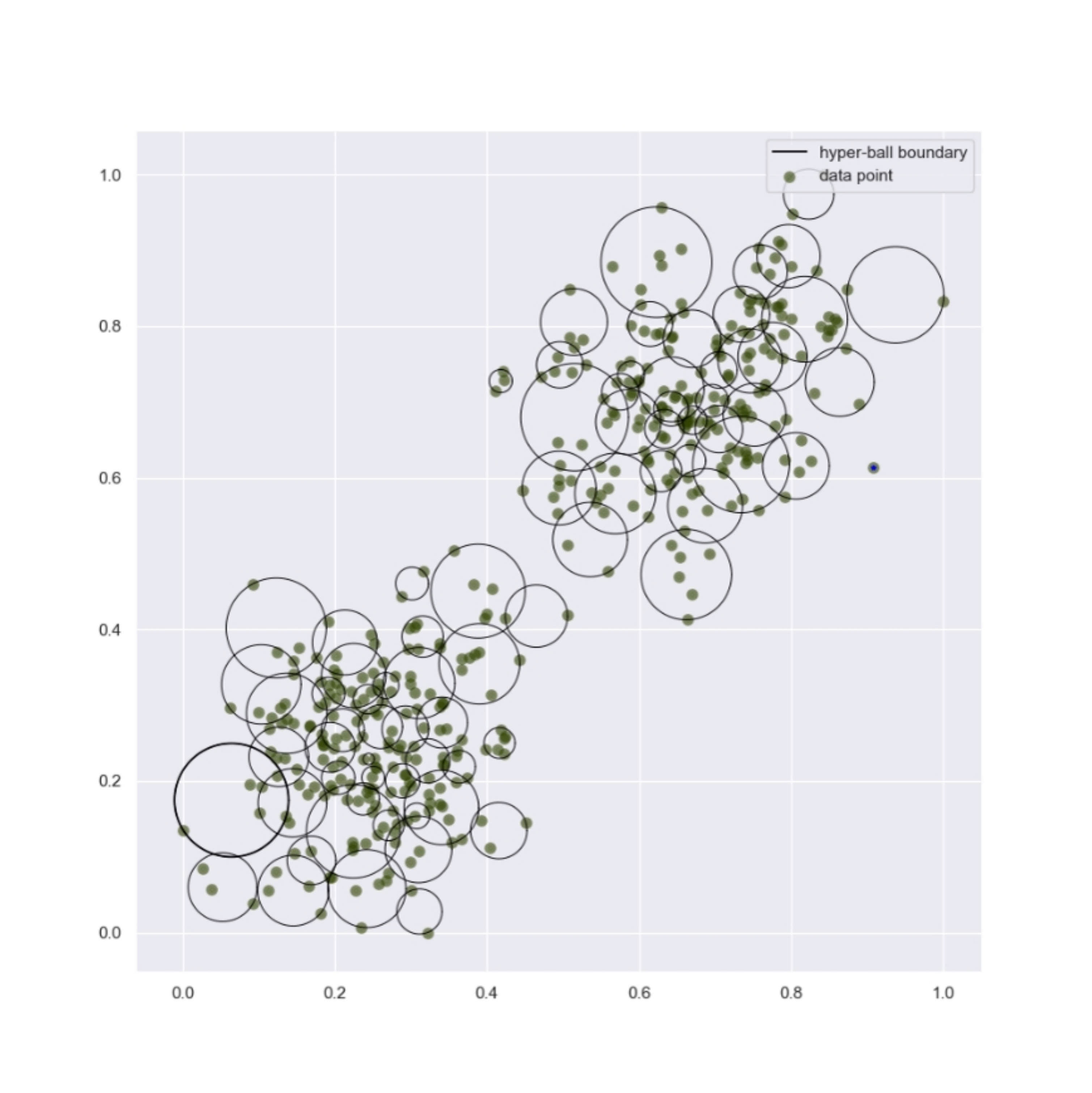}%
		\label{fig_2_4}}
	\caption{The process of  generateing granular-balls on a simplified example. (a) Raw data points in hyperball A. (b) Child hyperball of A. (c) Blue balls with radii that are too large. (d) The final hyperball sets after removing hyperball with a radius that is too large.}
	\label{fig_2}
\end{figure*}

\subsection{The Process of Generating Granular-Balls}
In this section, we mainly improve the granular-ball generation method proposed in\cite{xia2022adaptive} to better deal with the noise.    

\textbf{Definition 1.}Given a dataset $D\in R^d$, there are two quantities: Granular-Ball ($GB_j$) and Distribution Measure ($DM_j$). For each $GB_j$, the $c_{j}$ is the center of $GB_j$ and $r_{j}$ is the radius of $GB_j$. The definitions of $c_j$ and $r_j$ are as follows:
\begin{equation}
	\label{deqn_ex1a}
	c_j =\frac{1}{n_j}\sum_{i=1}^{n_j}p_i.
\end{equation}
\begin{equation}
	\label{deqn_ex2a}
	r_j =max(||p_i-c_j||).
\end{equation}
Where $\left||.\right||$ denotes the 2-norm and $n_j$ represents the number of data located in $GB_j$. The optimization model of granular-ball needs to distinguish between supervised and unsupervised cases. For unsupervised cases, we define the $DM_j$ value based on Formula (1) to guide the adaptive iterative generation process of granular-ball.

\textbf{Definition 2 (DM).} $DM_j$ is measured by computing the ration of the number data point $n_{j}$ and the sum radius $s_{j}$ in $GB_j$, Where $s_{j}=\sum_{i=1}^{n_j}||p_i-c_j||$,
\begin{equation}
	\label{deqn_ex1a}
	DM_j = \frac{s_j}{n_j}.
\end{equation}
The center is the data point that best represents the granular-ball, so a smaller value of $DM_{j}$ means that the data points in the granular-ball are closer to the center, and these data points more similar to each other. In other words,  $DM_{j}$ is the average distance from all data points in the granular-ball to the center, which is used to measure the similarity of the data in the granular-ball. The value of  ranges from 0 to $r_{j}$ , and the smaller the value, the higher the similarity.

The granular-ball was divided according to $DM$. As shown in Fig.2: Firstly, we treat the whole dataset as a granular-ball $A$; Then, we choose the two farthest points $C_1$ and $C_2$ to split the ball $A$ into two sub-balls $A_1$ and $A_2$; third, We calculate the $DM_A$, $DM_{A_1}$ and $DM_{A_2}$ values for $A$, $A_1$ and $A_2$; At last, We judge whether to split the ball by comparing $DM_A$ and $DM_{A_1}$ with $DM_{A_2}$. 
In\cite{xia2022adaptive}, both $DM_{A_1}$  and $DM_{A_2}$ need to be greater than $DM_{A}$ for $GB_A$ to split. However, when there is a lot of noise, this splitting rule will cause a large number of granular-balls to fail to split. 

\textbf{Definition 3(Weighted DM value).} Therefore, we use a weighted DM value for comparison in this paper, which can better adapt to noisy situations. $DM_{weight}$ is defined as follows:
\begin{equation}
	\label{deqn_ex5a}
	DM_{weight} = \dfrac{n_{A1}}{n_{A}}DM_{A1}+\dfrac{n_{A2}}{n_{A}}DM_{A2}.
\end{equation}

\begin{figure*}[!h]
	\centering
	\subfloat[]{\includegraphics[width=1.4in]{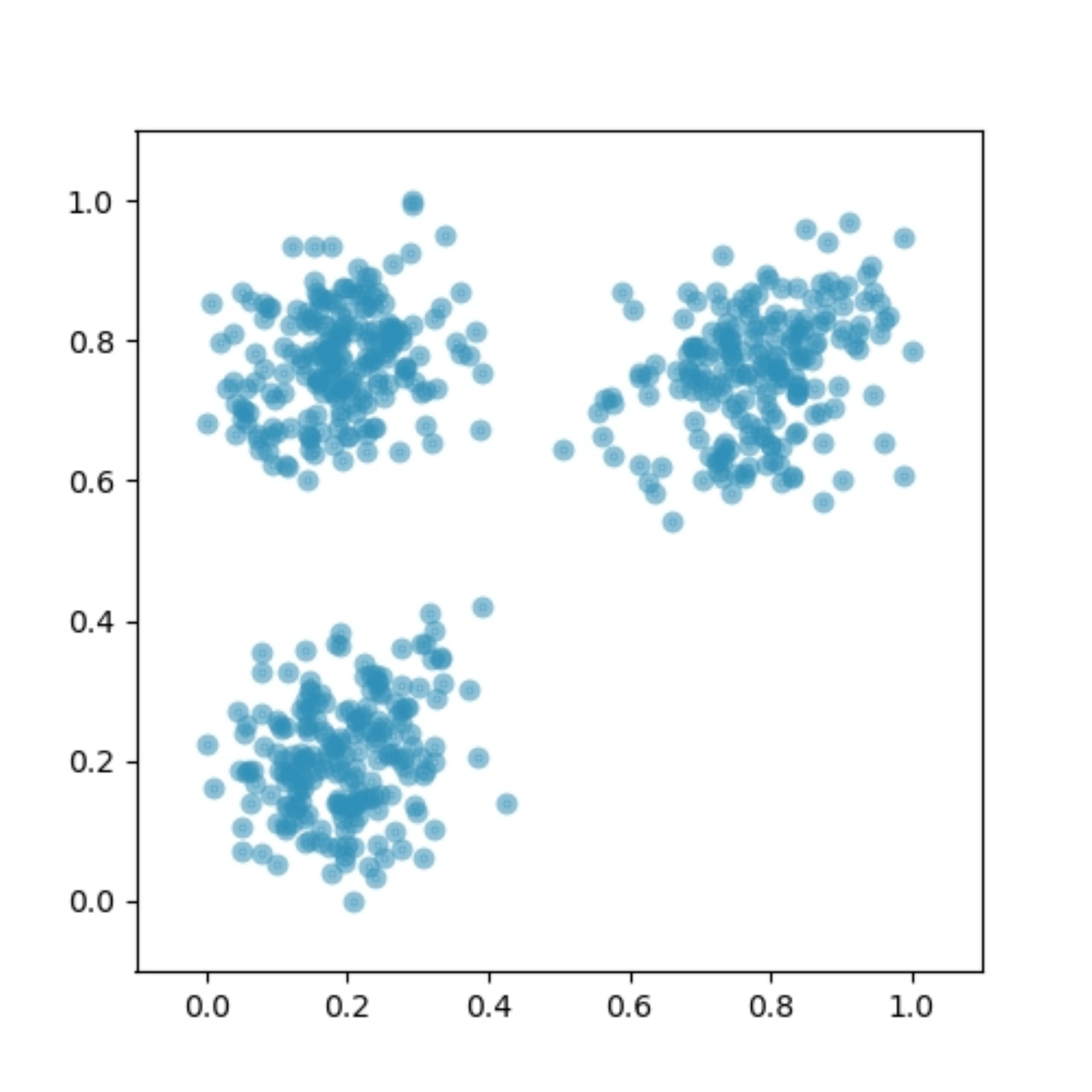}%
		\label{fig_3_1}}
	\subfloat[]{\includegraphics[width=1.4in]{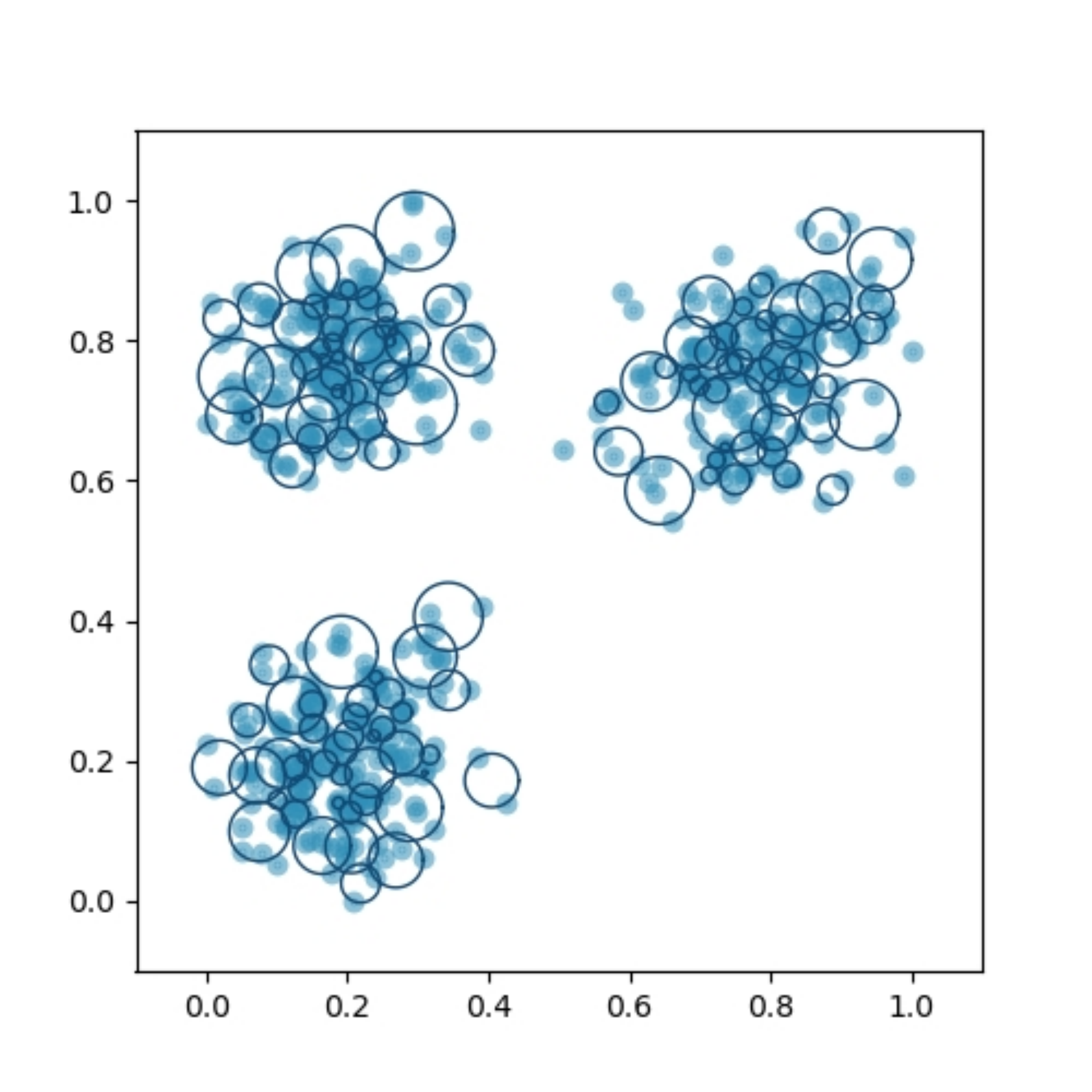}%
		\label{fig_3_2}}
	\subfloat[]{\includegraphics[width=1.4in]{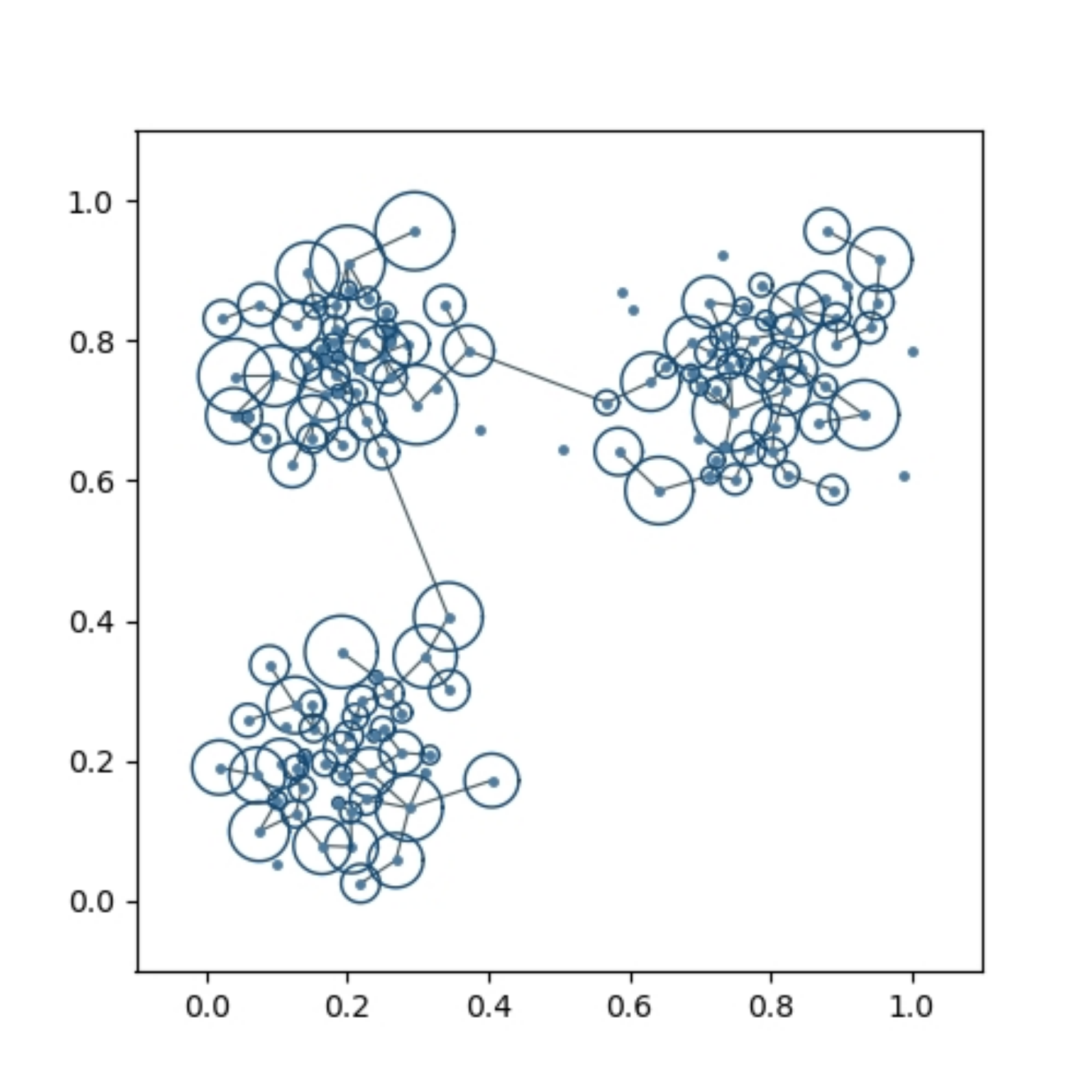}%
		\label{fig_3_3}}
	\subfloat[]{\includegraphics[width=1.4in]{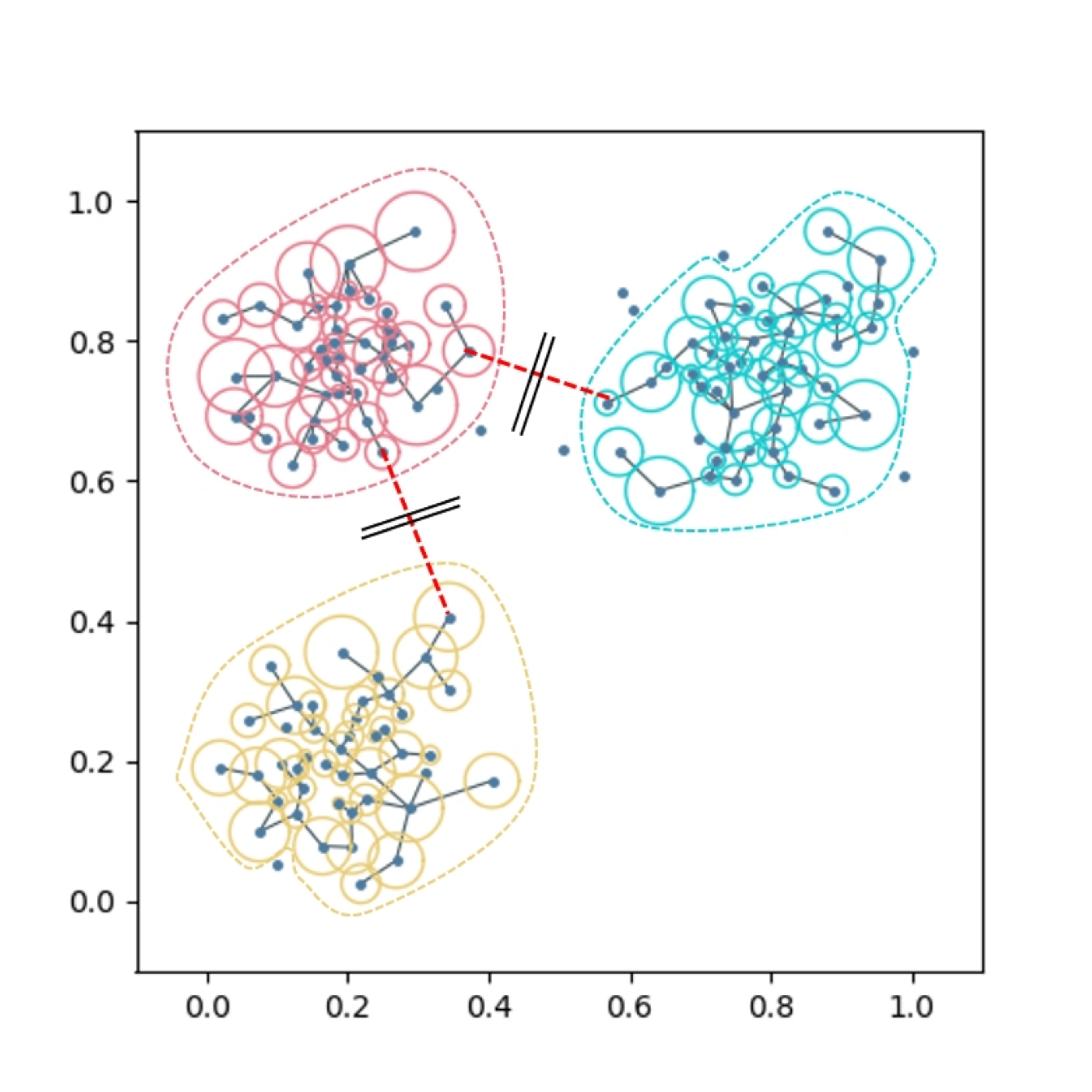}%
		\label{fig_3_4}}
	\subfloat[]{\includegraphics[width=1.4in]{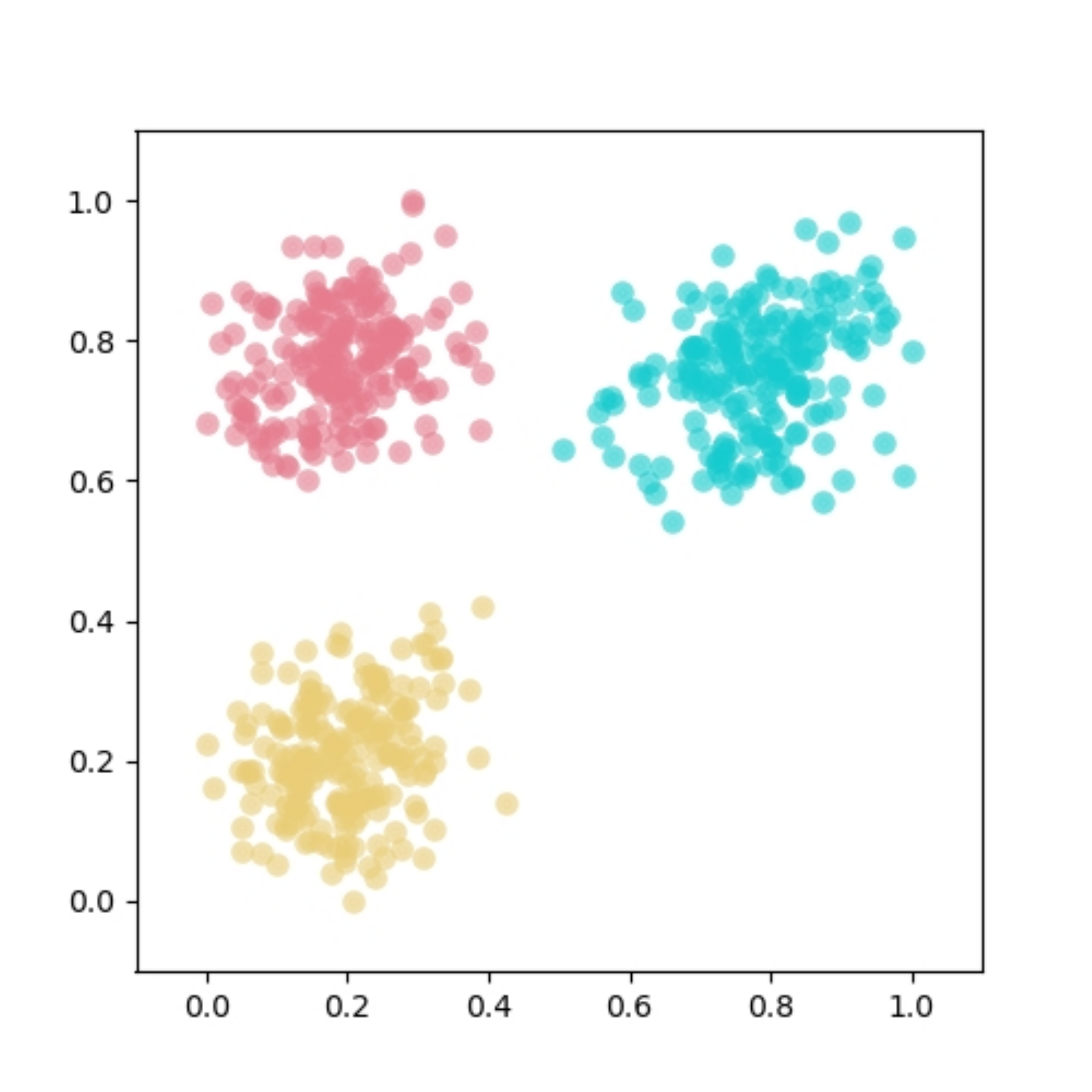}%
		\label{fig_3_5}}
	\caption{The overview of GBMST. (a) Original data set. (b) The granular-balls set. (c) The MST based on granular-balls. (d) The red edge is cut edge. (e) The clustering result of GBMST}
	\label{fig_3}
\end{figure*}

where $n_{A}$, $n_{A_1}$ and $n_{A_2}$ represent the number of data in the corresponding granular-ball, respectively. If $DM_{weight}$ is greater than $DM_A$, the granular-ball $A$ splits. In  Fig. 2c, the final splitting result is illustrated. However, in Fig. 2c, some granular-balls with radii that are too large may still be affected by some boundary or noise points and need to be split; if $r_j > 2\times\max(mean(r),median(r))$ , $GB_j$ needs to be split. $mean(r)$ and $median(r)$ represent the mean and median of all hyperball radii, respectively. After granular-balls with radii that are too large are removed, the splitting process is completed, and the result is shown in Fig. 2d. Based on the above descriptions, the generation of granular-balls algorithm is shown in Algorithm 1.

\begin{algorithm}
	\caption{Generation of Granular-Balls.}\label{alg:alg1}
	\begin{algorithmic}
		\STATE 
		\STATE \textbf{Input}\hspace{0.5cm}D(data set) 
		\STATE \textbf{Output}\hspace{0.3cm}GB sets
		\STATE 1\hspace{1.2cm}\textbf{For} each granular-ball $GB_j$ in D \textbf{do}
		\STATE 2\hspace{1.6cm}calculate $DM_A$, $DM_{weight}$, 
		\STATE 3\hspace{1.8cm} according to Eq.1, Eq.2, Eq.3,Eq.4;
		\STATE 4\hspace{1.6cm}\textbf{If} $DM_{weight}\geq DM_A$ \textbf{Then}
		\STATE 5\hspace{1.8cm}Split $GB_k$;
		\STATE 6\hspace{1.6cm}\textbf{End If}
		\STATE 7\hspace{1.3cm}\textbf{If} the number of GBs is not changing \textbf{Then}  
		\STATE 8\hspace{1.4cm} \space \space \space break;
		\STATE 9\hspace{1.3cm}\textbf{End For}
       	\STATE 10\hspace{1.2cm}\textbf{For} each granular-ball $GB_j$ in D \textbf{do}
        \STATE 11\hspace{1.6cm}calculate $mean(r)$, $median(r)$, 
        \STATE 12\hspace{1.6cm}\textbf{If} $r_{j}\geq 2\times\max(mean(r),median(r))$ \textbf{Then}
        \STATE 13\hspace{1.8cm}\space \space Split $GB_j$;
        \STATE 14\hspace{1.6cm}\textbf{End If}
        \STATE 15\hspace{1.3cm}\textbf{If} the number of GBs is not changing \textbf{Then}  
        \STATE 16\hspace{1.4cm} break;
        \STATE 17\hspace{1.3cm}\textbf{End For}
		\STATE 18\hspace{1.1cm}\textbf{return} GB sets;
	\end{algorithmic}
	\label{alg1}
\end{algorithm}
\subsection{Constructing a MST based on Granular-Balls}
In this section, we construct the MST (MST)  with  granular-balls. In order to solve the defects of the traditional MST clustering method, we propose a method of using the granular-ball as a sample point to construct a MST, and this method can also effectively solve the problem of outliers participating. As shown in Fig.3a-Fig.3c, We substitute the data sample with the center of the granular-ball. The distance of two granular-ball is defined as the distance between the centers of the two granular-balls minus the radii of the two granular-balls. If the two balls overlap, the distance is set 0. The granular-balls participating in the construction of the MST do not include all granular-balls belonging to the outlier set. The distance $dis(GB_{j1},GB_{j2})$ and  $outlier$ set are defined as follows:
\begin{equation}
	\label{deqn_ex7a}
	dis(GB_{j1},GB_{j2}) = dis(c_{j1},c_{j2})-(r_{j1}+r_{j2}).
\end{equation}
\begin{equation}
	\label{deqn_ex7a}
	outlier = \{GB_j| n_j\leq2\}.
\end{equation}
\subsection{Clustering Based on Granular-Ball and MST (GBMST)}
In this section, we prune the MST generated in the previous section to generate clusters. First, according to the parameter k, the k edges with the largest distance are removed to generate k+1 connected branches, each branch representing a cluster. Then, we assign outlier ball to its nearest neighbor granular-ball with cluster label. As shown in Fig.3d, the two longest edges are cut off to get the clustering result. The red dotted lines in the graph are the edges we cut, and the final result of the cluster is shown in Fig.3e. Based on the above descriptions,  the granular-ball based MTS clustering algorithm is designed and shown in Algorithm 2.
\begin{algorithm}
	\caption{GBMST algorithm.}\label{alg:alg2}
	\begin{algorithmic}
		\STATE 
		\STATE \textbf{Input}\hspace{0.5cm}GB sets, $outlier=\emptyset$,k
		\STATE \textbf{Output}\hspace{0.2cm}The Clutering result
		\STATE 1\hspace{1.2cm}\textbf{For} each granular-ball $GB_j$ in GB set \textbf{do}
		\STATE 2\hspace{1.6cm}\textbf{If} $n_j \leq 2 $ \textbf{Then}
		\STATE 3\hspace{1.8cm} $outlier = outlier \cup GB_j$ ;
		\STATE 4\hspace{1.2cm}\textbf{End For}
		\STATE 5\hspace{1.2cm}$V=\{GB_j\|GB_j\notin outlier\}$
		\STATE 6\hspace{1.2cm}$construct MST(V,E)$ 
		\STATE 7\hspace{1.2cm}Cut the longest k-1 edge
		\STATE 8\hspace{1.2cm}Assign cluster labels to k connected components
		\STATE 9\hspace{1.2cm}\textbf{For} each granular-ball $GB_j$ in outlier set \textbf{do}
		\STATE 10\hspace{1.8cm}Assign cluster labels according to 
		\STATE 11\hspace{1.8cm}its nearest granular-ball's cluster lable;
		\STATE 12\hspace{1.1cm}\textbf{return} clustering result;
	\end{algorithmic}
	\label{alg1}
\end{algorithm}

As shown in the above algorithm, GBMST mainly consists of three stages: generating granular-balls, generating a MST with granular-balls, and clustering Based on Granular-Ball and MST. Suppose the number of hyper balls is $N$, $n$ is the number of points in a data set and $m$ is the number of edges. The time complexity of first step  is $O(nlogn)$. And the time complexity of constructing the MST of granular-ball is $O(mlogN)$. Then we cut the longest edge each time and repeat $k-1$ times to generate $k$ clusters. The time complexity of this step is $O(N)$. Finally, if the number of outliers is $u$, the time complexity of process of  outliers ball into the corresponding clusters is O(u*N). Because $N<<n$, $u<<n$ and $m<<n$, the complexity of GBMST is close to $O(nlogn)$.

\section{Experimental evaluations}
To validate the GBMST algorithm, we conduct experimental analysis on synthetic data, biological cell segmentation data, and UCI data set. To verify the effectiveness,We choose to compare with three classical clustering methods K-means\cite{J.M1967Some}, DBSCAN\cite{1996AEster} and DP\cite{Rodriguez2014Clustering}, and compare with two methods based on MST, Normal\_MST\cite{Zahn1971Graph} and LDP\_MST\cite{Cheng2019Clustering}. 

We used the python scikit learning toolbox to conduct experiments on a normal performance PC. The detailed configuration is as follows: Intel(R) Core(TM) i9-10920X CPU @ 3.50GHz, 64GB memory, Windows 10 operating system. 
\subsection{Experiment on Synthetic Data Sets without Noise}
We first conduct experiments on 6 complex synthetic datasets without noise. Table 1 shows the detailed information of the 6 data sets, and the visualization results are shown in Figure 4.
\begin{table}[!h]
	\caption{Synthetic data sets without noise.\label{table1}}
	\renewcommand\arraystretch{1.4}
	\centering
	\begin{tabular}{lccc}
		\hline & Instances  & Clusters & Source \\
		\hline D1   & 899  & 2 & \cite{Aristides2007Clustering}   \\
		       D2   & 944  & 2 & \cite{Aristides2007Clustering}   \\
		       D3   & 630  & 4 & \cite{Aristides2007Clustering}      \\
		       D4   & 788  & 7 & \cite{Aristides2007Clustering}   \\
		       D5   & 1016 & 4 & \cite{Aristides2007Clustering}     \\ 
		      D6   & 1741 & 6 & \cite{Aristides2007Clustering}      \\ 
		\hline
	\end{tabular}
\end{table}

\begin{table}[!h]
	\caption{Parameters settings on Synthetic Data Sets without Noise.\label{table2}}
	\renewcommand\arraystretch{1.4}
	\centering
	\begin{tabular}{lccc}
		\hline             & K-means      & DBSCAN                 & DP \\
		\hline D1   & K=2         & Minpts=3, Eps=0.06     & K=2, dc=2\%   \\
		D2          & K=2         & Minpts=4, Eps=0.06     & K=2, dc=2\%   \\
		D3          & K=4         & Minpts=4, Eps=0.05     & K=2, dc=2\%   \\
		D4          & K=7         & Minpts=5, Eps=0.04     & K=2, dc=2\%   \\
		D5          & K=4         & Minpts=4, Eps=0.05     & K=2, dc=2\%   \\
		D6          & K=6         & Minpts=4, Eps=0.04     & K=2, dc=2\%   \\		       		       		       
		\hline
		\hline             & Normal\_MST  & LDP\_MST              & GBMST  \\
		\hline D1   & K=2         & K=2, Minsize=0.018   & K=2   \\
		D2          & K=2         & K=2, Minsize=0.018   & K=2   \\
		D3          & K=4         & K=4, Minsize=0.018   & K=4   \\
		D4          & K=7         & K=7, Minsize=0.018   & K=7   \\
		D5          & K=4         & K=4, Minsize=0.018   & K=4   \\
		D6          & K=6         & K=6, Minsize=0.018   & K=6   \\		       
		\hline
	\end{tabular}
\end{table}

\begin{figure*}[!h]
	\centering
	\subfloat[]{\includegraphics[width=1.65in]{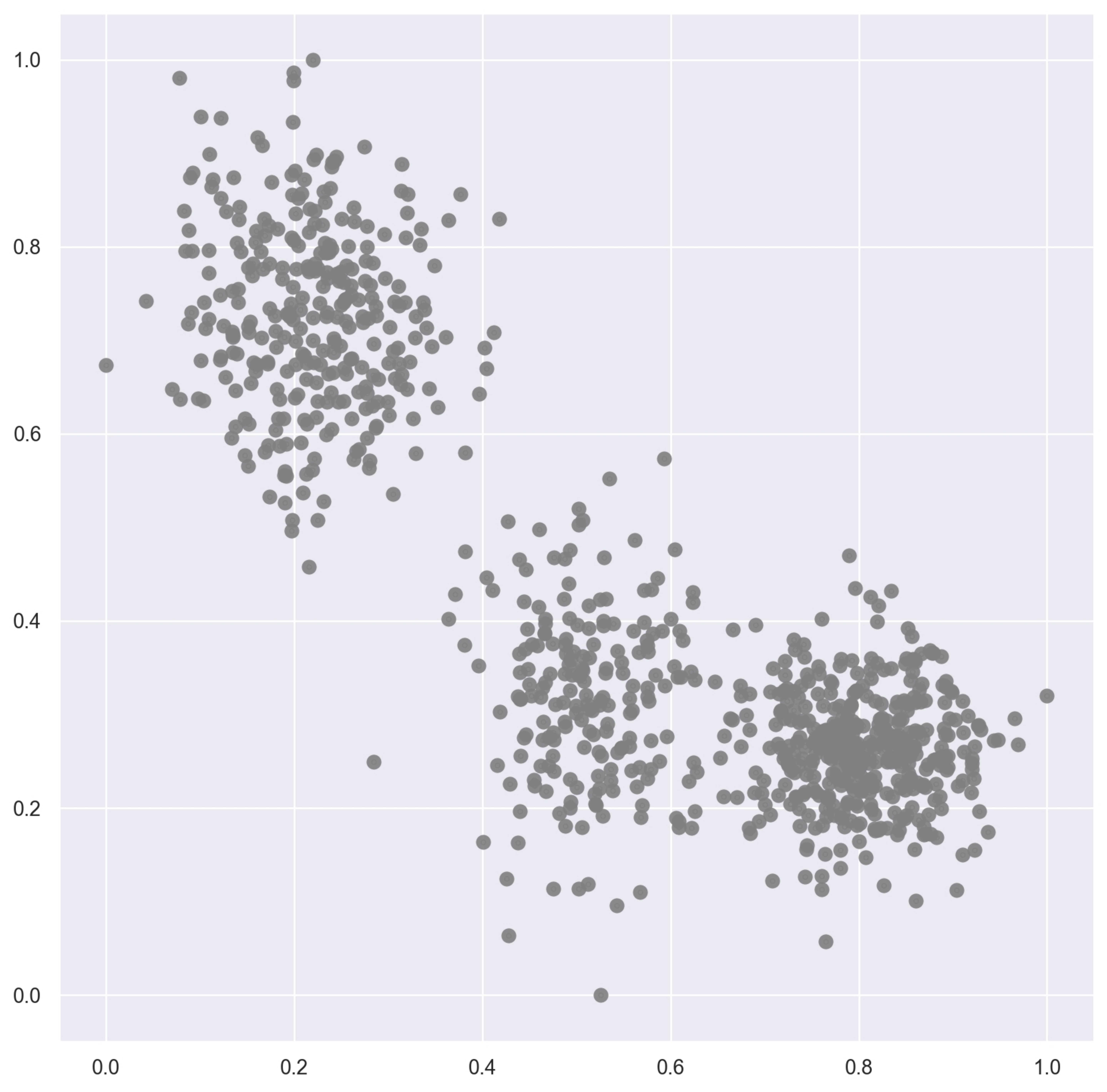}%
		\label{fig_4_1}}
	\subfloat[]{\includegraphics[width=1.65in]{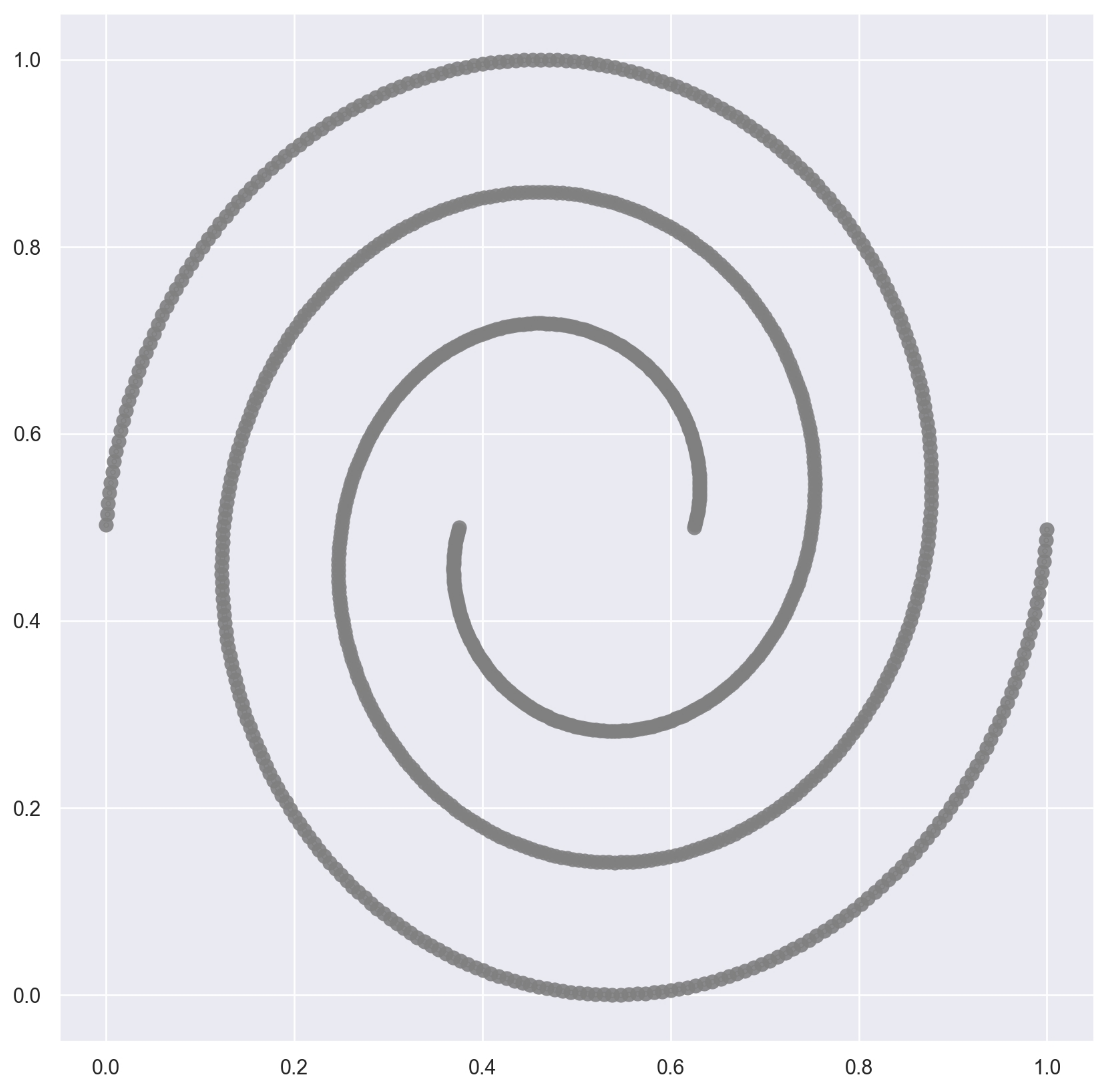}%
		\label{fig_4_2}}
	\subfloat[]{\includegraphics[width=1.65in]{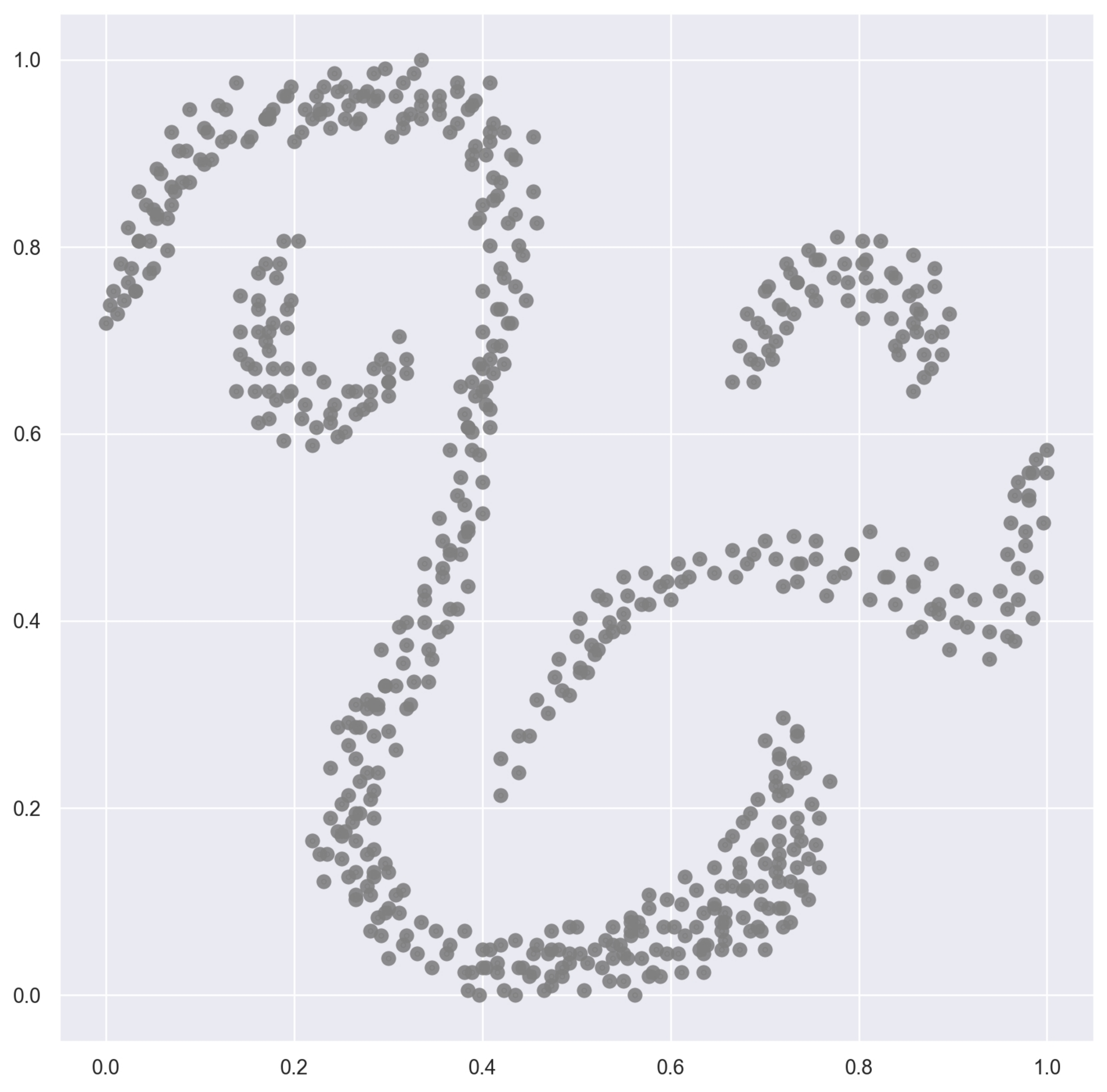}%
		\label{fig_4_3}}
	
	\subfloat[]{\includegraphics[width=1.65in]{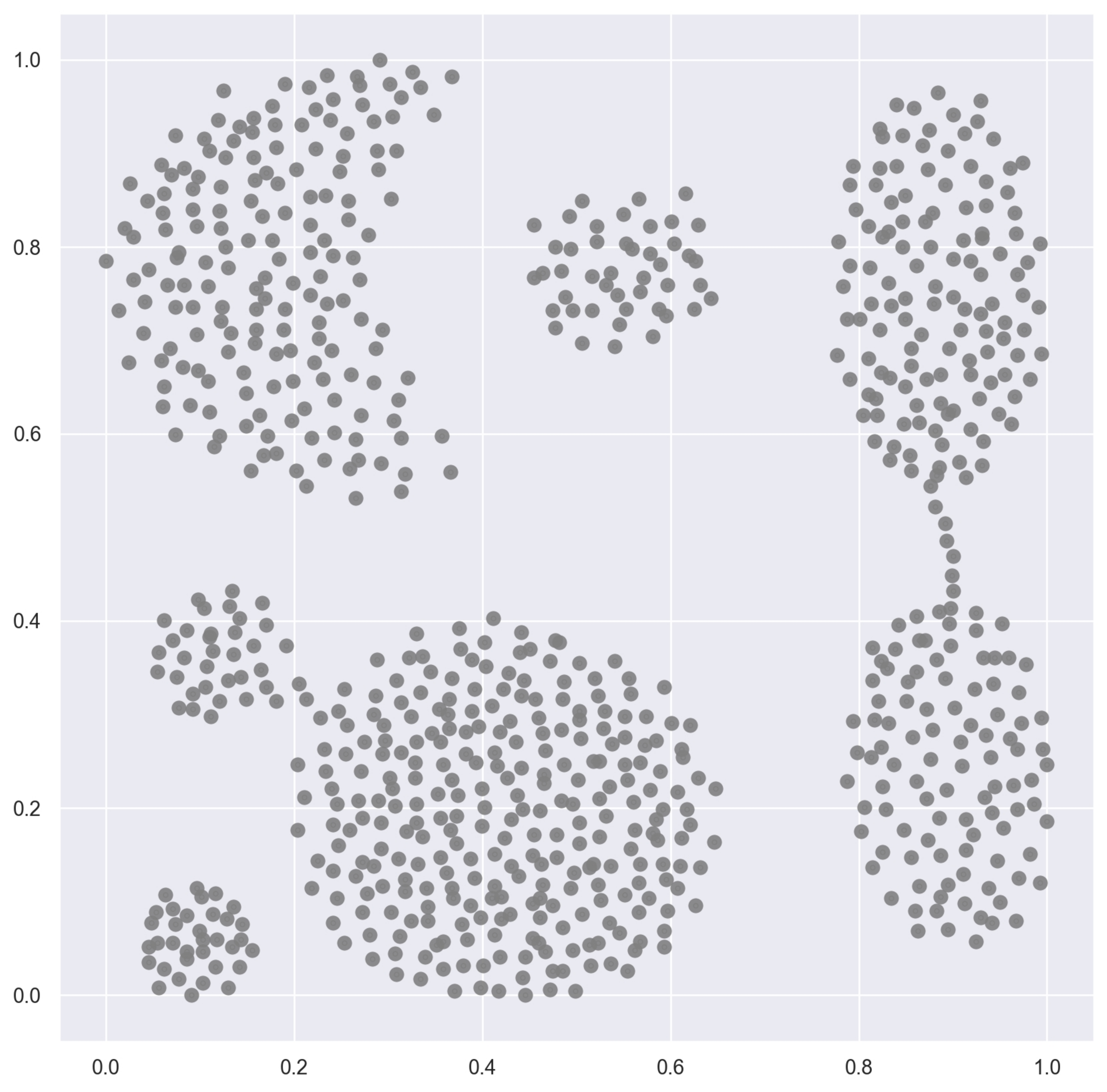}%
		\label{fig_4_4}}
	\subfloat[]{\includegraphics[width=1.65in]{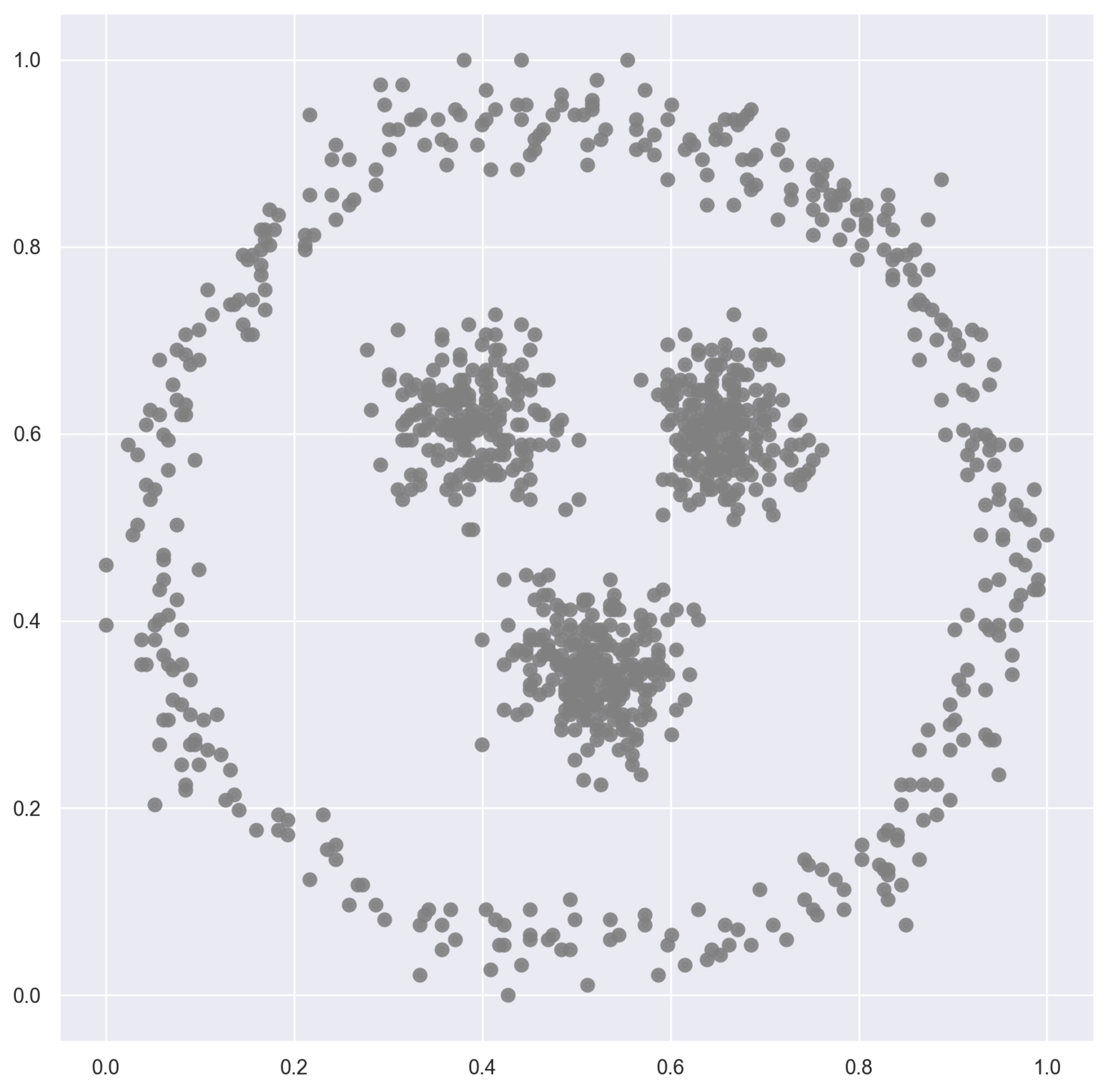}%
		\label{fig_4_5}}
	\subfloat[]{\includegraphics[width=1.65in]{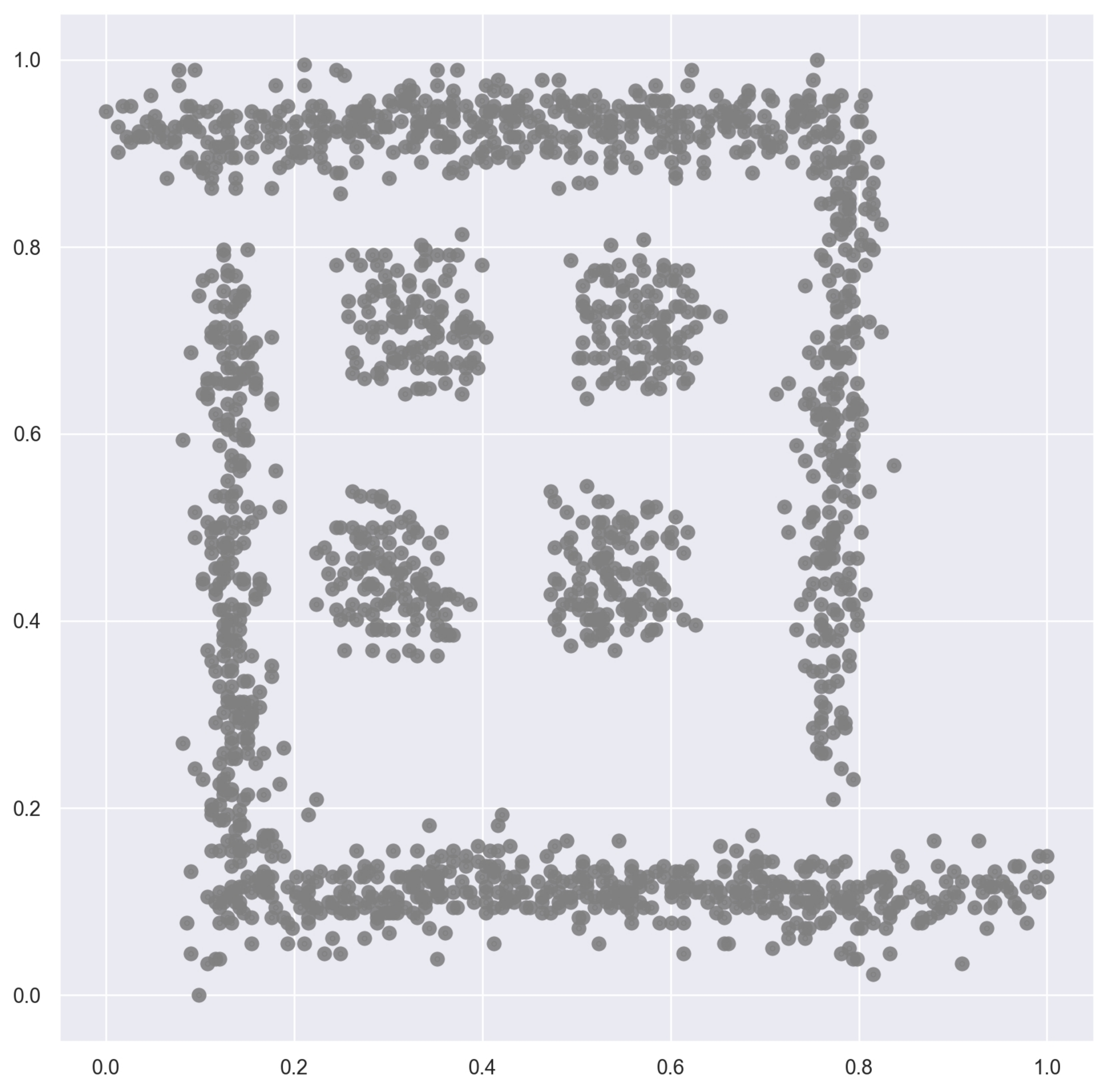}%
		\label{fig_4_6}}
	\caption{The original synthetic data sets without noise. (a) D1. (b) D2. (c) D3. (d) D4. (e) D5. (6) D6.}
	\label{fig_4}
\end{figure*}
\begin{figure*}[!h]
	\centering
	\subfloat[]{\includegraphics[width=1.65in]{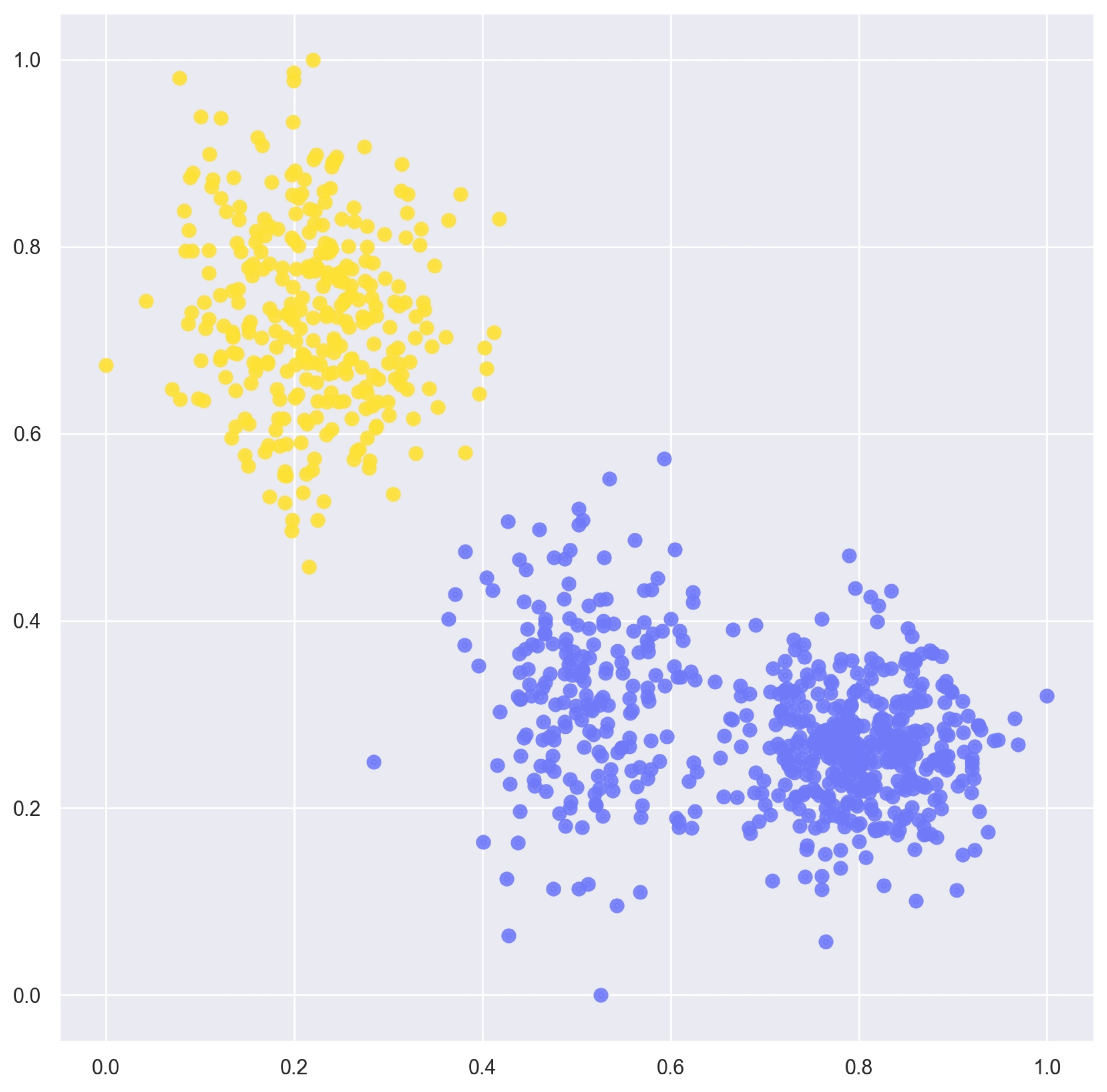}%
		\label{fig_5_1}}
	\subfloat[]{\includegraphics[width=1.65in]{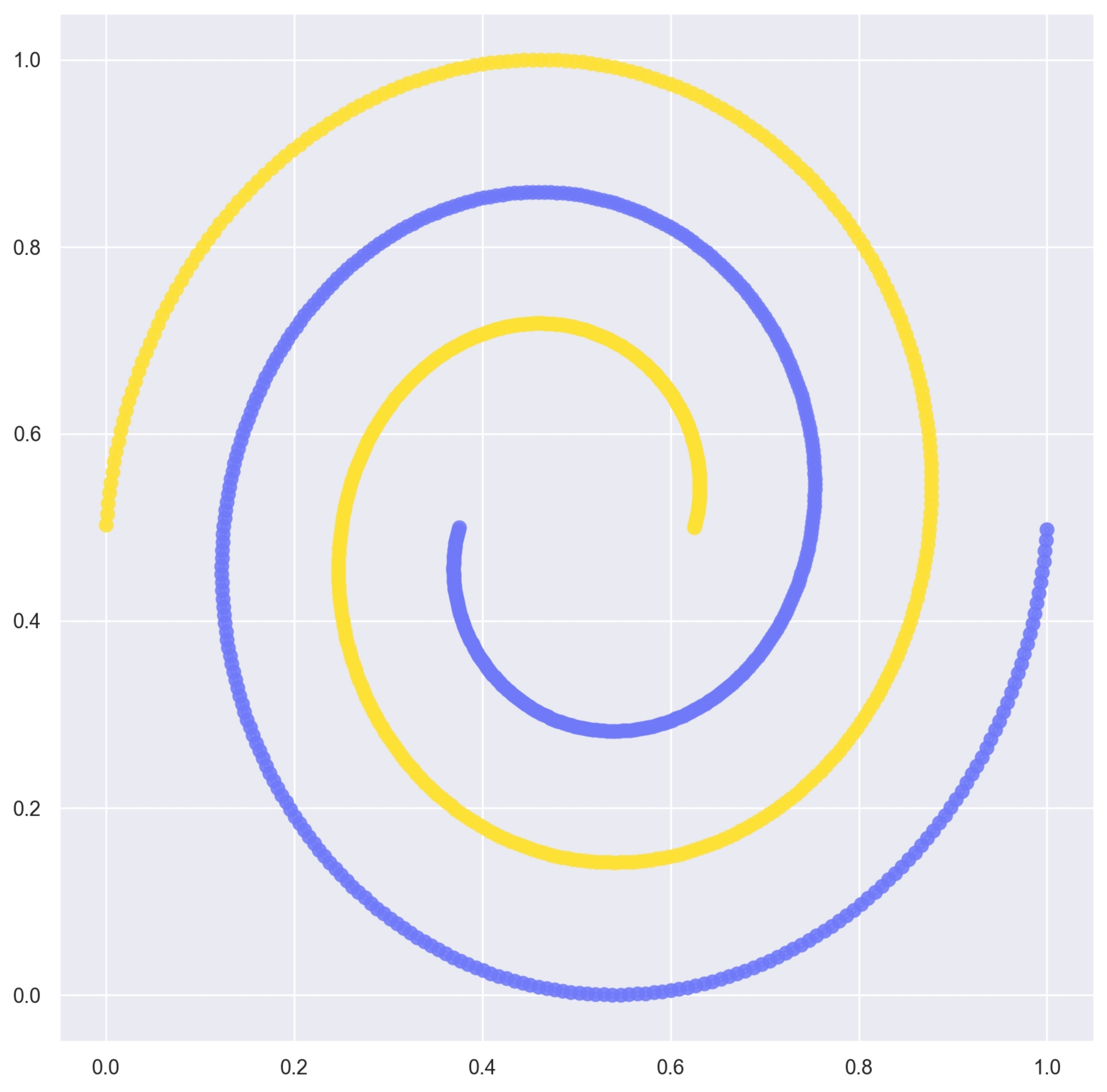}%
		\label{fig_5_2}}
	\subfloat[]{\includegraphics[width=1.65in]{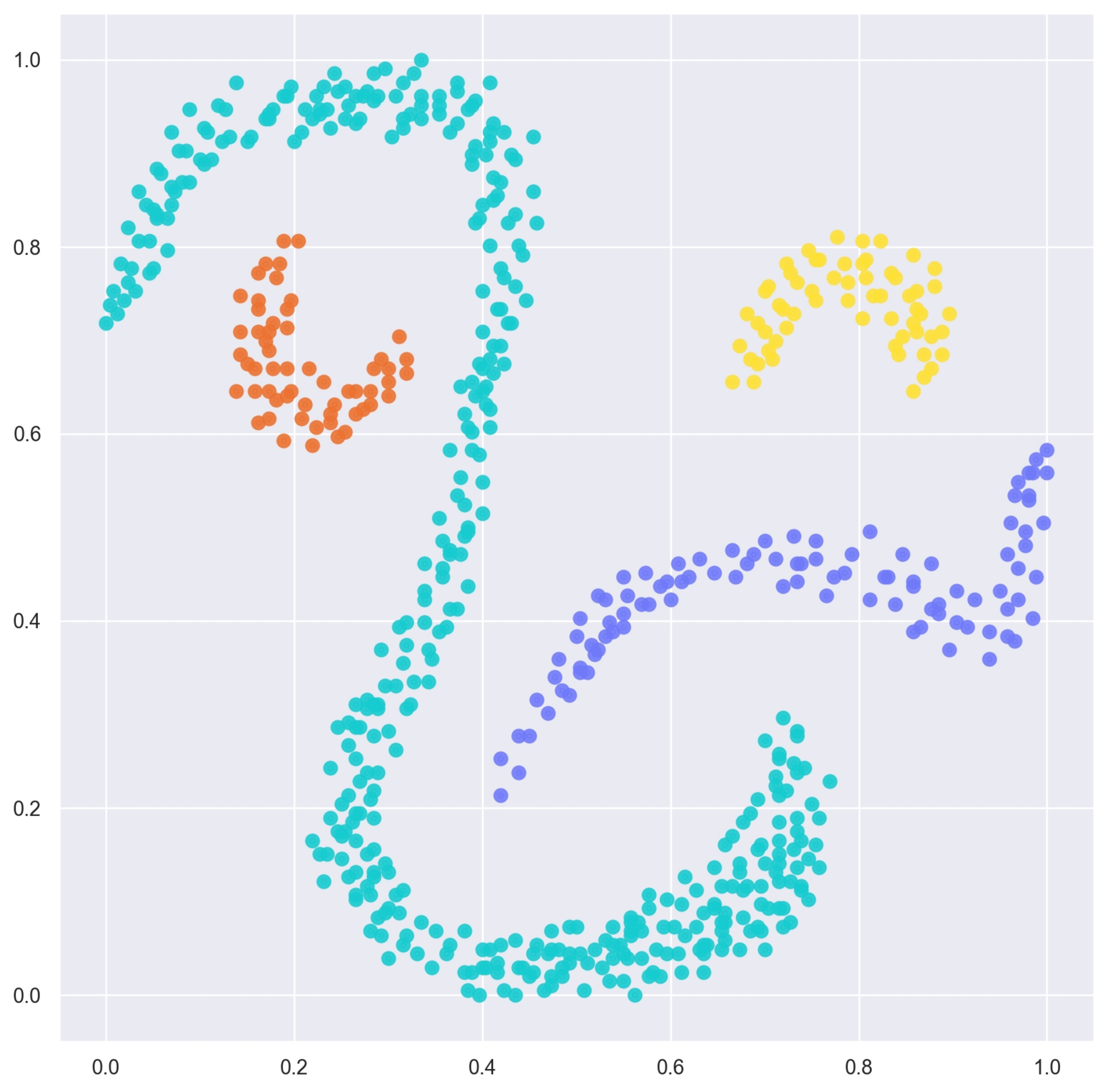}%
		\label{fig_5_3}}
	
	\subfloat[]{\includegraphics[width=1.65in]{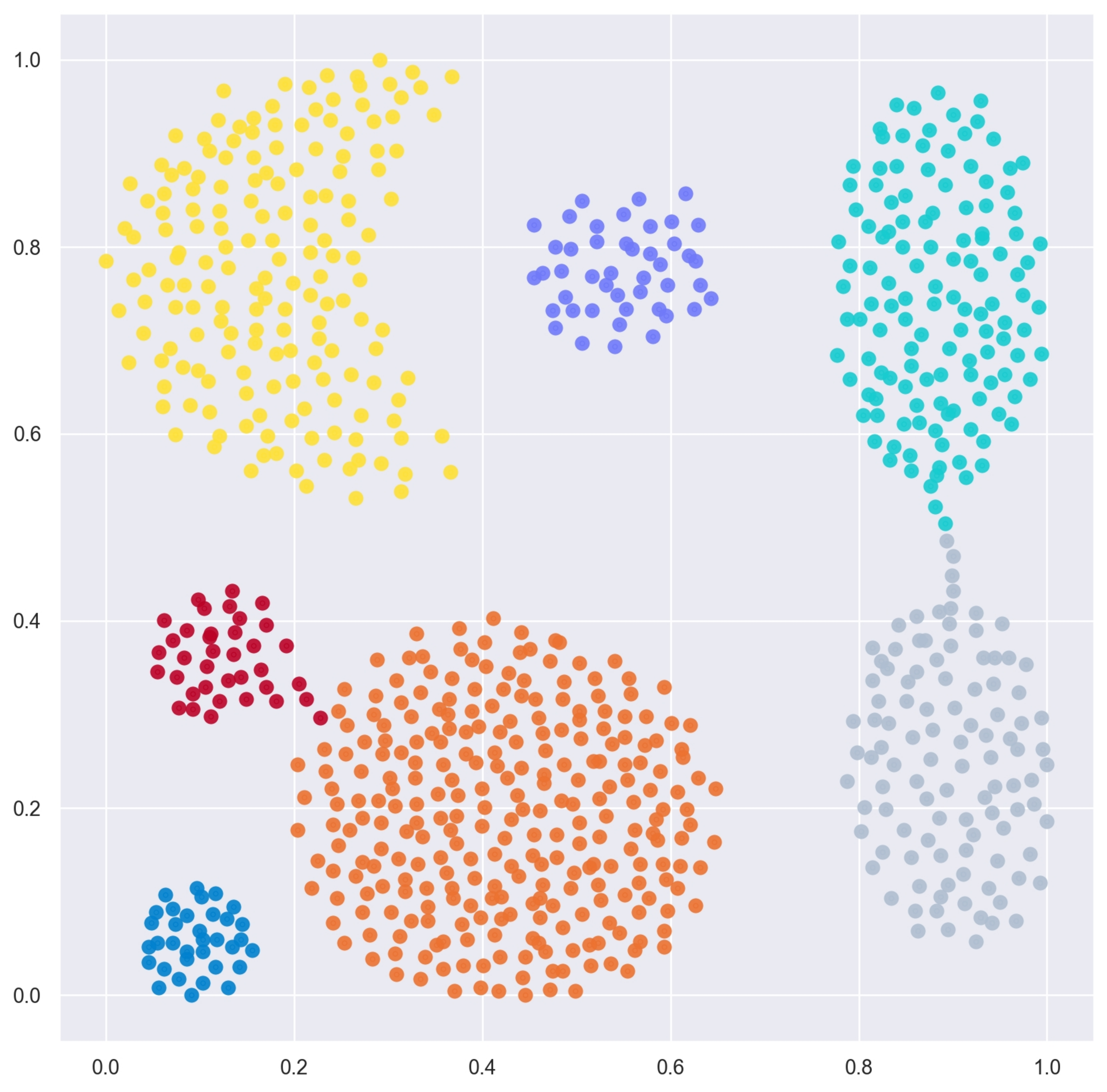}%
		\label{fig_5_4}}
	\subfloat[]{\includegraphics[width=1.65in]{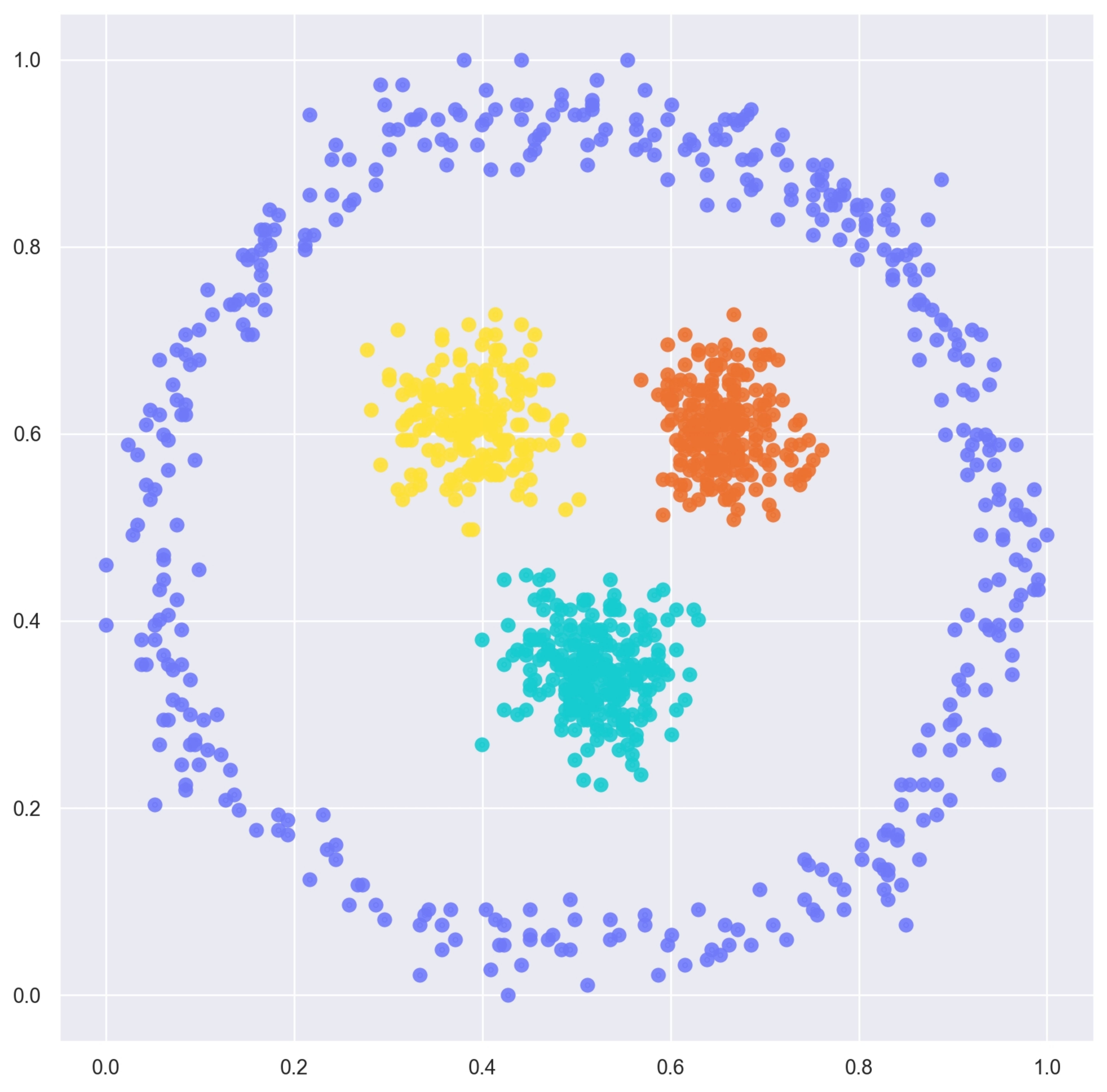}%
		\label{fig_5_5}}
	\subfloat[]{\includegraphics[width=1.65in]{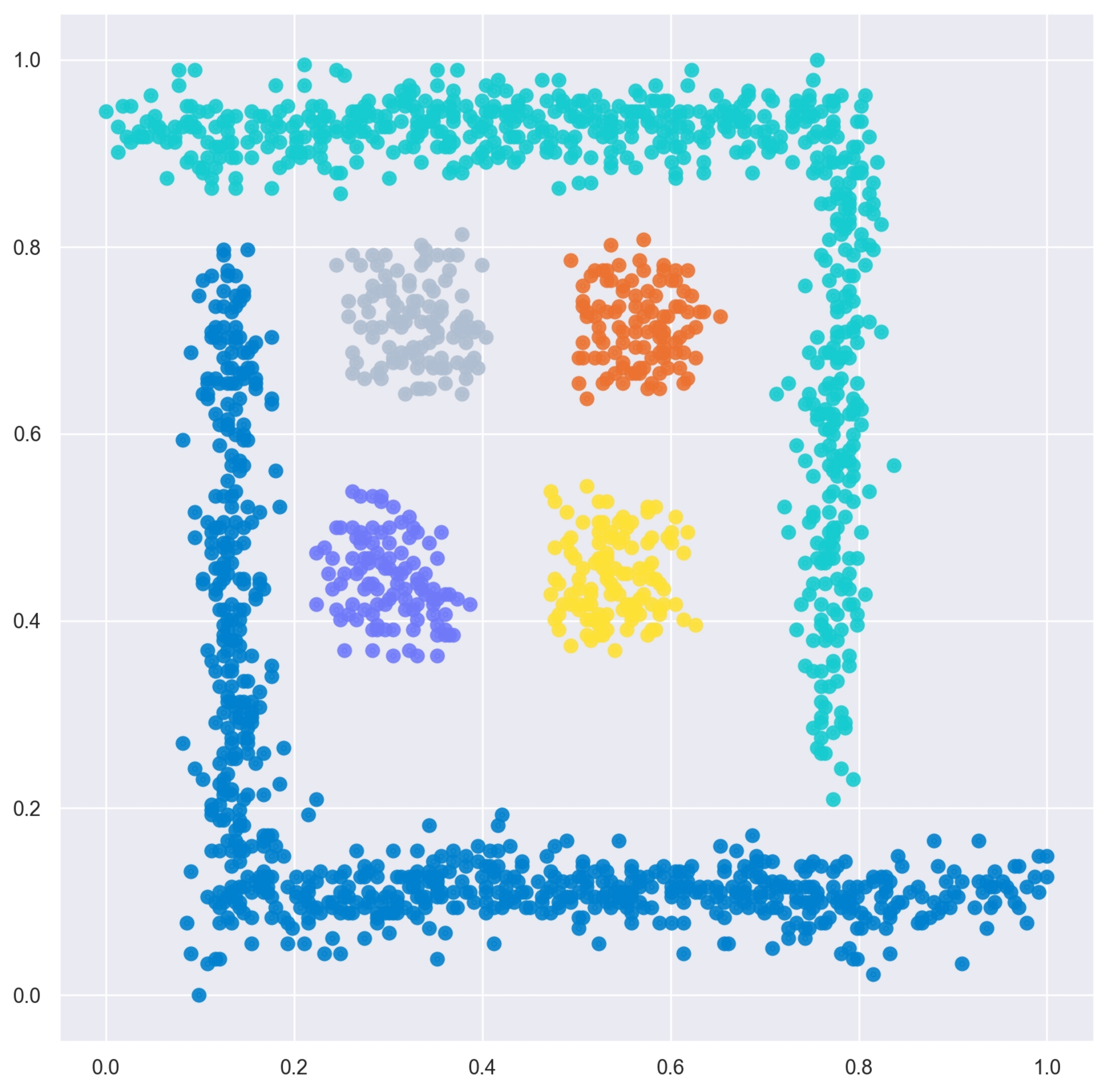}%
		\label{fig_5_6}}
	\caption{The clustering results of GBMST on synthetic data set without noise.}
	\label{fig_5}
\end{figure*}
K-means, Normal\_MST and GBMST only need to set the desired number of clusters K. The settings of the two parameters of DBSCAN have a great influence on the clustering results. In order to obtain better clustering results, it is necessary to repeatedly test with different values.The cutoff distance dc for DP is set to be 2\%. The parameters of LDP\_MST include the desired number K and the Minsize. According to\cite{Cheng2019Clustering}, the K for LDP\_MST is set to the true number of clusters and the Minsize is set to 0.018. The detailed parameter settings for synthetic data set without noise are shown in Table 2.

As shown in Fig.5, the clustering result of GBMST on synthetic data sets without noise indicate that GBMST can get ideal results for all the synthetic data sets without noise. Fig.1-Fig.5 of the supplementary shows the results of the other five comparison algorithms on the noise-free synthetic dataset. Fig.1 displays the clustering results of K-means. We can see that K-means cannot deal with manifold data sets. The clustering results shown in Fig.2 indicate that DBSCAN fail to detect clusters in Dataset 3 and Dataset 4. The clustering results in Fig.3 show that DP is not suitable to manifold data sets. Due to the interference from boundary points, as shown in Fig.4a and Fig.4d, Normal\_MST cannot get correct clustering results. But the results on several other data sets in Fig.4 show that Normal\_MST is able to handle complex manifold data. As shown in Fig.5, by setting two parameters, K and Minsize, LDP\_MST can get correct clustering results in several data sets.
 
\subsection{Experiment on Synthetic Data Sets with Noise}

\begin{table}[!h]
	\caption{Experiment data sets with noise.\label{table3}}
	\renewcommand\arraystretch{1.4}
	\centering
	\begin{tabular}{lcccc}
		\hline              & Instances  & Noises & Clusters & Source  \\
		\hline  D7   & 1043       & 43    & 2        & \cite{Ha2015A,Ha2014Robust}   \\
	        	D8   & 1039       & 41    & 4        & \cite{Ha2015A,Ha2014Robust}   \\
		        D9   & 1641       & 45    & 3        & \cite{Ha2015A,Ha2014Robust}   \\
		        D10  & 1427       & 71    & 4        & \cite{Ha2015A,Ha2014Robust}   \\
		\hline
	\end{tabular}
\end{table}
\begin{table}[!h]
	\caption{Parameters settings on Synthetic Data Sets with Noise.\label{table4}}
	\renewcommand\arraystretch{1.4}
	\centering
	\begin{tabular}{lccc}
		\hline             & K-means  & DBSCAN           & DP \\
		\hline D7   & K=2     & Minpts=4, Eps=0.04 & K=2, dc=2\%   \\
	       	   D8   & K=4     & Minpts=4, Eps=0.04 & K=2, dc=2\%   \\
		       D9   & K=3     & Minpts=4, Eps=0.03 & K=2, dc=2\%   \\
		       D10  & K=4     & Minpts=5, Eps=0.04 & K=2, dc=2\%   \\	       		       		       
		\hline
		\hline             & Normal\_MST  & LDP\_MST              & GBMST  \\
		\hline D7   & K=2         & K=2, Minsize=0.018   & K=2   \\
		       D8   & K=4         & K=4, Minsize=0.018   & K=4   \\
	           D9   & K=3         & K=3, Minsize=0.018   & K=3   \\
		       D10  & K=4         & K=4, Minsize=0.018   & K=4   \\		       
		\hline
	\end{tabular}
\end{table}
\begin{figure*}[!h]
	\centering
	\subfloat[]{\includegraphics[width=1.65in]{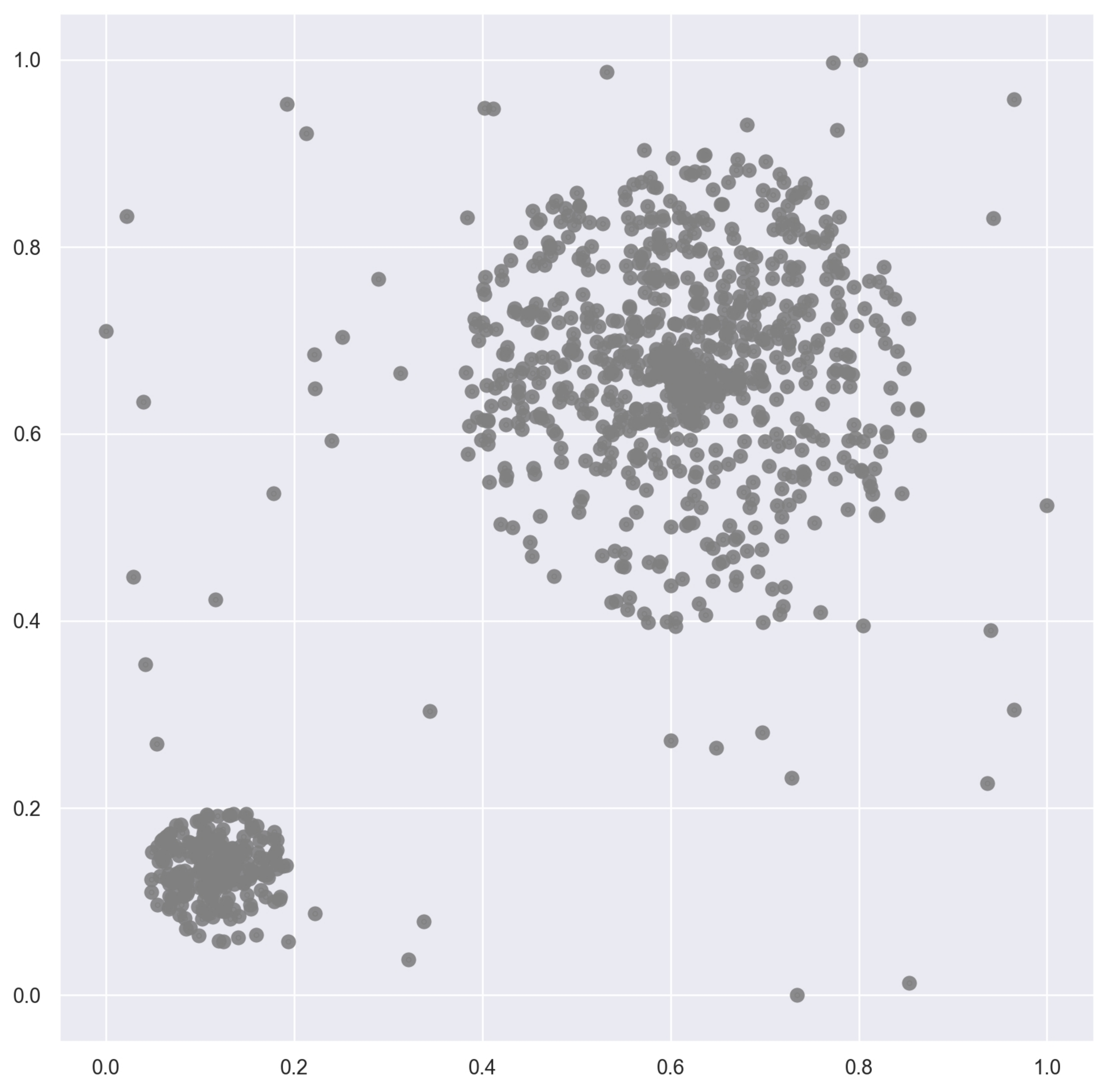}%
		\label{fig_6_1}}
	\subfloat[]{\includegraphics[width=1.65in]{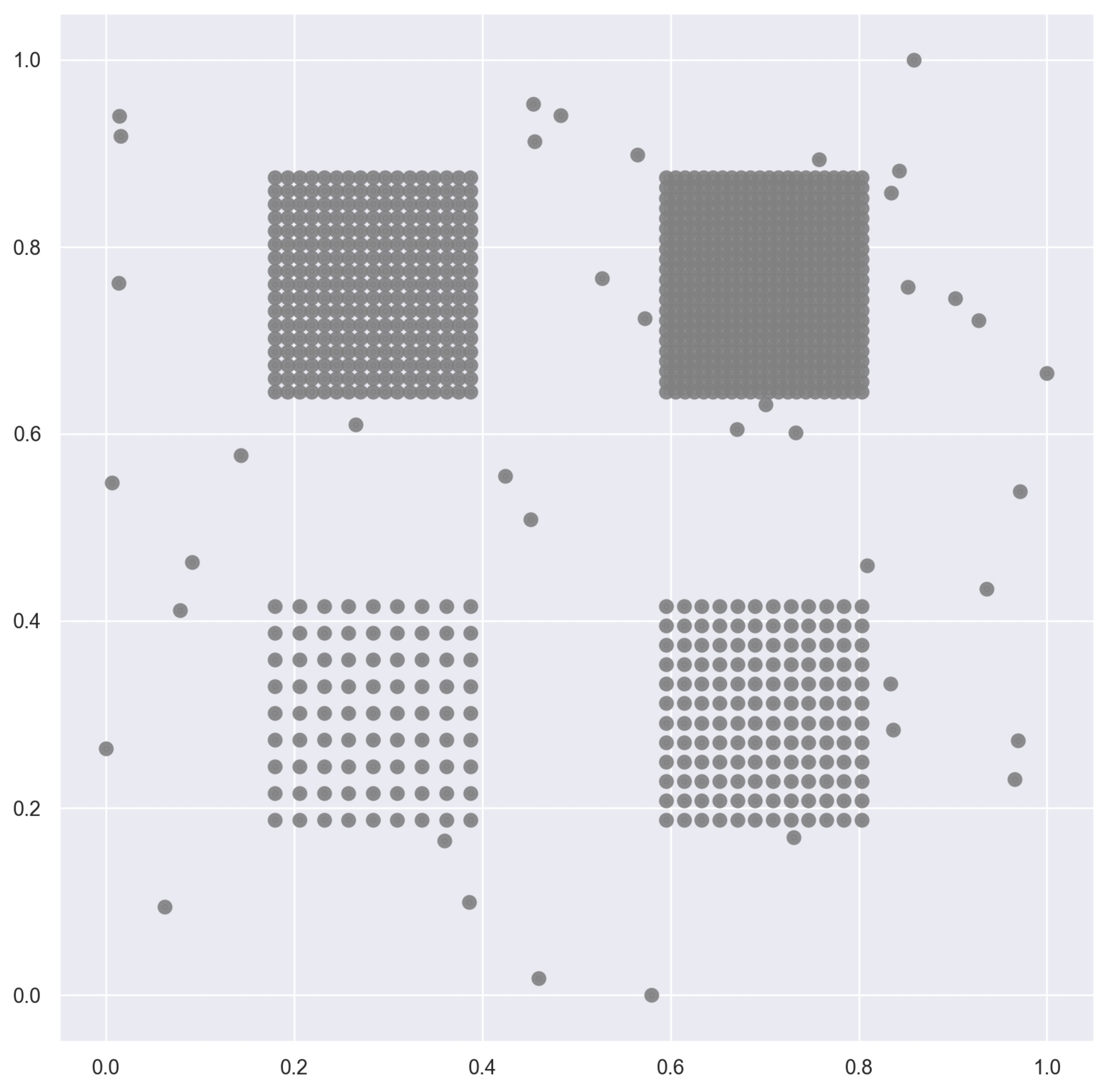}%
		\label{fig_6_2}}
	\subfloat[]{\includegraphics[width=1.65in]{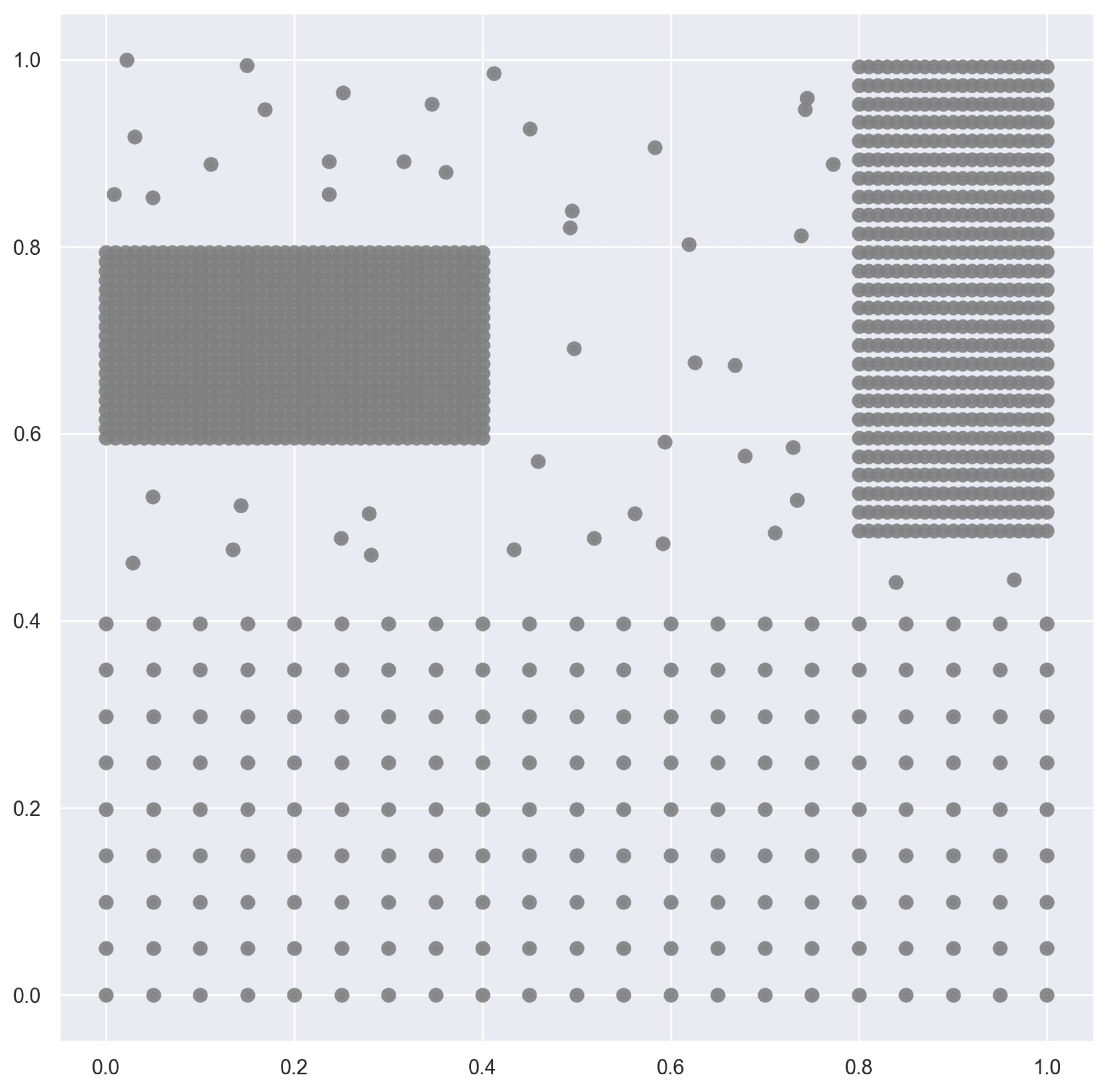}%
		\label{fig_6_3}}
	\subfloat[]{\includegraphics[width=1.65in]{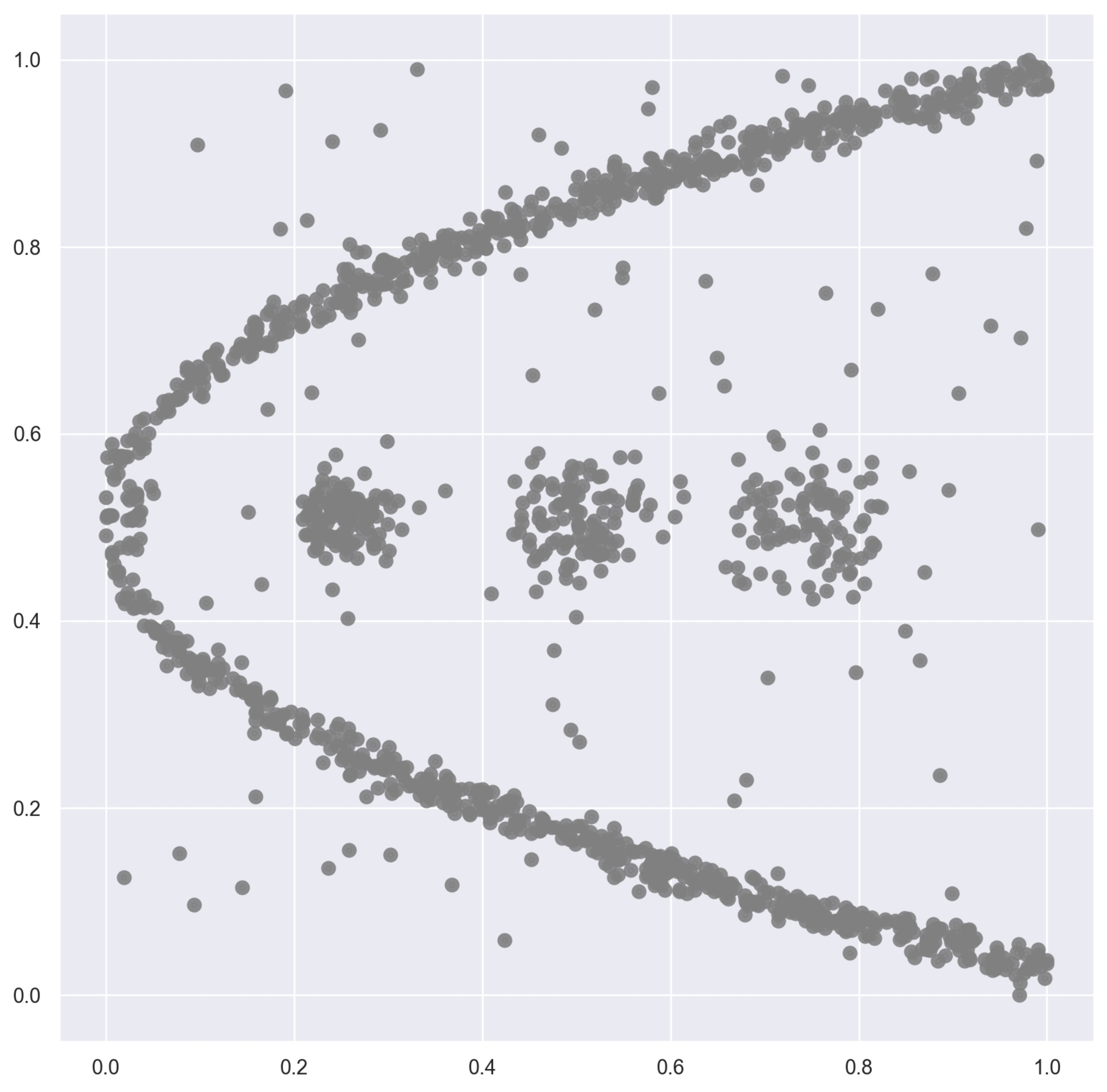}%
		\label{fig_6_4}}
	\caption{The original synthetic data sets with noise. (a) D7. (b) D8. (c) D9. (d) D10.}
	\label{fig_6}
\end{figure*}

\begin{figure*}[!h]
	\centering
	\subfloat[]{\includegraphics[width=1.65in]{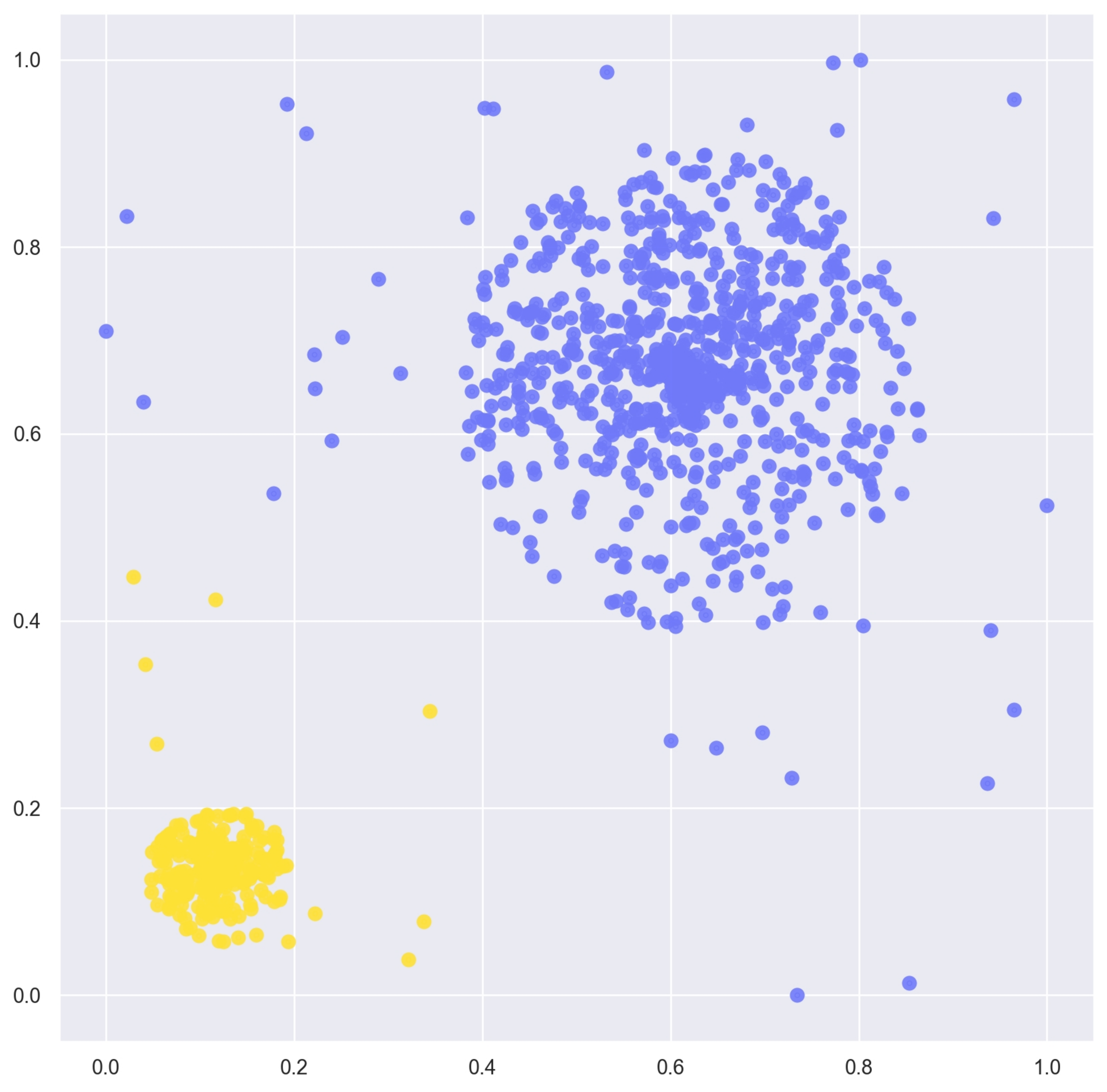}%
		\label{fig_7_1}}
	\subfloat[]{\includegraphics[width=1.65in]{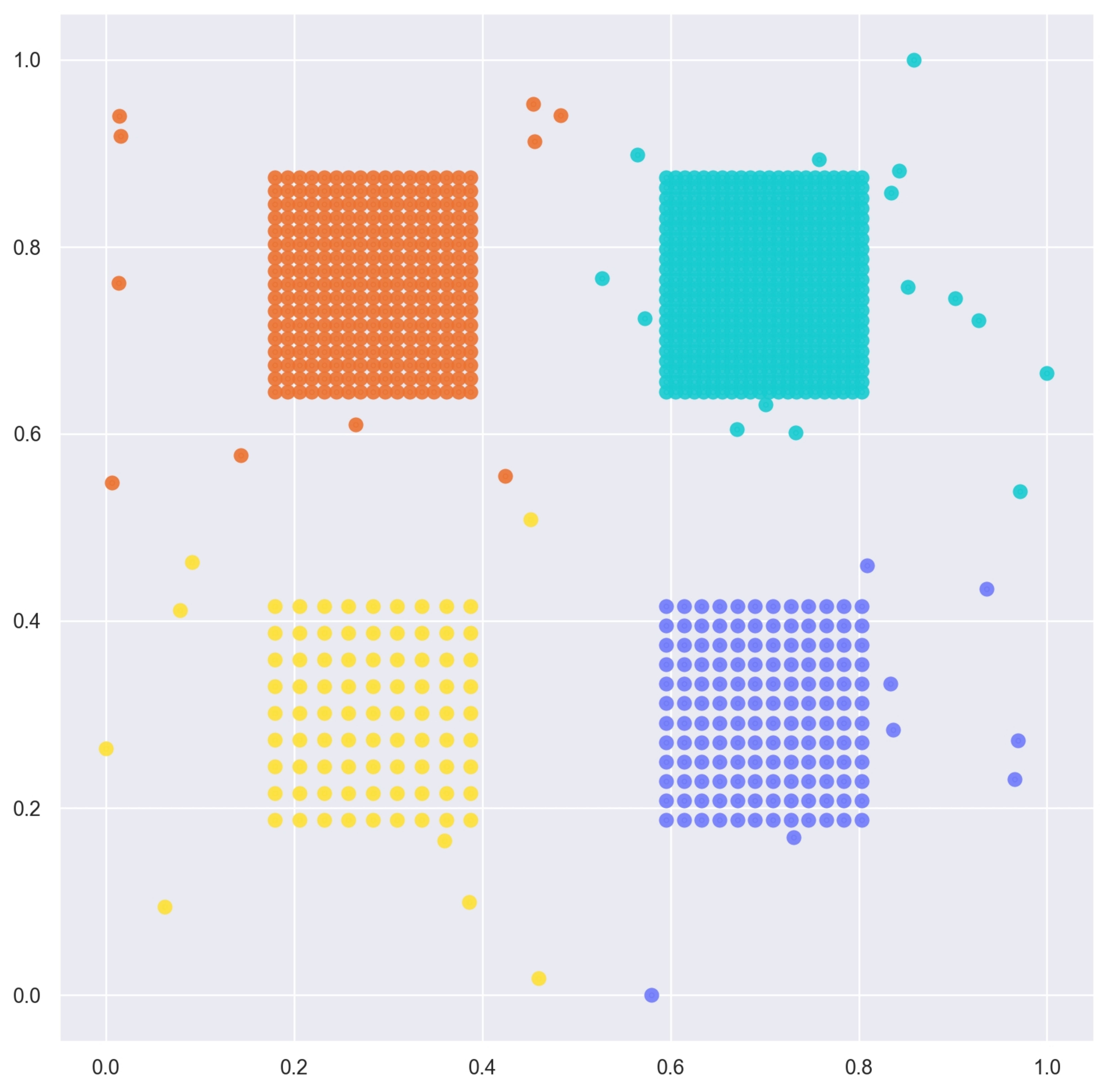}%
		\label{fig_7_2}}
	\subfloat[]{\includegraphics[width=1.65in]{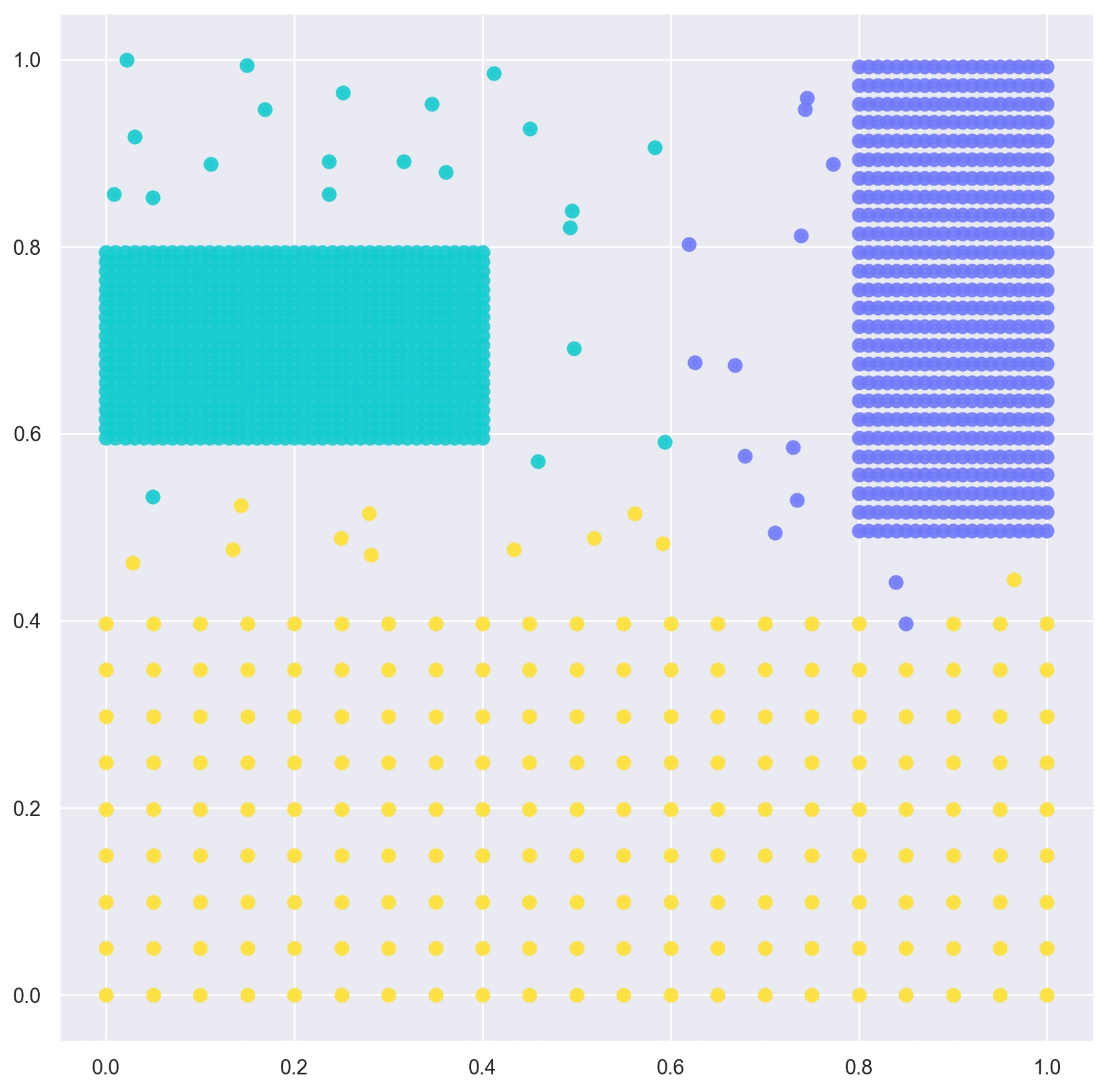}%
		\label{fig_7_3}}
	\subfloat[]{\includegraphics[width=1.65in]{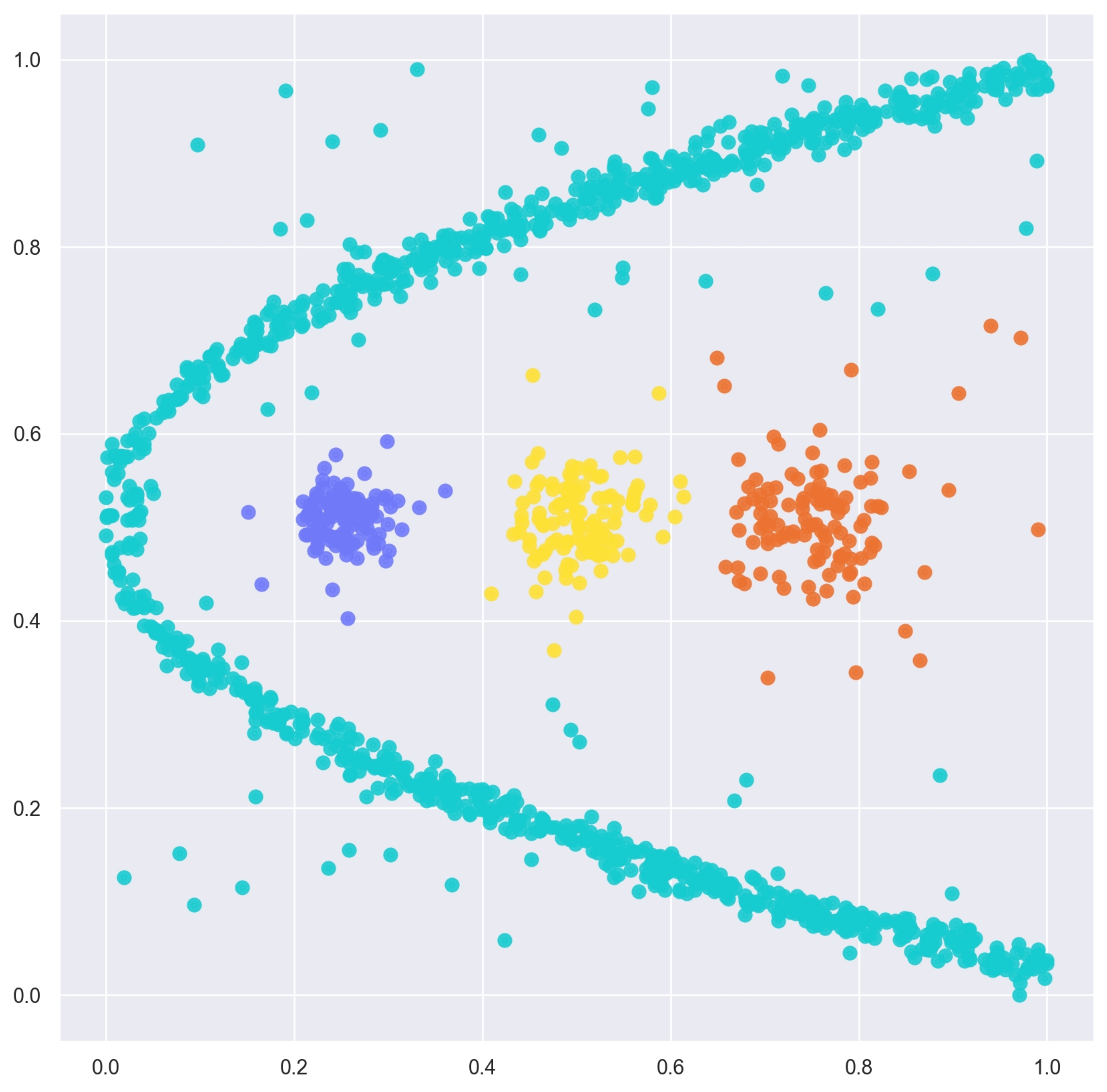}%
		\label{fig_7_4}}
	\caption{The clustering results of GBMST on synthetic data set with noise.}
	\label{fig_7}
\end{figure*}
To verify the algorithm's adaptability to noise, we conduct experiments on 4 synthetic data sets with noise.The detailed information of data sets is shown in Table 3, and the visualization of synthetic data sets is shown in Fig.6. The detailed parameter settings for synthetic data set with noise are shown in Table 4.

Since GBMST uses hypers-balls to build a MST, it is more robust to noise, which is illustrated by the clustering results in Fig.7. The clustering results of the other five comparison algorithms on the noisy synthetic dataset are shown in Supplementary Fig.6-Fig.10. As shown in Fig.6c and Figure.8c, k-means and DP algorithms cannot obtain correct clustering results on data with large variations in density. The results of DBSCAN are shown in Figure 7, which shows that DBSCAN is robust to noise and can identify clusters with more complex shapes, but it requires manual adjustment of two parameters and cannot handle data with large density changes. As shown in Figure 9, traditional MST clustering Normal\_MST fails on all four noisy data due to noise interference. By combining density kernel and MST, LDP\_MST can eliminate the interference of noise, as shown in Fig.10, its clustering results are robust to noises.

From the clustering results of the above synthetic datasets, it can be seen that the DBSCAN algorithm can handle clusters of complex shapes without being sensitive to noise, but cannot handle datasets of various densities. But K-means and DP algorithms cannot handle complex-shaped clusters. Normal\_MST are suitable for complex manifold data, but are sensitive to noise and boundary points. By setting Minsize, LDP\_MST can handle data containing complex manifolds and noise. Based on hypber-balls, GBMST can obtain suitable clustering results for all datasets with large density variation or datasets containing multiple clusters and noise.
\subsection{Clustering on Biomedical Data Set}
\begin{figure*}[!h]
	\centering
	\subfloat[Ground Truth]{\includegraphics[width=1.9in,height=1.6in]{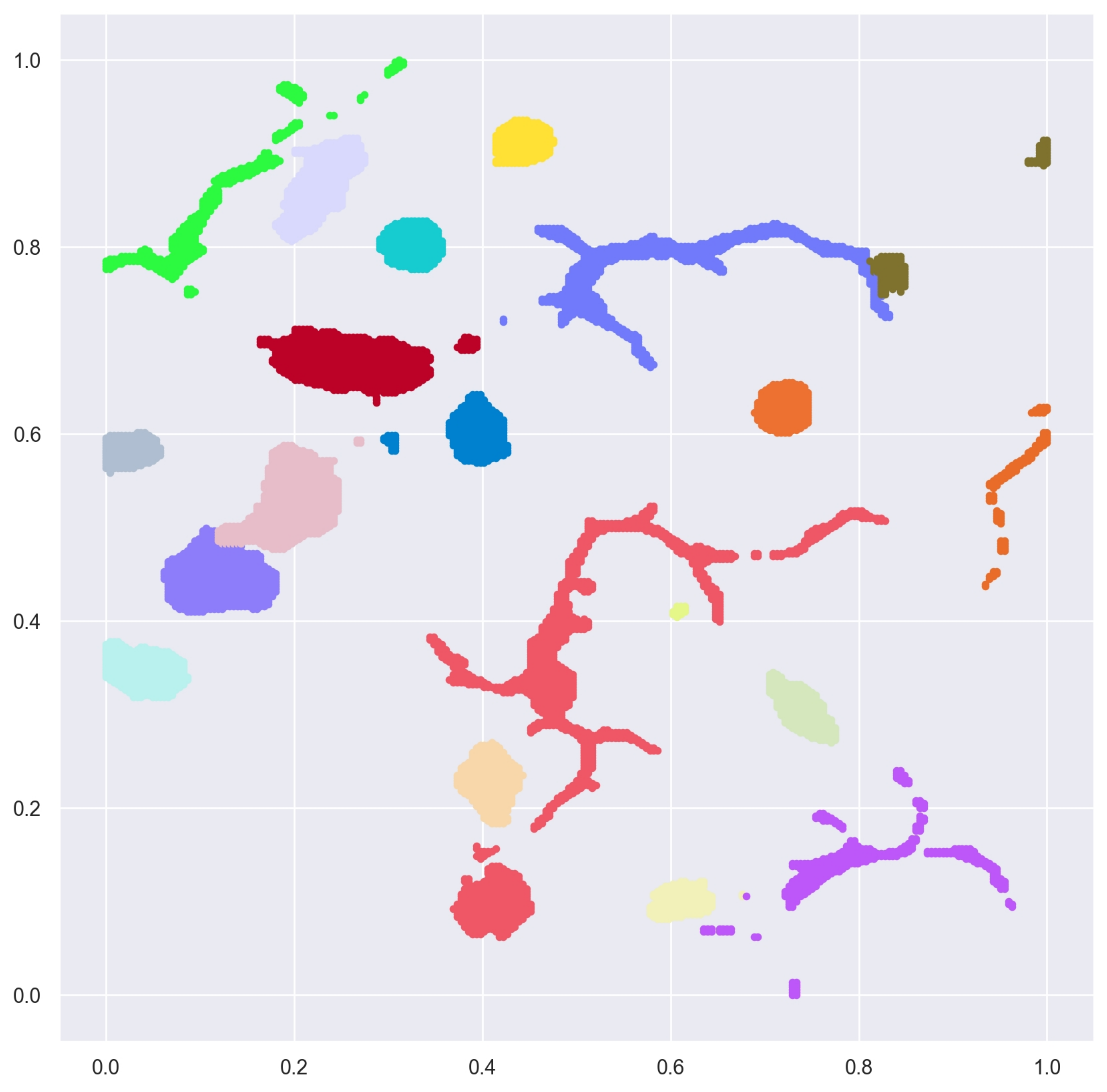}%
		\label{fig_9_1}}
	\subfloat[K-means]{\includegraphics[width=1.6in]{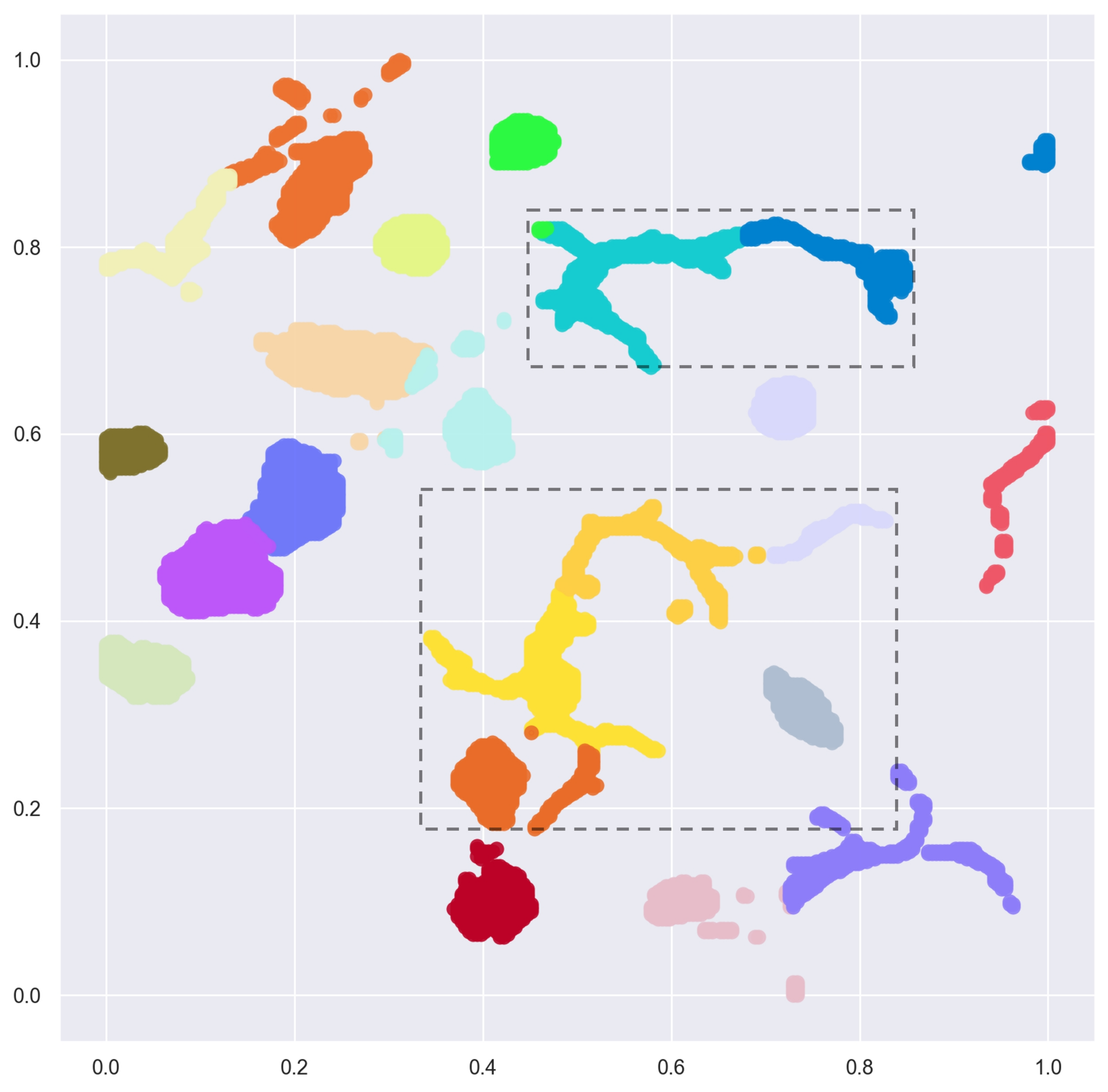}%
		\label{fig_9_2}}
	\subfloat[DBSCAN]{\includegraphics[width=1.6in]{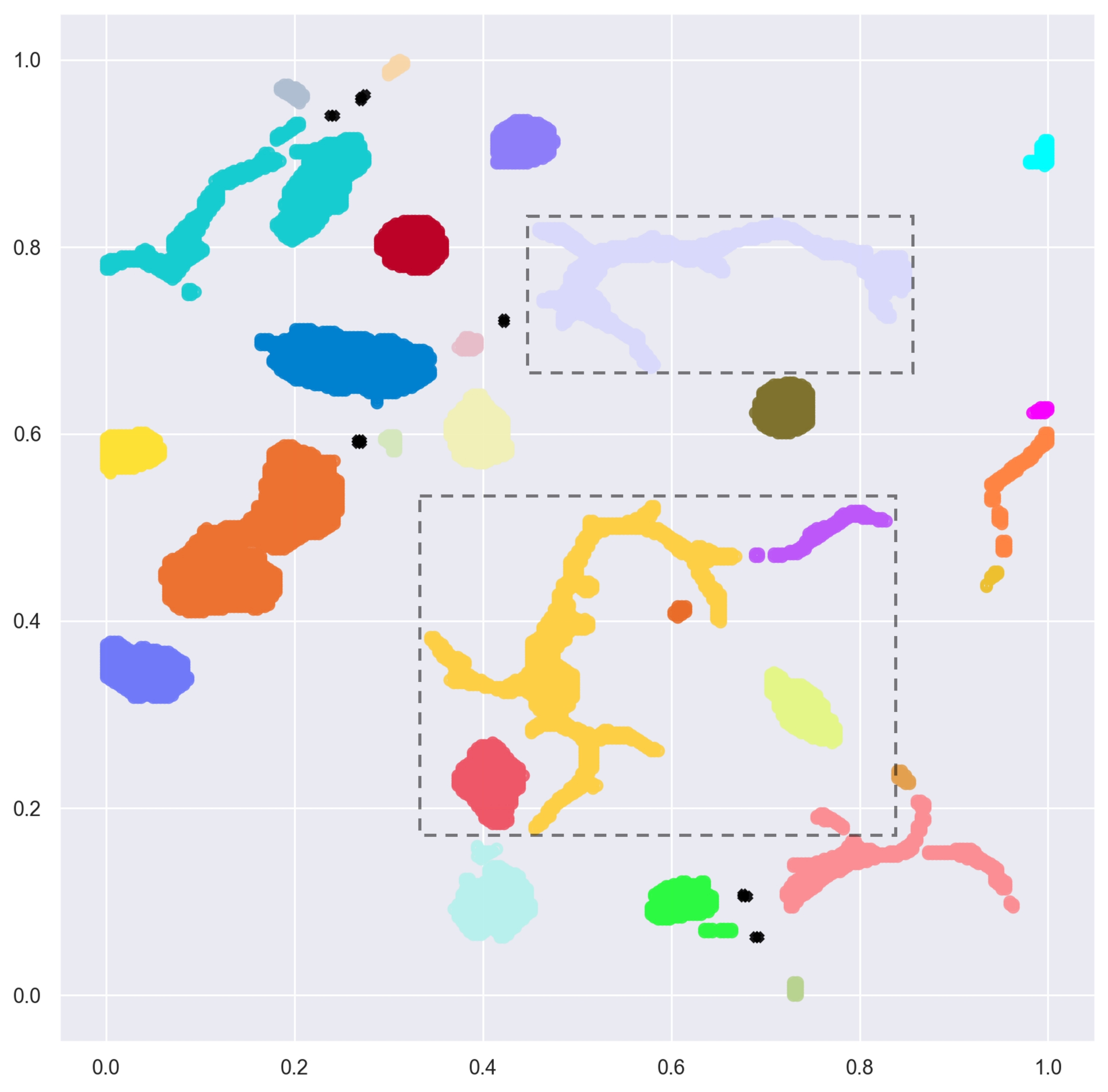}%
		\label{fig_9_3}}
	\subfloat[DP]{\includegraphics[width=1.6in]{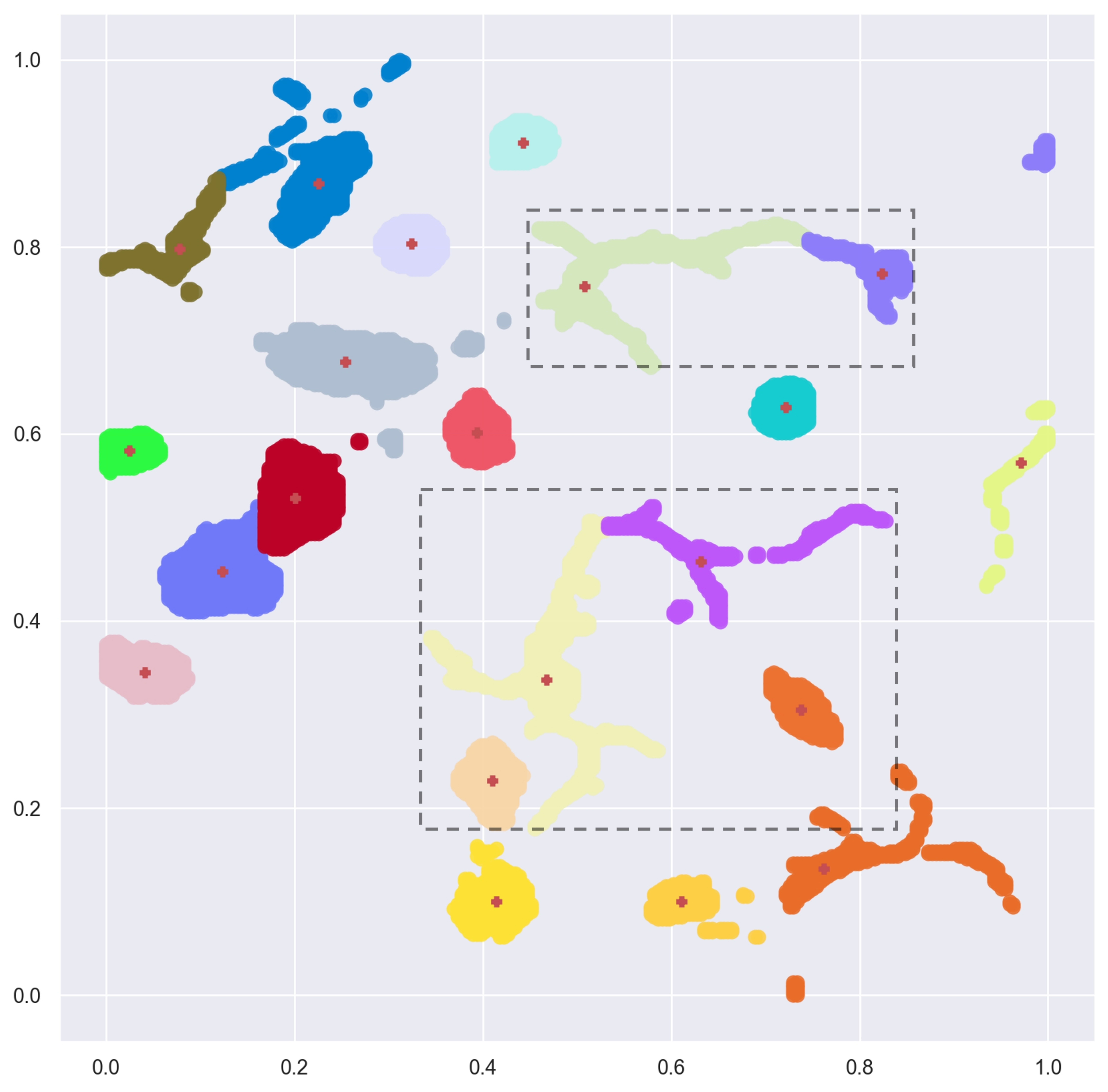}%
		\label{fig_9_4}}
	
	\subfloat[ARI\_Index]{\includegraphics[width=1.9in,height=1.6in]{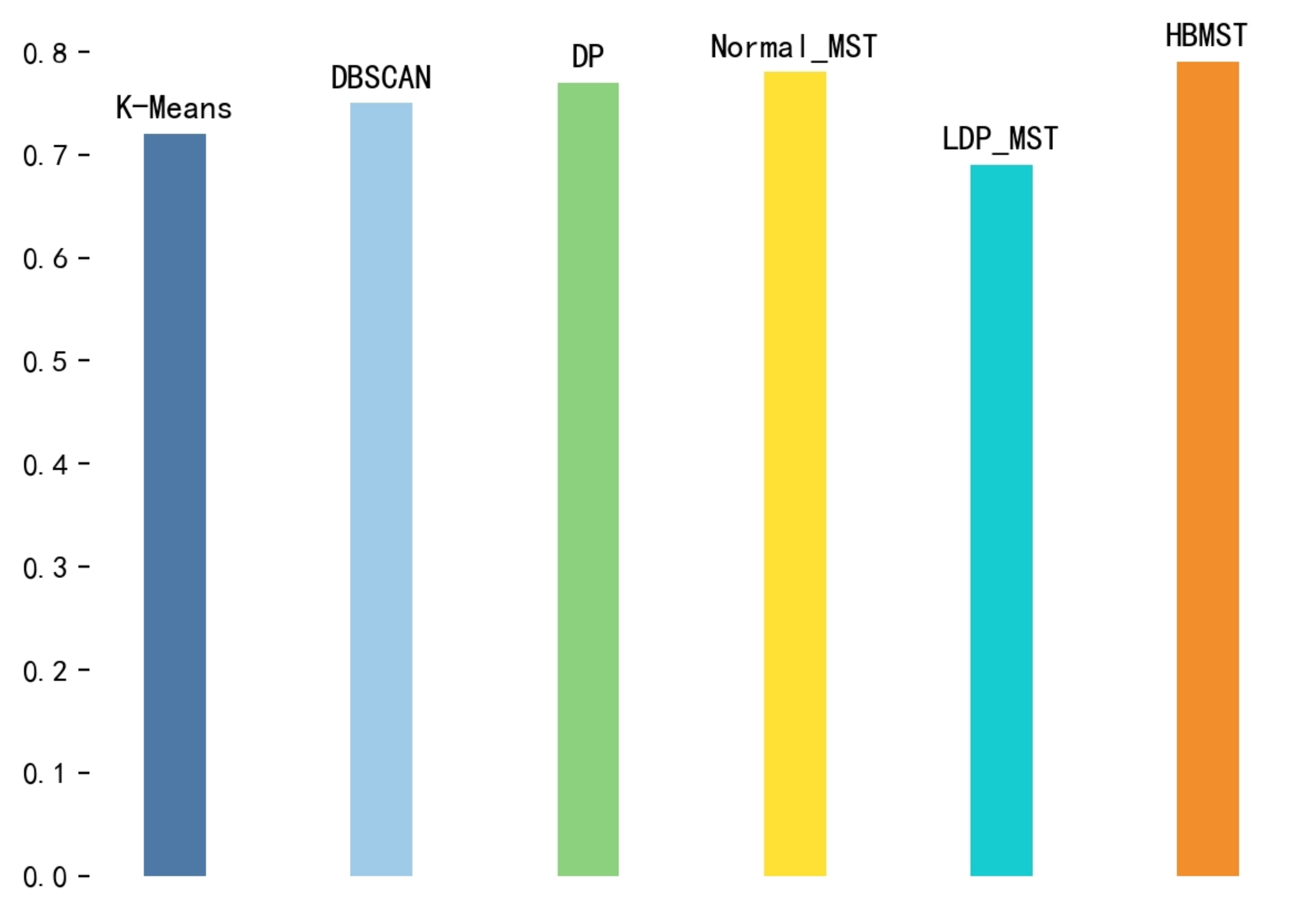}%
		\label{fig_9_5}}
	\subfloat[Normal\_MST]{\includegraphics[width=1.6in]{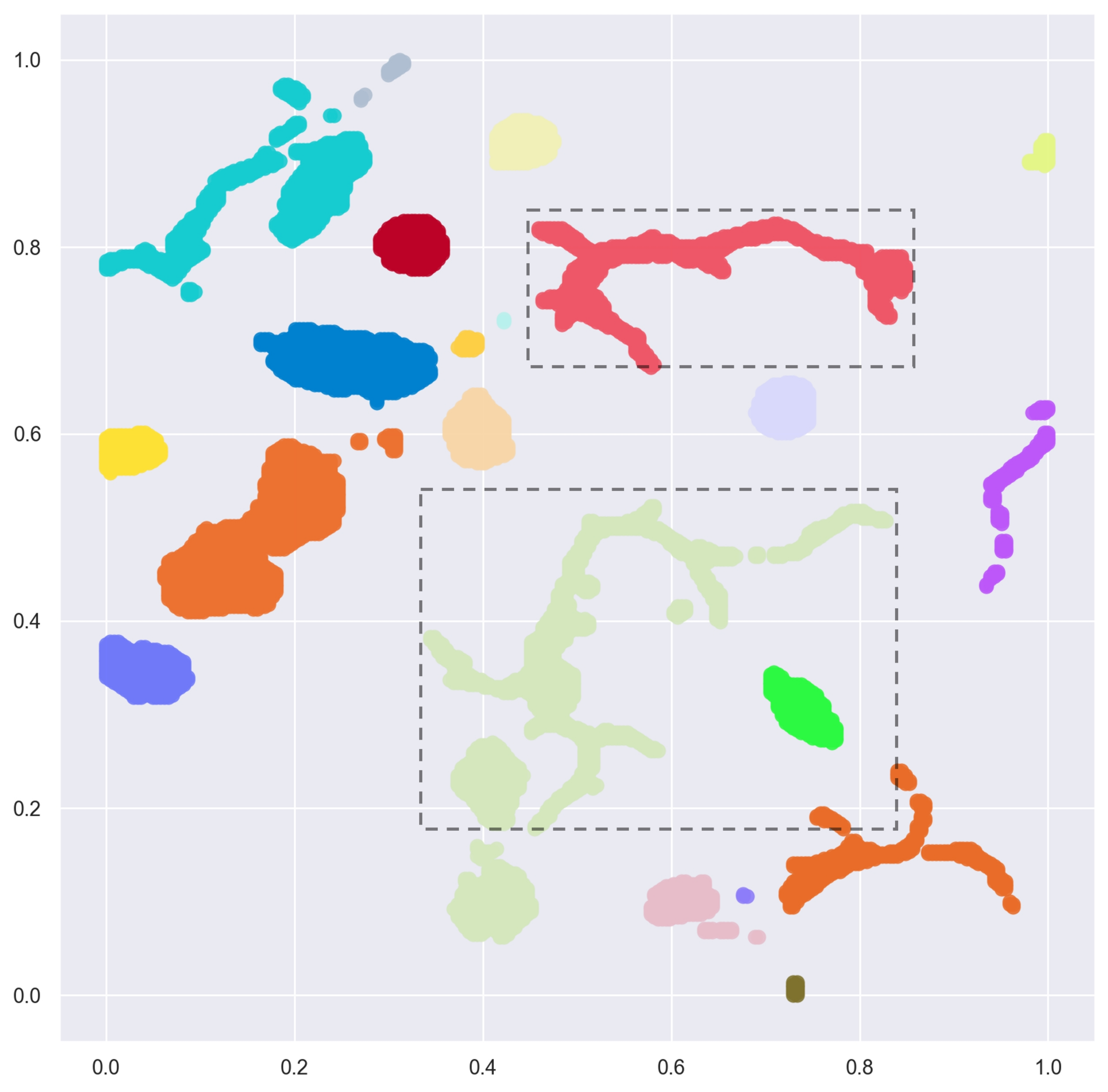}%
		\label{fig_9_6}}
	\subfloat[LDP\_MST]{\includegraphics[width=1.6in]{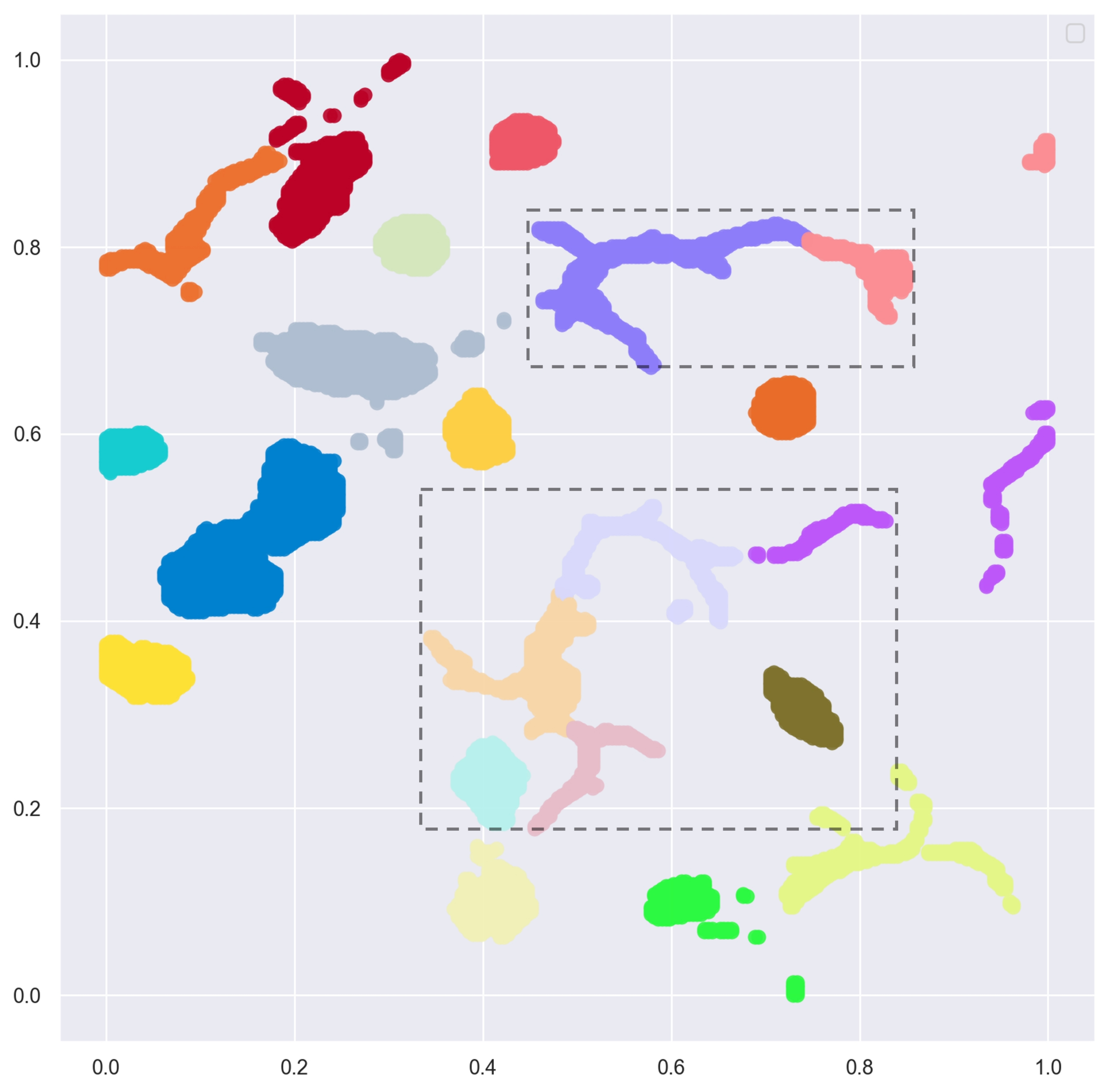}%
		\label{fig_9_7}}
	\subfloat[GBMST]{\includegraphics[width=1.6in]{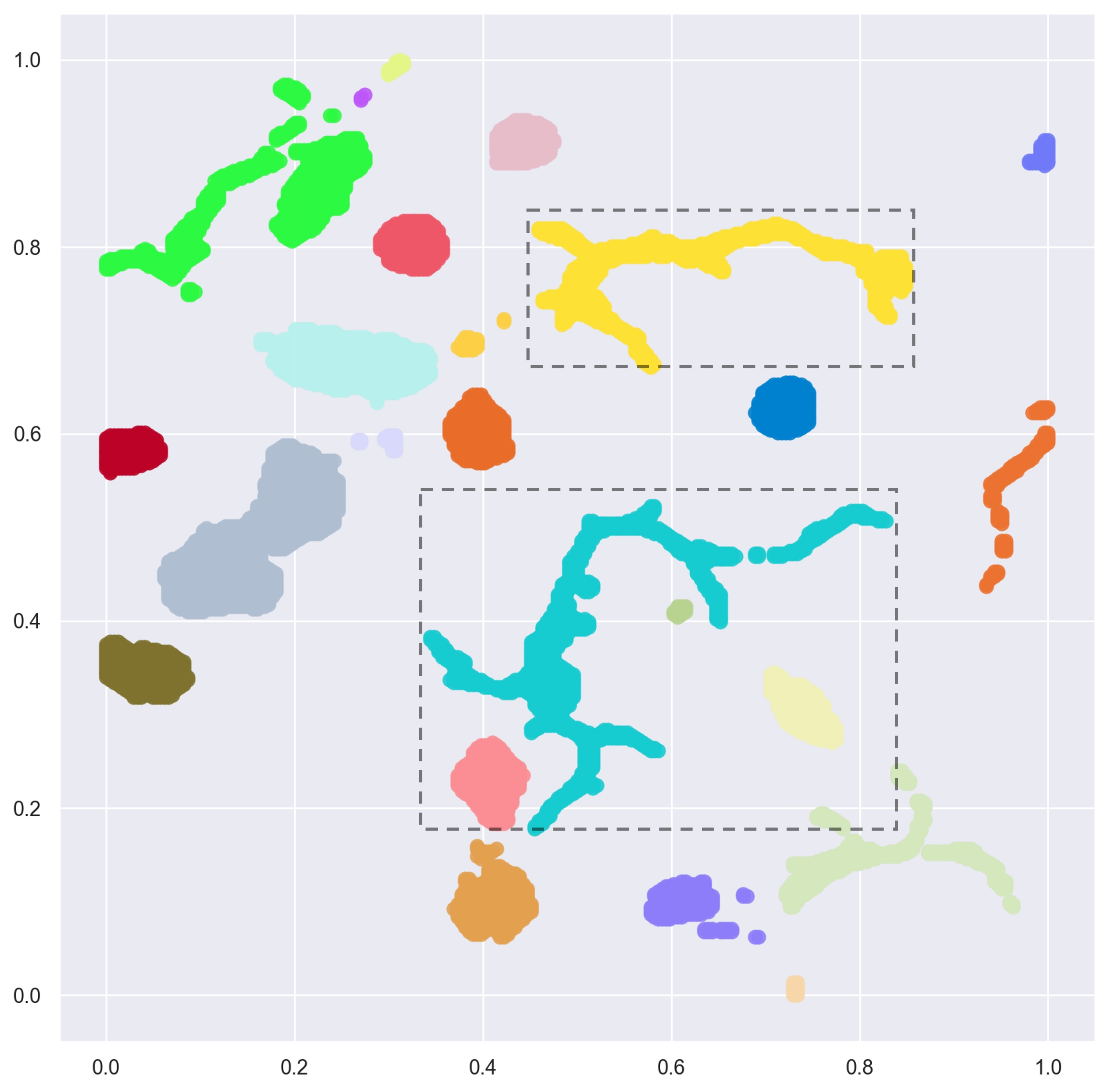}%
		\label{fig_9_8}}
	\caption{Cluster analysis of immune cells by confocal microscopy.  (a) The ground truth immune cells. (e)The performance of K-Means, DBSCAN, DP, Normal\_MST, LDP\_MST and GBMST. (b) The clustering result of the K-Means. (c) The clustering result of the DBSCAN. (d) The clustering result of the DP. (f)The clustering result of the Normal\_MST. (g) The clustering result of the LDP\_MST. (h) The clustering result of the GBMST.}
	\label{fig_9}
\end{figure*}
We apply our method to biomedical data segmentation, a challenging spatiotemporal biomedical dataset including 8681 instances\cite{Beltman2009Analysing,Pizzagalli2018Leukocyte}. This dataset contains both spherical and non-spherical shapes with potential contact\cite{Pizzagalli2019A}, and images of immune cells that exhibit high plasticity are particularly relevant.

In Fig.8, we show the results of an analysis performed for the immune cells. The ground truth of the immune cells is displayed in Fig.8a. The bar graph (Fig.8e) shows the Rand Index\cite{Steinley2004Properties} performance of K-Means, DBSCAN, DP, Normal\_MST, LDP\_MST and GBMST. This shows that GBMST achieves the best performance. In Fig.8, we depict the clusters with different colors. We can see that K-Means, DPeak and LDP\_MST split one cell into multiple parts, and disjoint cells are classified into one cluster in Fig.8b, Fig.8d and Fig.8g. As shown in Fig.8c, we find that DBSCAN can detect the upper cluster in the dotted box. In Fig.8f, Normal\_MST can also detect the upper cluster in the dotted box. However, multiple clusters are split into one cluster represented in green. From the Fig.8h, GBMST can detect immune cells well in the biomedical data segmentation. Only two clusters in proximity are spited into one cluster in the dotted box.

\subsection{Clustering on UCI Data sets}
We further validate the algorithm GBMST by conducting experiments on the UCI dataset. The details of the UCI data are shown in Table 5. The comparison of ACC and NMI scores\cite{chen2011Parallel} on clustering results of UCI are shown in Table 6.

\begin{table}[!h]
	\caption{UCI data sets.\label{table5}}
	\renewcommand\arraystretch{1.4}
	\centering
	\begin{tabular}{lcccc}
		\hline&Datasets & Instances  & Clusters & Dimensions \\
		\hline& Iris    & 150  & 3 & 4   \\
		& Wine    & 178  & 3 & 13   \\
		& Control    & 600  & 6 & 60      \\
		& Segment    & 2310  & 7 & 19   \\
		& letter    & 20000 & 26 & 16     \\ 
		\hline
	\end{tabular}
\end{table}

\begin{table}[!h]
	\caption{The Scores of Clustering Results on UCI Data Sets.\label{table6}}
	\renewcommand\arraystretch{1}
	\centering
	\resizebox{\linewidth}{!}{
		\begin{tabular}{lcccccccc}
			\hline  &Datasets    &      & K-means  & DBSCAN  & DP      & Normal\_MST  &LDP\_MST   &GBMST   \\
			\hline  & Iris       & ACC  & 0.875    & 0.658   & 0.907   & 0.532        & 0.973     & 0.973  \\
			        &            & NMI  & 0.732    & 0.751   & 0.806   & 0.621        & 0.901     & 0.901  \\
                    & Wine       & ACC  & 0.944    & 0.691   & 0.853   & 0.456        & 0.983     & 0.991  \\
                    &            & NMI  & 0.816    & 0.527   & 0.656   & 0.412        & 0.928     & 0.953  \\  
                    & Control    & ACC  & 0.582    & 0.312   & 0.567   & 0.332        & 0.678     & 0.723  \\
                    &            & NMI  & 0.709    & 0.122   & 0.757   & 0.084        & 0.700     & 0.715  \\     
                    & Segment    & ACC  & 0.476    & 0.525   & 0.907   & 0.512        & 0.780     & 0.801  \\
                    &            & NMI  & 0.456    & 0.601   & 0.806   & 0.522        & 0.817     & 0.792  \\
                    & letter     & ACC  & 0.375    & 0.528   & 0.907   & 0.391        & 0.546     & 0.532  \\
                    &            & NMI  & 0.563    & 0.588   & 0.806   & 0.488        & 0.699     & 0.682  \\            
           \hline
		\end{tabular}	
		
	}	
\end{table}

In general, their ACC values are lower than GBMST and LDP\_MST. For Iris and Wine, K-means and DP have high accuracy, but they do not perform well in several other more complex datasets. Since the Normal\_MST is sensitive to noise, the scores of clustering results are generally low. For the data sets of control and segment with complex shapes, the NMI value of DBSCAN is relatively high. Except for Segment data set, algorithm GBMST scores higher than other methods on all other data sets.
\subsection{Evaluation on Running Time}
In this section, we analyze the time of all the contrasting algorithms on the experimental dataset. The time complexity of each comparison algorithm is listed in Table 7. We take the average of 10 runs of the algorithm on each data set as the running time displayed in Table 8.
\begin{table}[!h]
	\caption{The Time Complexity.\label{table7}}
	\renewcommand\arraystretch{1.4}
	\centering
	\begin{tabular}{lccc}
		\hline     & K-means     & DBSCAN      & DP \\
		\hline     & $O(nkt)$    & $O(n^2)$,   & $O(n^2)$   \\	       		       		       
		\hline
		\hline     & Normal\_MST & LDP\_MST    & GBMST  \\
		\hline     & $O(nlogn)$  & $O(nlogn)$  & $O(nlogn)$   \\	       
		\hline
	\end{tabular}
\end{table}

\begin{table}[!h]
	\caption{The Running Time on All Data Sets (s)\label{table8}}
	\renewcommand\arraystretch{1}
	\centering
	\resizebox{\linewidth}{!}{
		\begin{tabular}{lccccccc}
			\hline  &Datasets    & K-means  & DBSCAN   & DP      & Normal\_MST  &LDP\_MST   &GBMST   \\
			\hline  & D1  & 0.023    & 0.31     & 2.92    & 0.31        & 0.28     & 0.27    \\
				    & D2  & 0.037    & 0.34     & 3.36    & 0.34        & 0.27     & 0.29    \\
				    & D3  & 0.033    & 0.35     & 0.37    & 0.35        & 0.21     & 0.19    \\
				    & D4  & 0.069    & 0.34     & 2.28    & 0.34        & 0.20     & 0.18    \\
				    & D5  & 0.038    & 0.35     & 3.76    & 0.33        & 0.41     & 0.31    \\
                    & D6  & 0.085    & 0.42     & 10.58   & 0.40        & 0.91     & 0.80		\\
                    & D7  & 0.021    & 0.34     & 3.77    & 0.31        & 0.42     & 0.35    \\
                    & D8  & 0.028    & 0.35     & 3.86    & 0.31        & 0.87     & 0.39    \\
                    & D9  & 0.043    & 0.38     & 9.53    & 0.32        & 1.13     & 0.81    \\
                    & D10 & 0.044    & 0.37     & 7.25    & 0.33        & 0.63     & 0.58    \\
                    & Cell       & 0.27     & 0.70     & 263.36  & 0.59        & 64.24    & 33.89    \\
                    & Iris       & 0.011    & 0.23     & 0.29    & 0.12        & 0.13     & 0.27    \\
                    & Wine       & 0.013    & 0.11     & 0.27    & 0.16        & 0.18     & 0.27    \\
                    & Control    & 0.023    & 0.21     & 0.45    & 0.31        & 0.11     & 0.27    \\	
                    & Segment    & 0.025    & 0.51     & 3.86    & 0.45        & 1.15     & 1.01    \\
                    & letter     & 0.18     & 77.82    & 369.88  & 0.31        & 78.23    & 50.18    \\
            \hline
		\end{tabular}	
		
	}	
\end{table}

As shown in Table 8, DP algorithm has the lowest time efficiency and K-means has the highest time efficiency. Compared with the DP and LDP\_MST algorithms with higher clustering accuracy, the time complexity of GBMST is better than both. When the data set is small, the time efficiency of DBSCAN algorithm and GBMST algorithm is not much different, but when the data set is large, GBMST is obviously better than DBSCAN. Overall, GBMST outperforms other algorithms in both accuracy and efficiency.
\section{Conclusion}
In this paper, we improve the granular-ball generation process to obtain suitable granularity for the dataset. The MST construction efficiency can be improved and noise interference can be reduced by multi-granularity granular-ball. Therefore, we propose a clustering algorithm that combines multi-granularity Granular-Ball and MST (GBMST). The clustering results show that GBMST can efficiently identify datasets containing complex shapes and noise. The GBMST only needs to input a cluster number K, and the generation of both granular-ball and MST is done adaptively.

There are still many directions for future research and exploration. Next, we will investigate how to improve the splitting method of granular-ball to improve the efficiency of GBMST. We will try to construct the MST with a larger-grained structure by taking the overlapping granular-balls as a whole to improve the accuracy and efficiency of the algorithm.
\section*{Acknowledgments}
This work was supported in part by the National Natural Science Foundation of China under Grant Nos. 62176033 and 61936001.

\bibliographystyle{IEEEtran}
\bibliography{myref}{}

\newpage
\vspace{11pt}
\vspace{-33pt}

\vfill

\end{document}


\title{A Multi-Granularity Hyper-Ball Clustering Method Based on Minimum Spanning Tree}
	\author{Jiang~Xie, Tian~Xia, Shuyin~Xia*, Guoyin~Wang, Qizhu~Dai} 	

	\begin{figure*}[!h]
	  \centering
	  \subfloat[]{\includegraphics[width=1.65in]{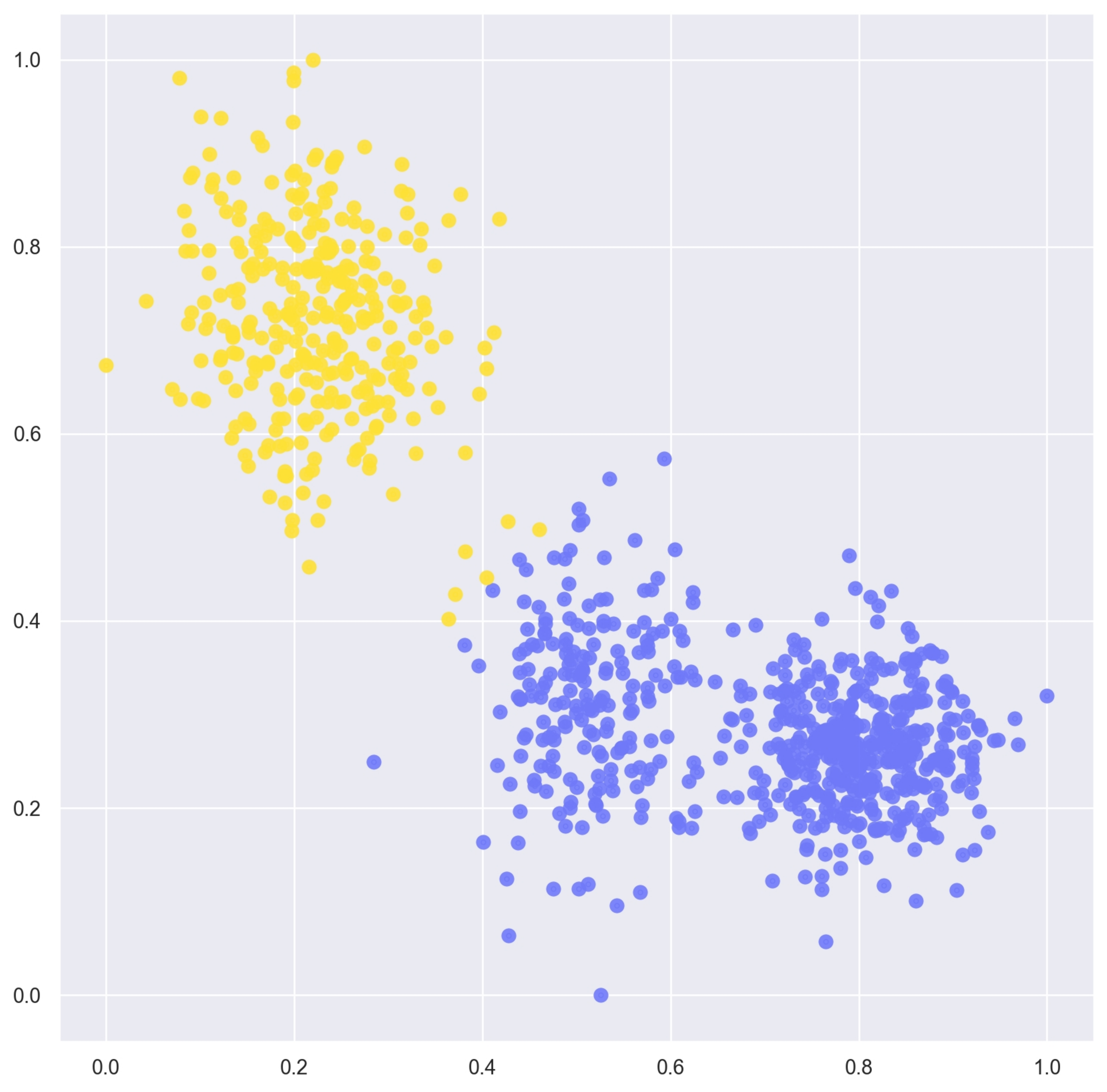}%
		\label{fig_1_1}}
	  \subfloat[]{\includegraphics[width=1.65in]{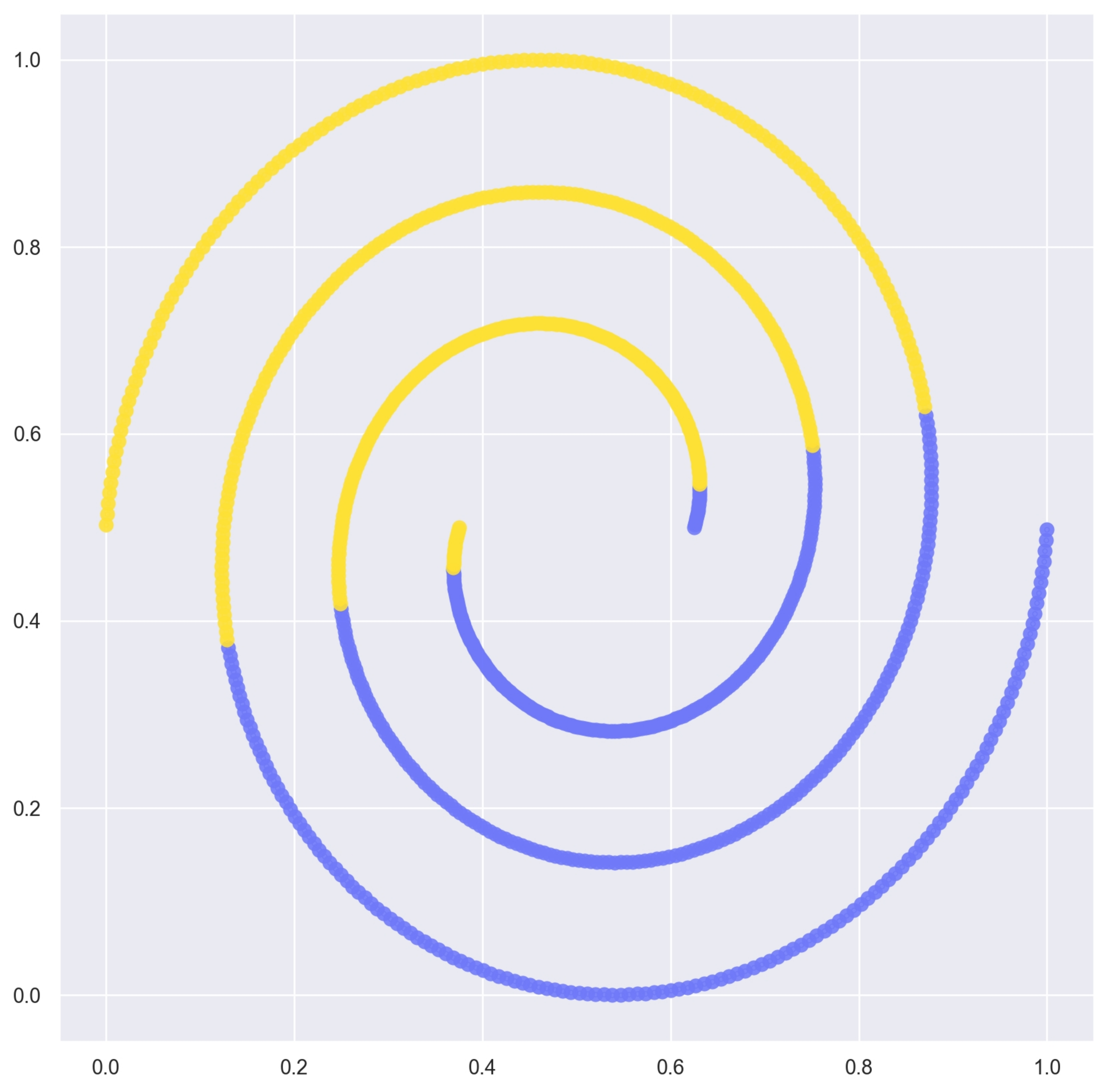}%
		\label{fig_1_2}}
	  \subfloat[]{\includegraphics[width=1.65in]{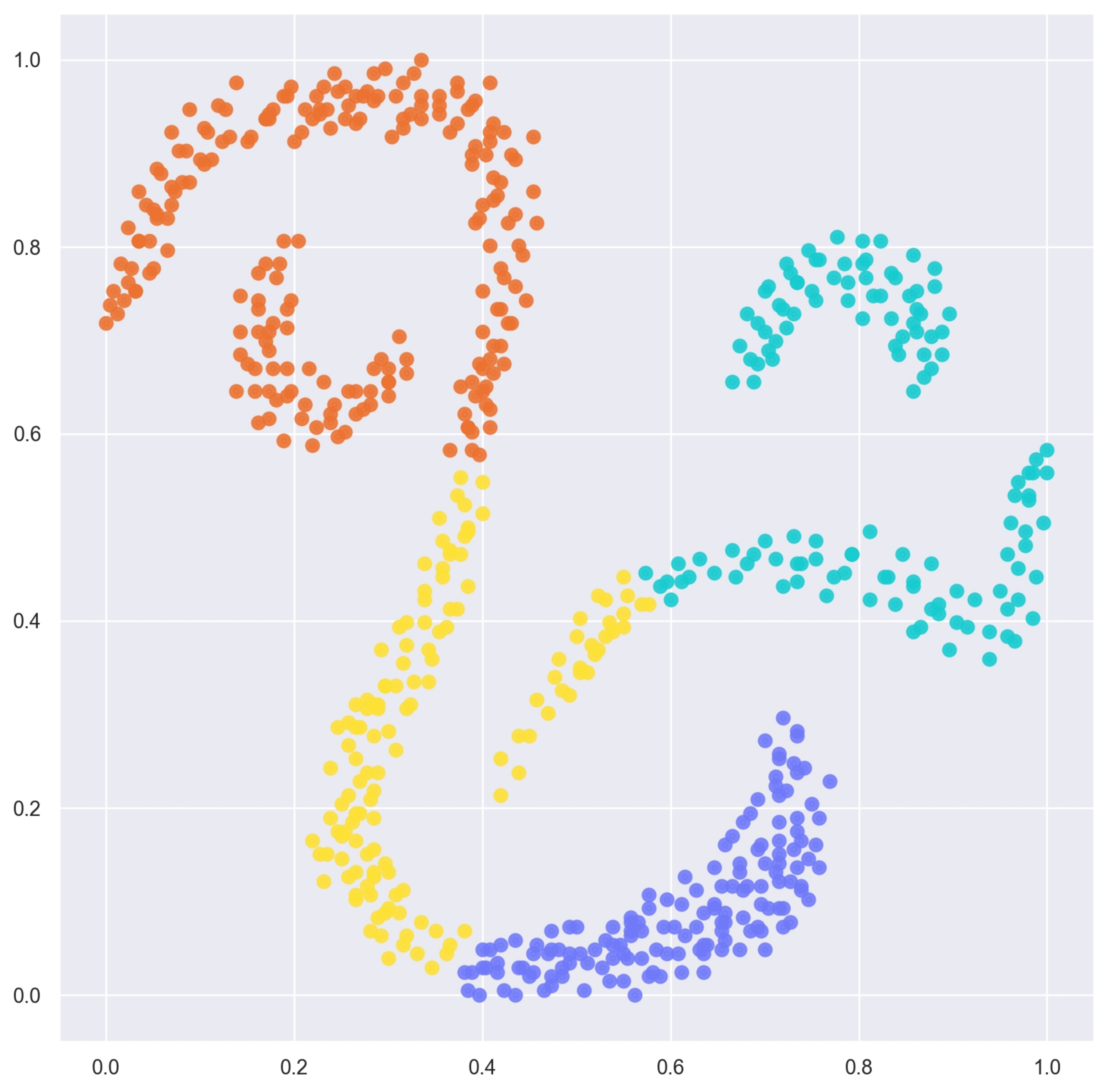}%
		\label{fig_1_3}}
	
	  \subfloat[]{\includegraphics[width=1.65in]{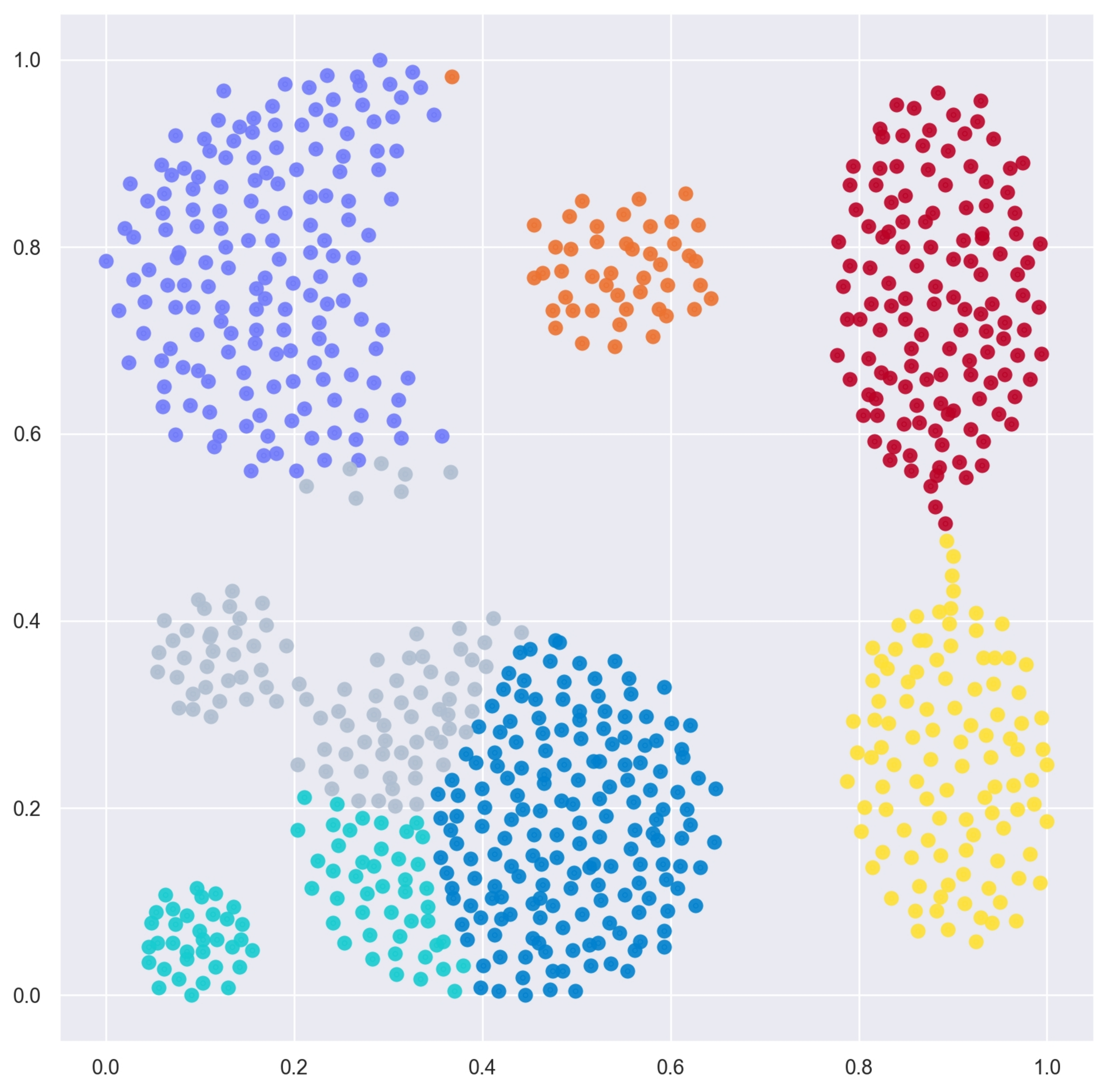}%
		\label{fig_1_4}}
	  \subfloat[]{\includegraphics[width=1.65in]{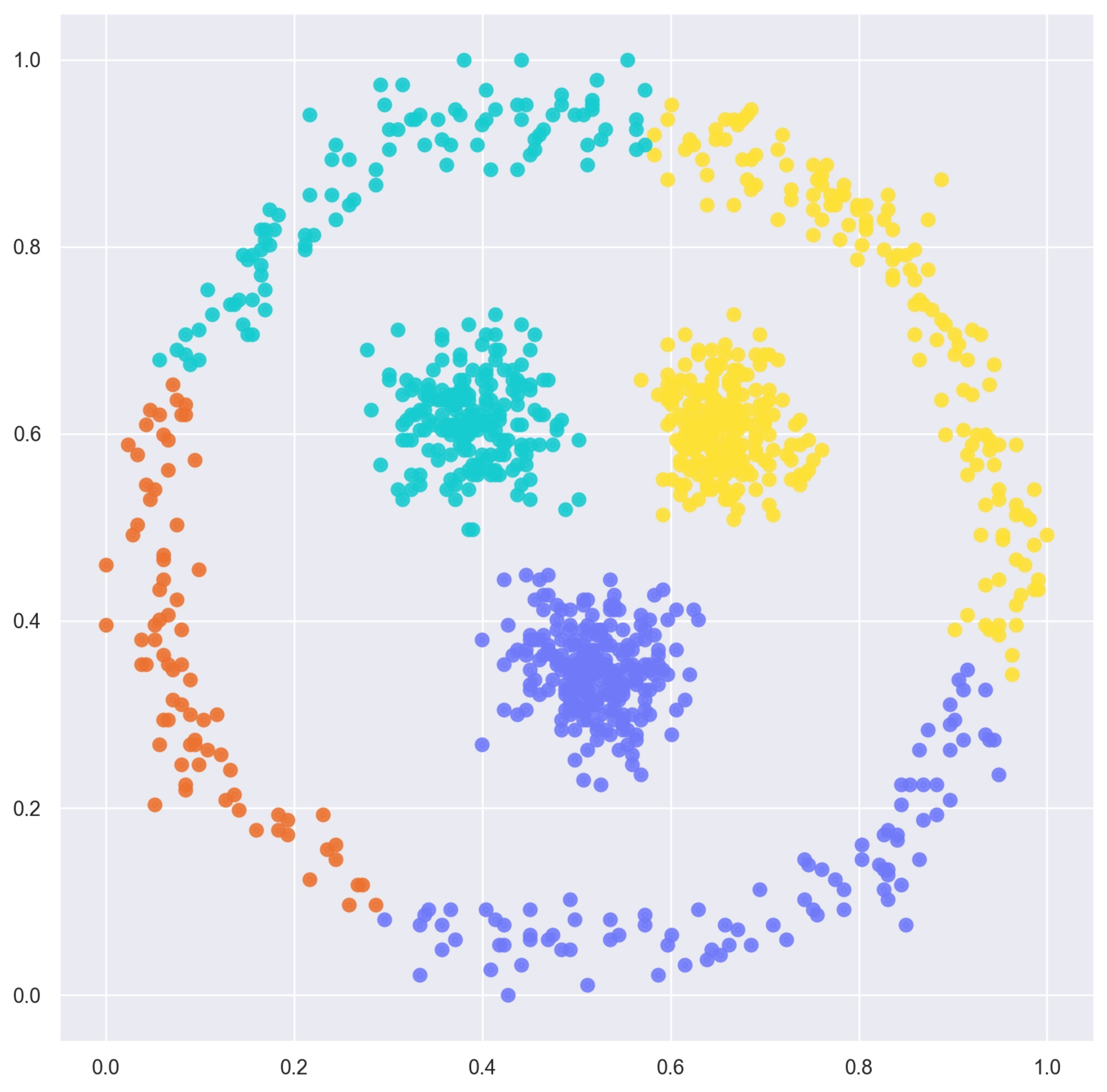}%
		\label{fig_1_5}}
	  \subfloat[]{\includegraphics[width=1.65in]{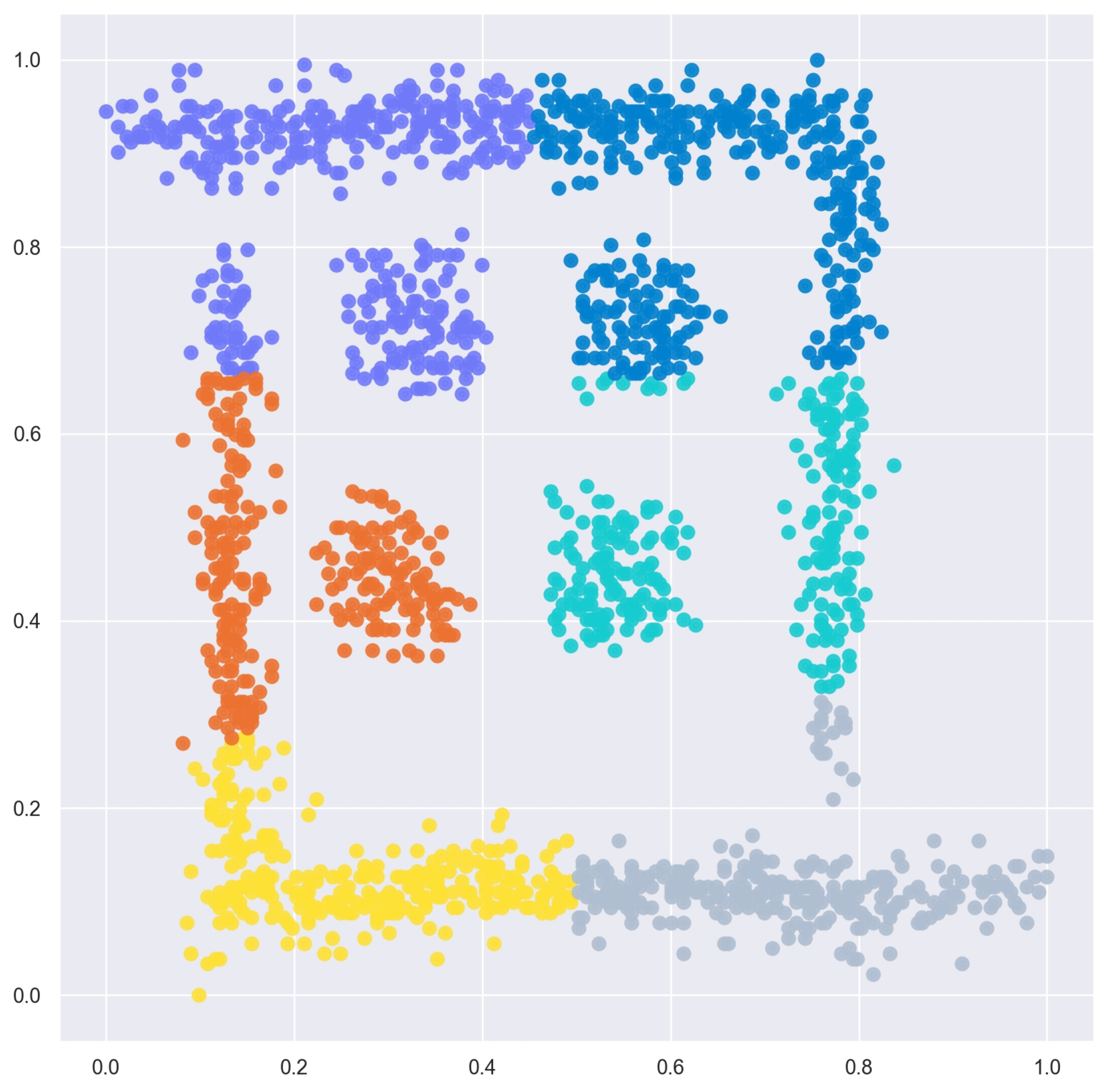}%
		\label{fig_1_6}}
	  \caption{The clustering results of K-means on synthetic data set without noise.}
	  \label{fig_1}
    \end{figure*}
	\begin{figure*}[!h]
	\centering
	\subfloat[]{\includegraphics[width=1.65in]{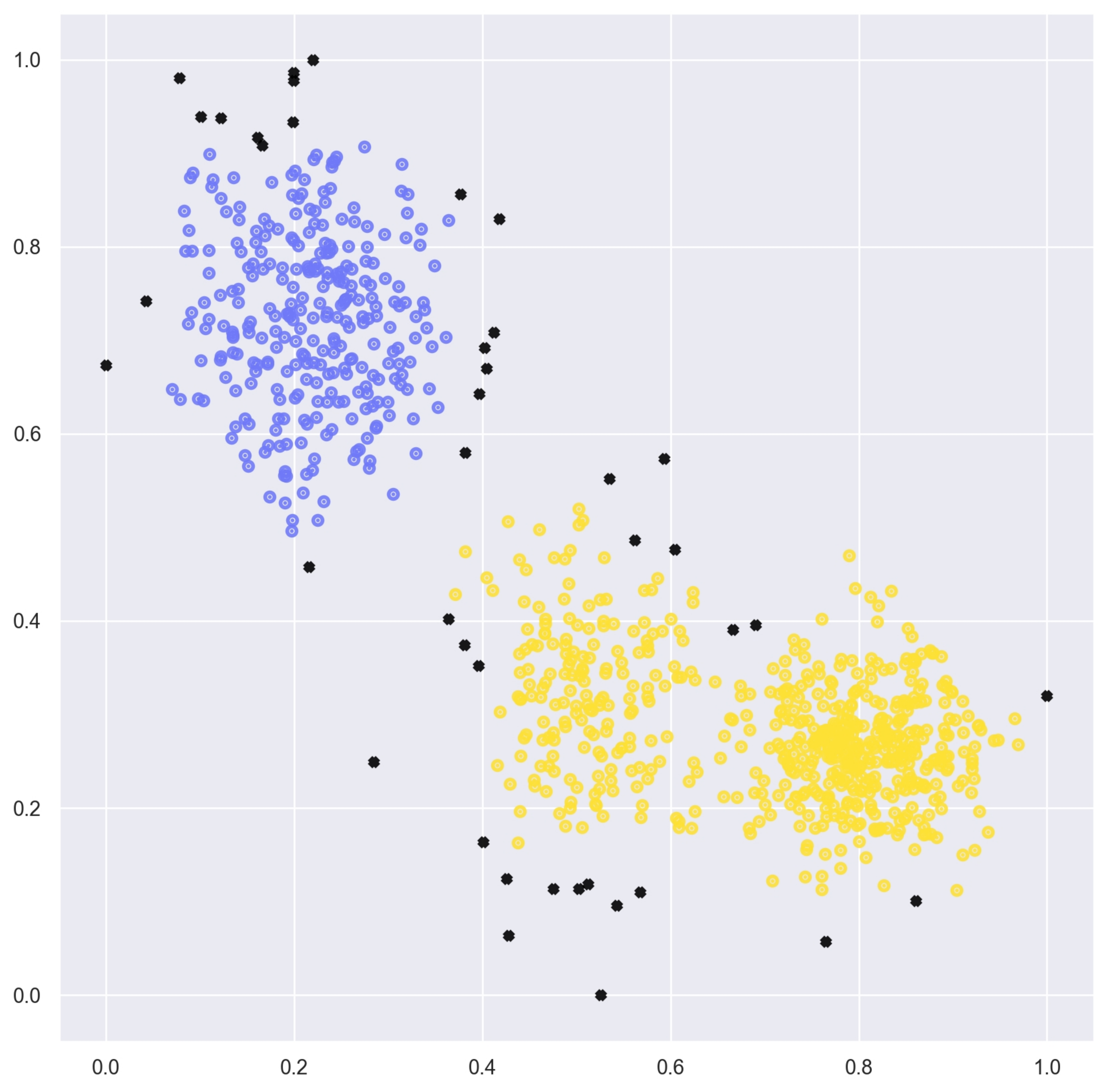}%
		\label{fig_2_1}}
	\subfloat[]{\includegraphics[width=1.65in]{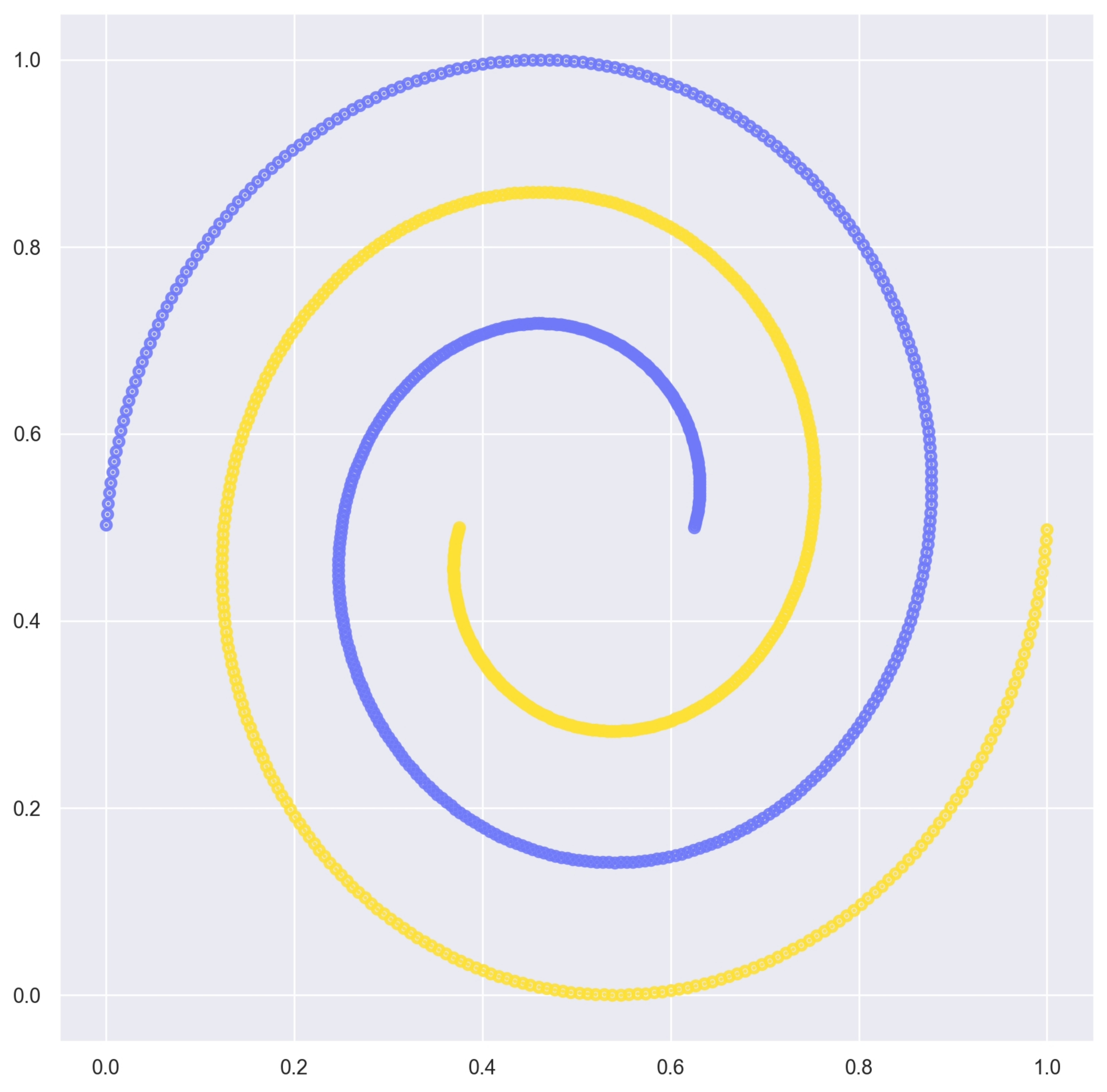}%
		\label{fig_2_2}}
	\subfloat[]{\includegraphics[width=1.65in]{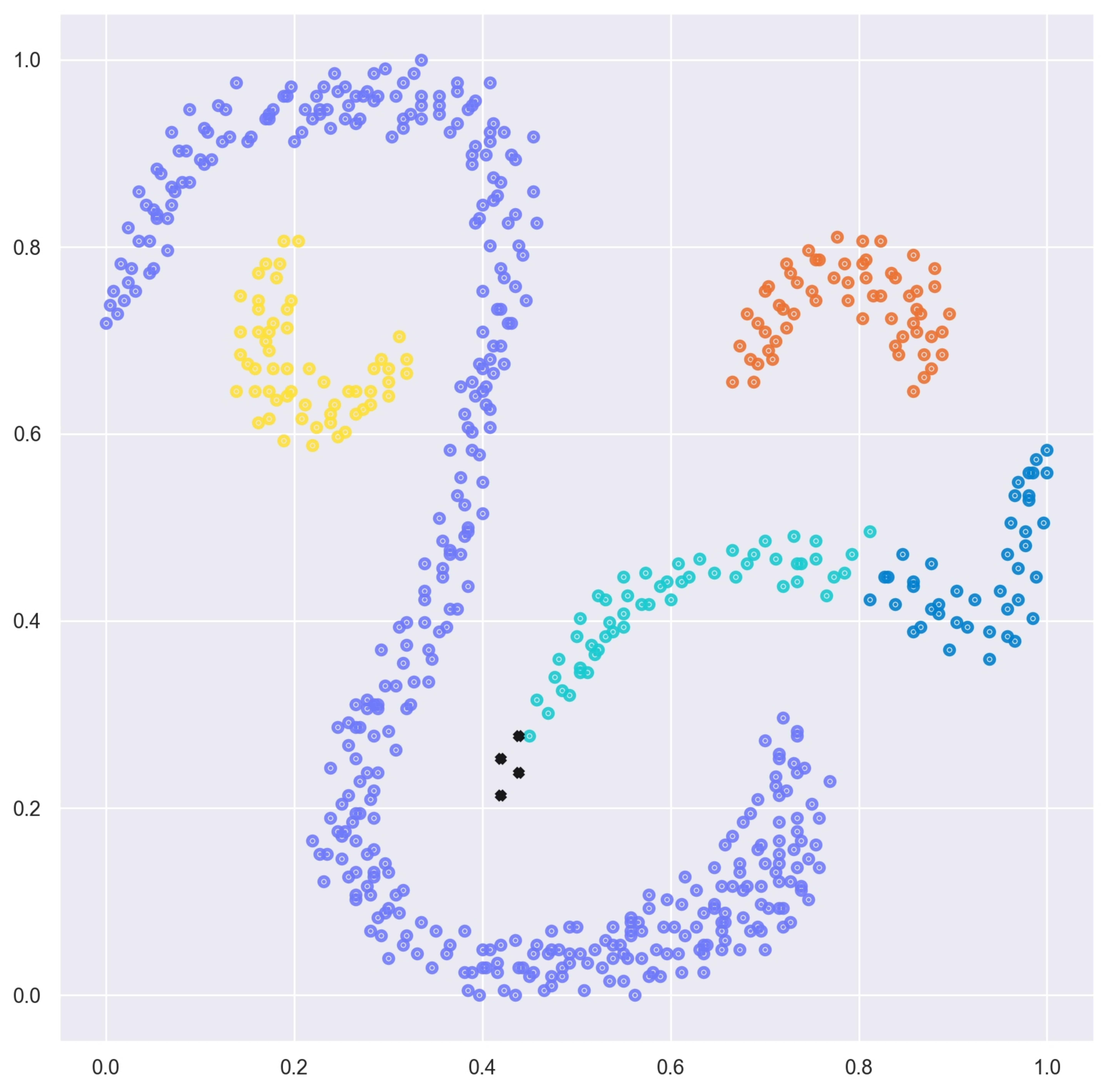}%
		\label{fig_2_3}}
	
	\subfloat[]{\includegraphics[width=1.65in]{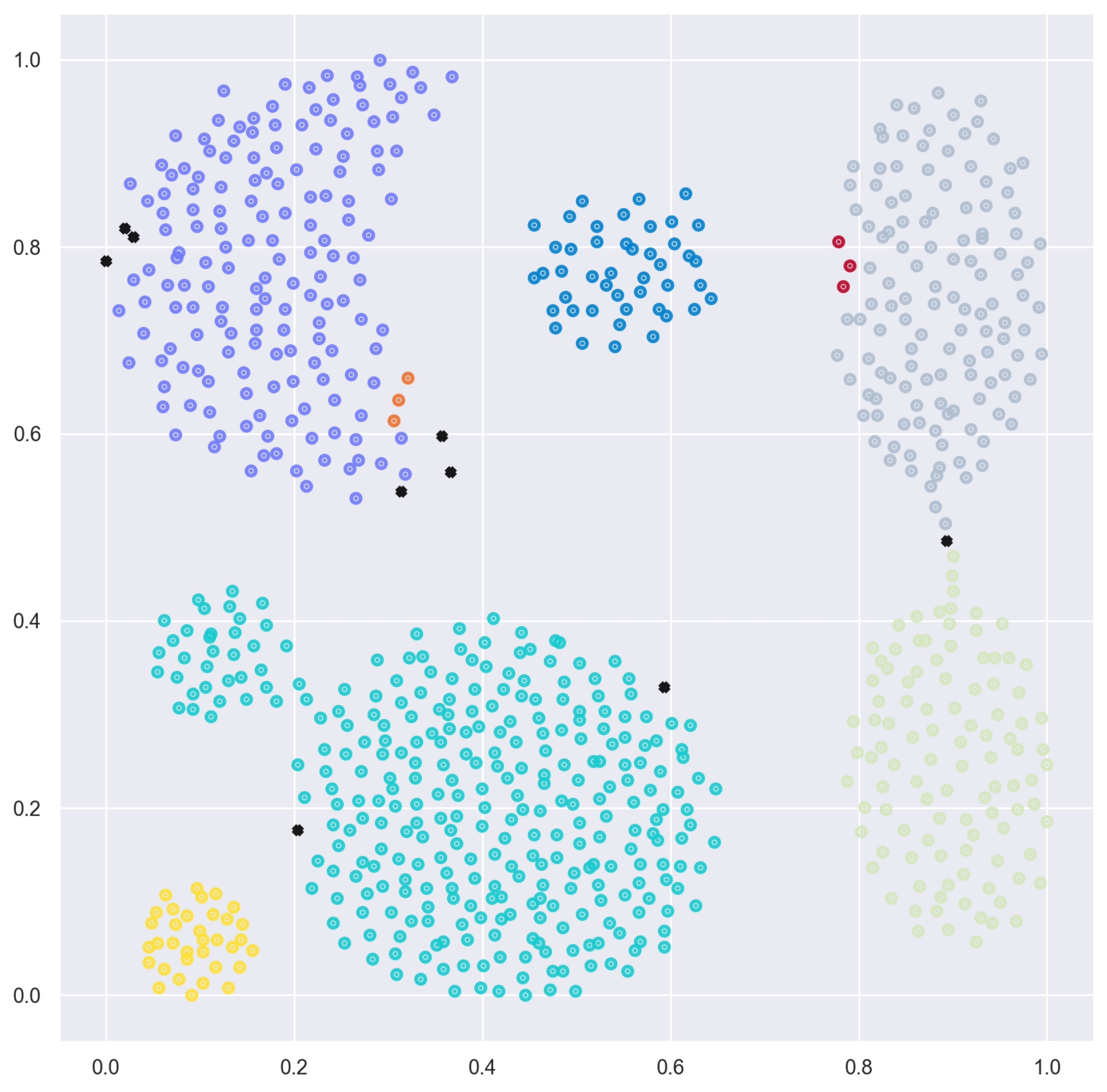}%
		\label{fig_2_4}}
	\subfloat[]{\includegraphics[width=1.65in]{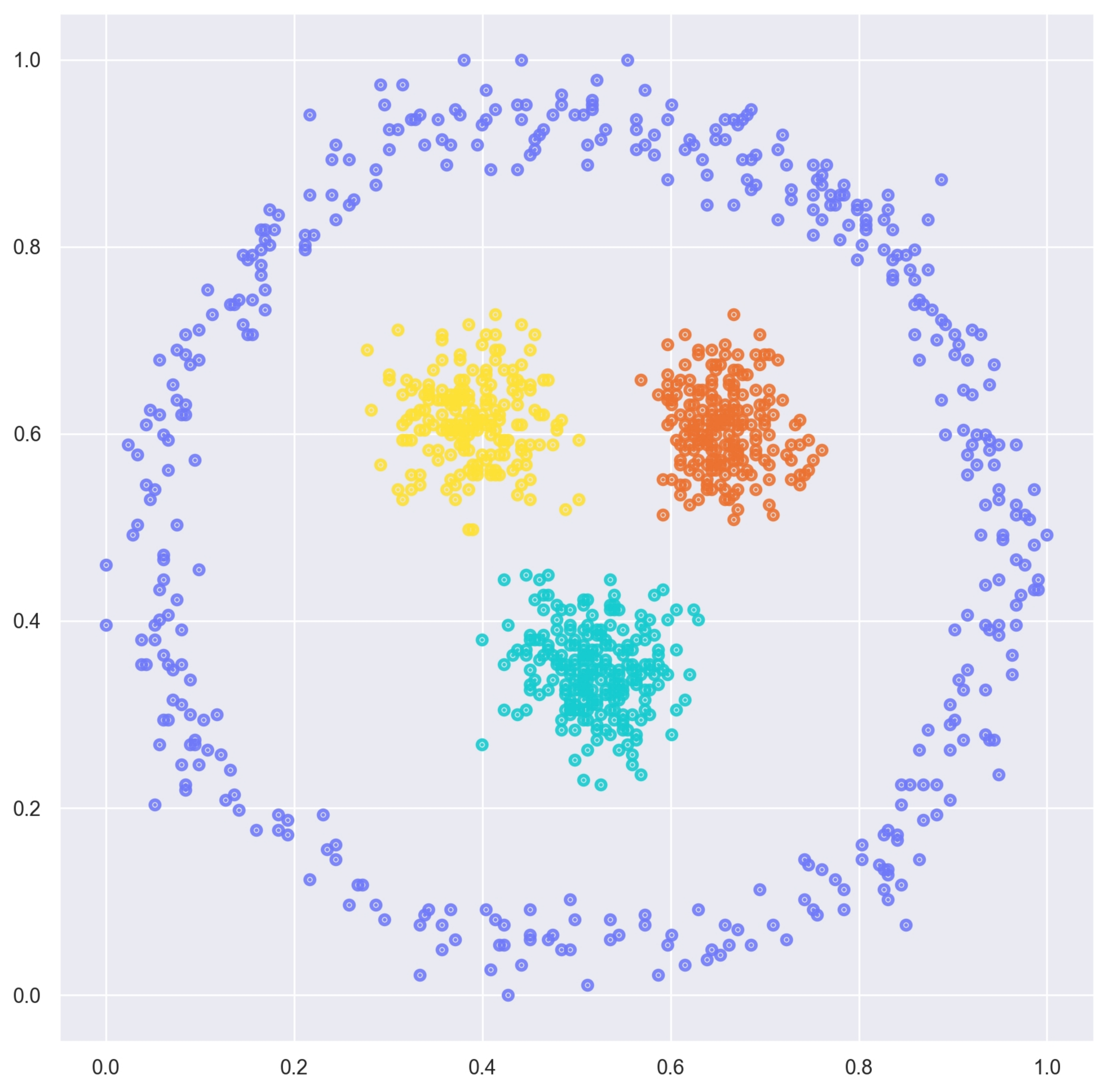}%
		\label{fig_2_5}}
	\subfloat[]{\includegraphics[width=1.65in]{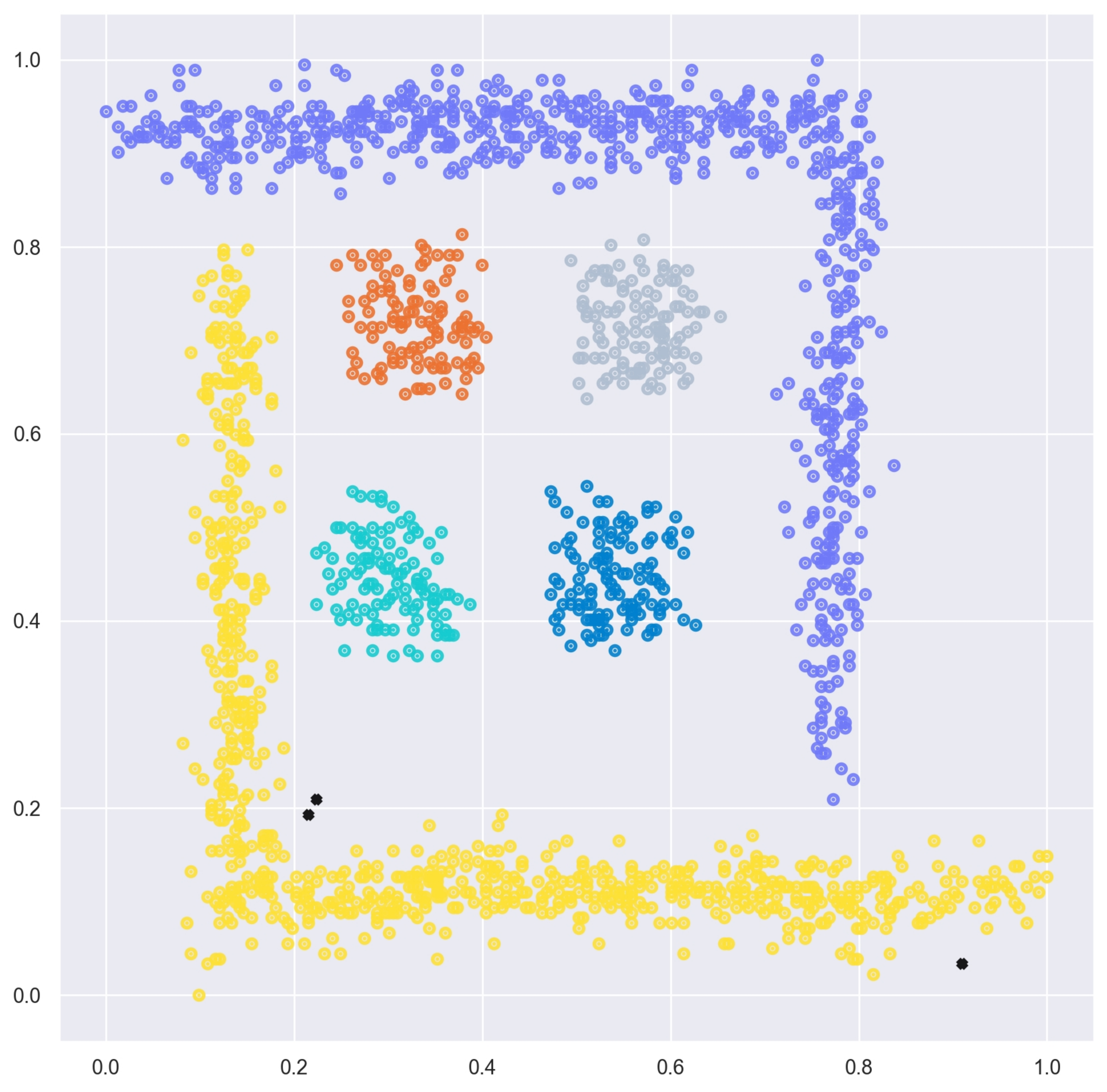}%
		\label{fig_2_6}}
	\caption{The clustering results of DBSCAN on synthetic data set without noise.}
	\label{fig_2}
    \end{figure*}
	\begin{figure*}[!h]
	\centering
	\subfloat[]{\includegraphics[width=1.65in]{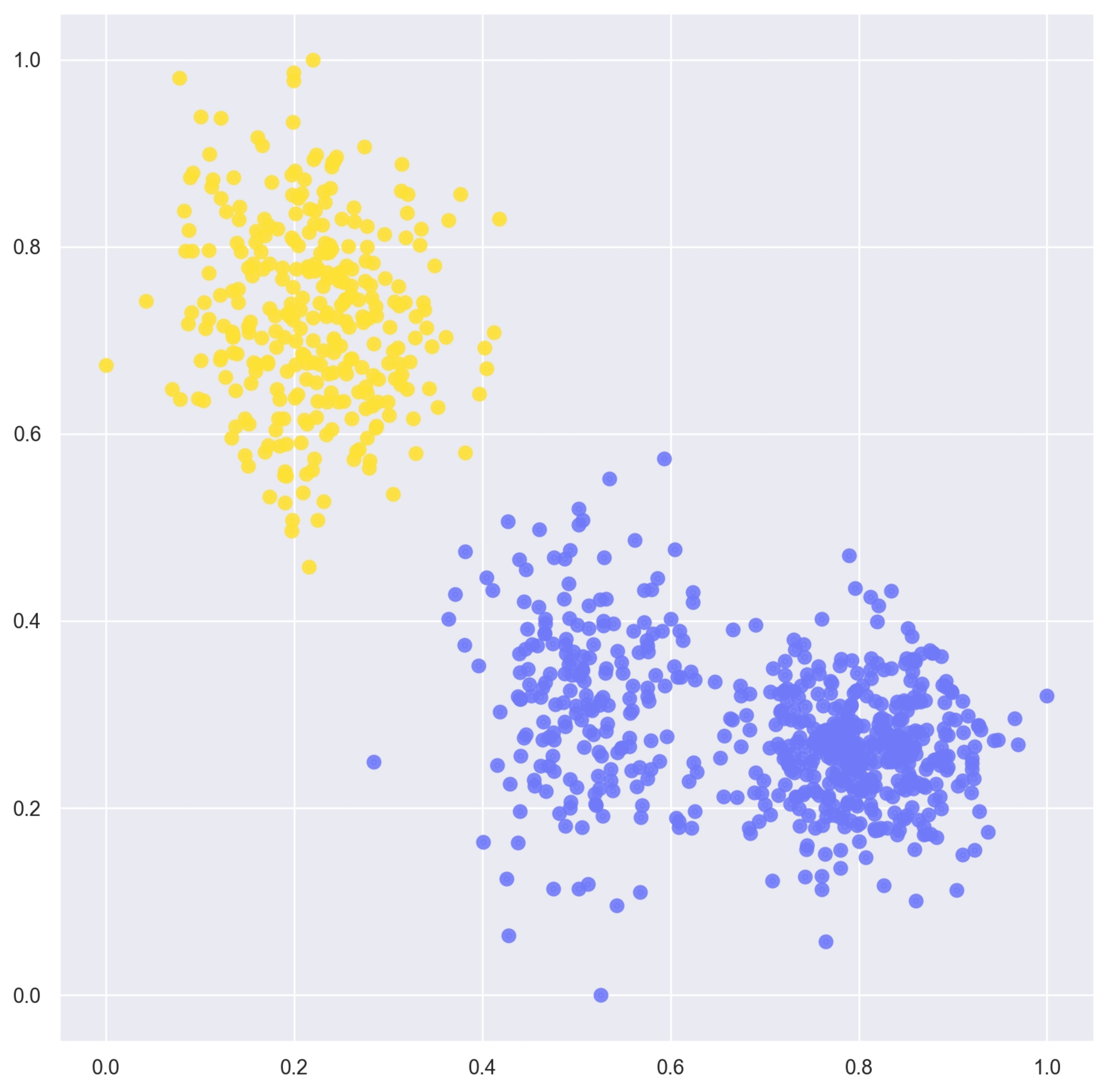}%
		\label{fig_3_1}}
	\subfloat[]{\includegraphics[width=1.65in]{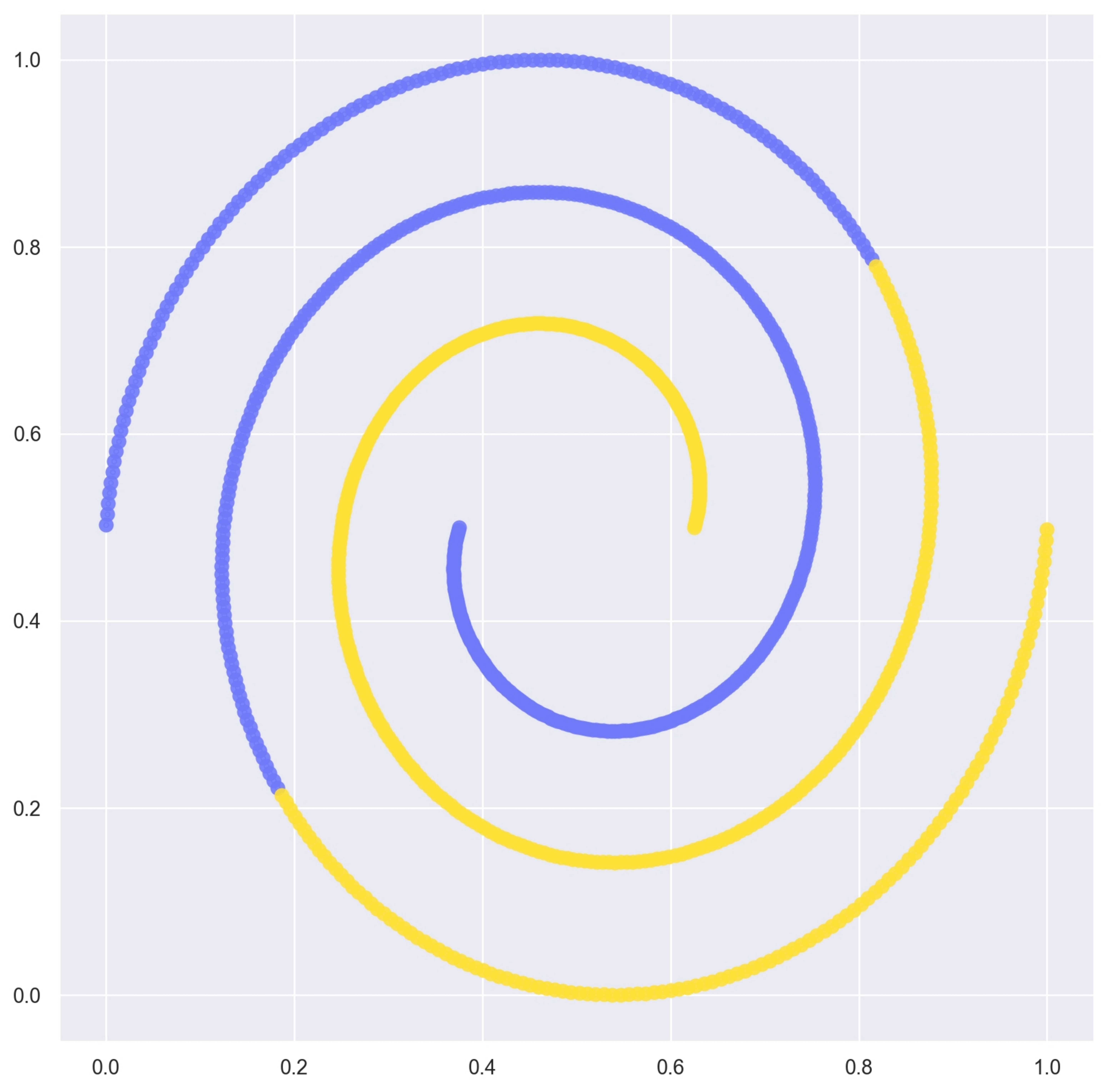}%
		\label{fig_3_2}}
	\subfloat[]{\includegraphics[width=1.65in]{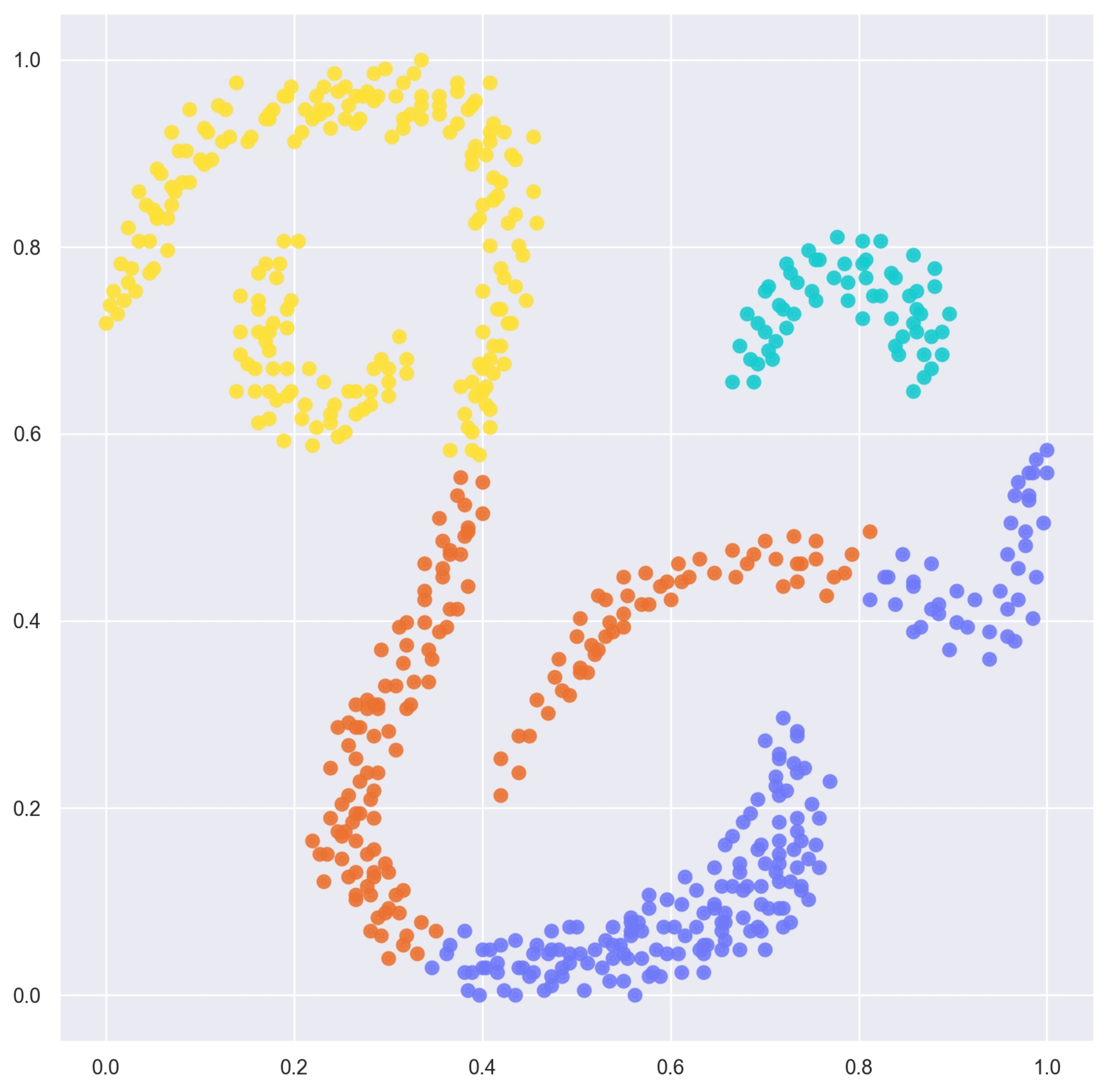}%
		\label{fig_3_3}}
	
	\subfloat[]{\includegraphics[width=1.65in]{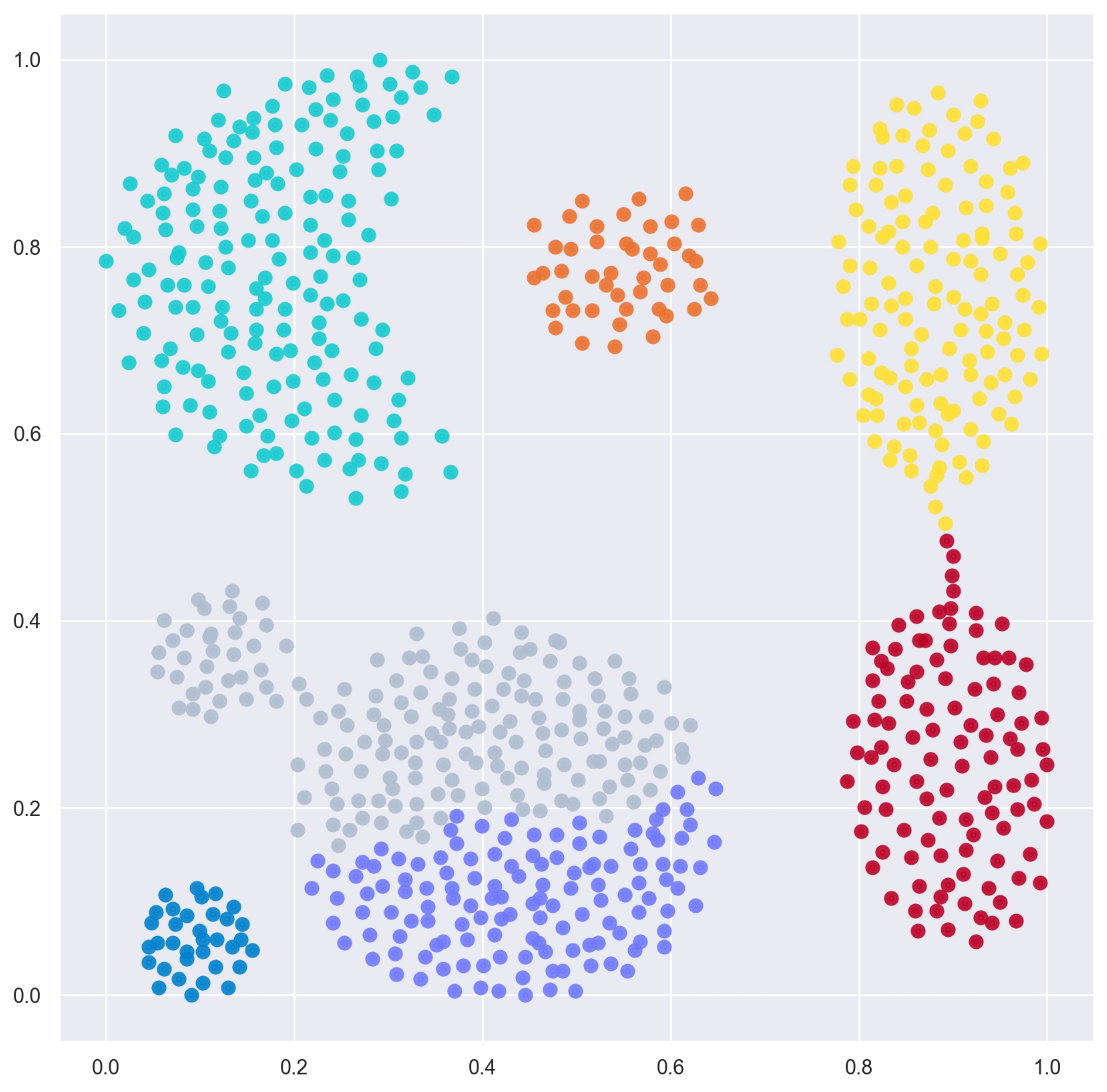}%
		\label{fig_3_4}}
	\subfloat[]{\includegraphics[width=1.65in]{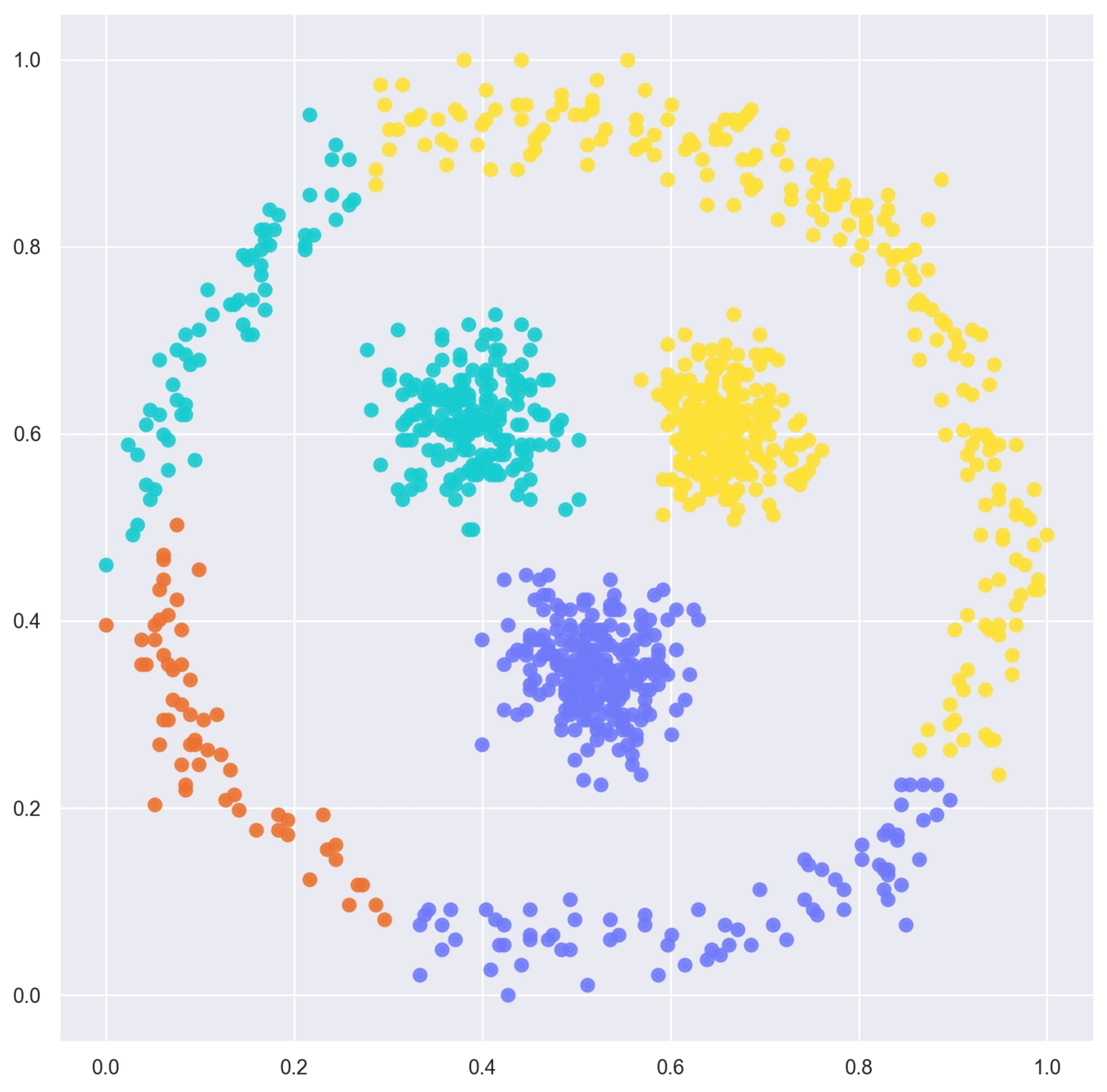}%
		\label{fig_3_5}}
	\subfloat[]{\includegraphics[width=1.65in]{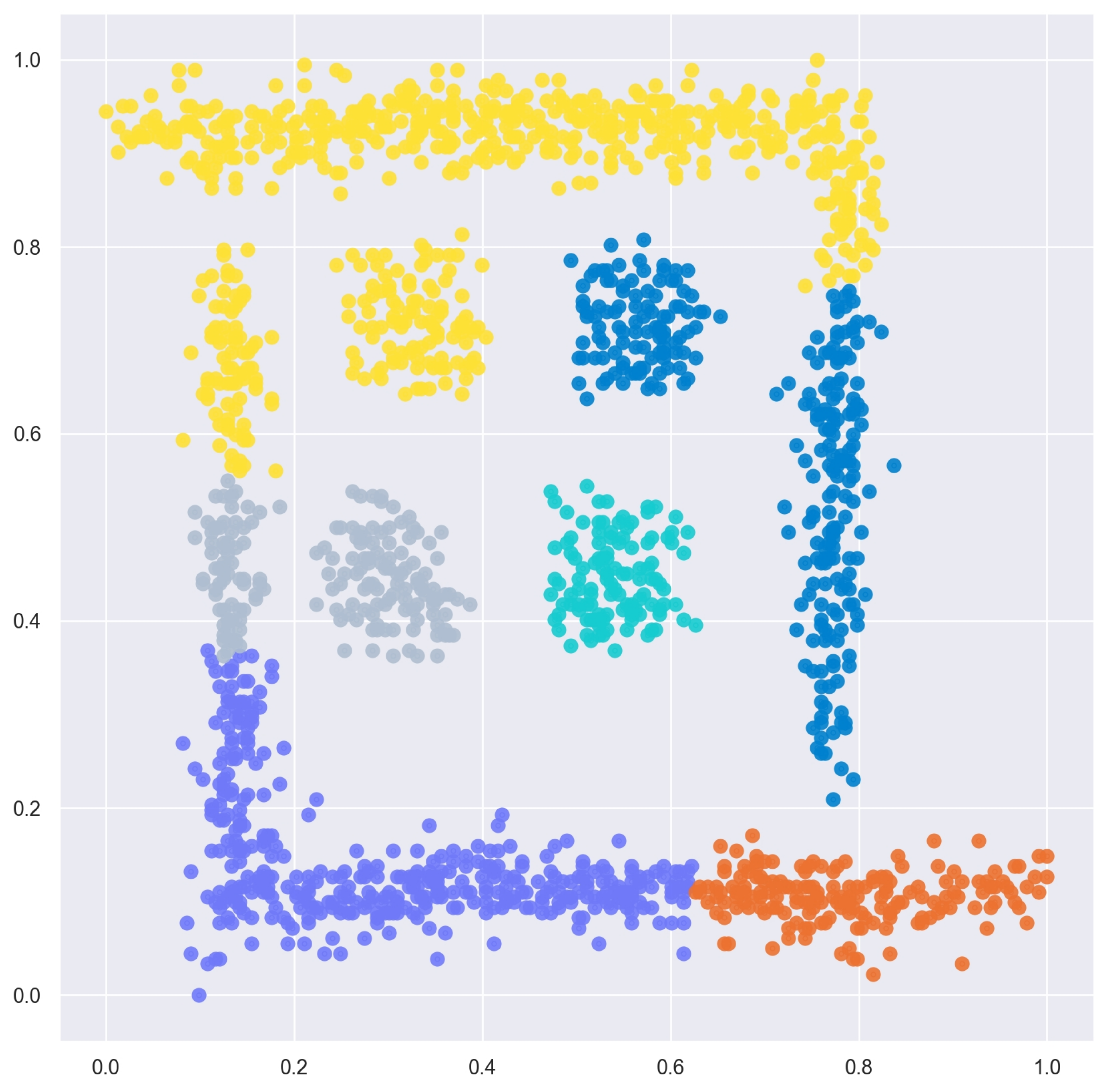}%
		\label{fig_3_6}}
	\caption{The clustering results of DP on synthetic data set without noise.}
	\label{fig_3}
    \end{figure*}
	\begin{figure*}[!h]
	\centering
	\subfloat[]{\includegraphics[width=1.65in]{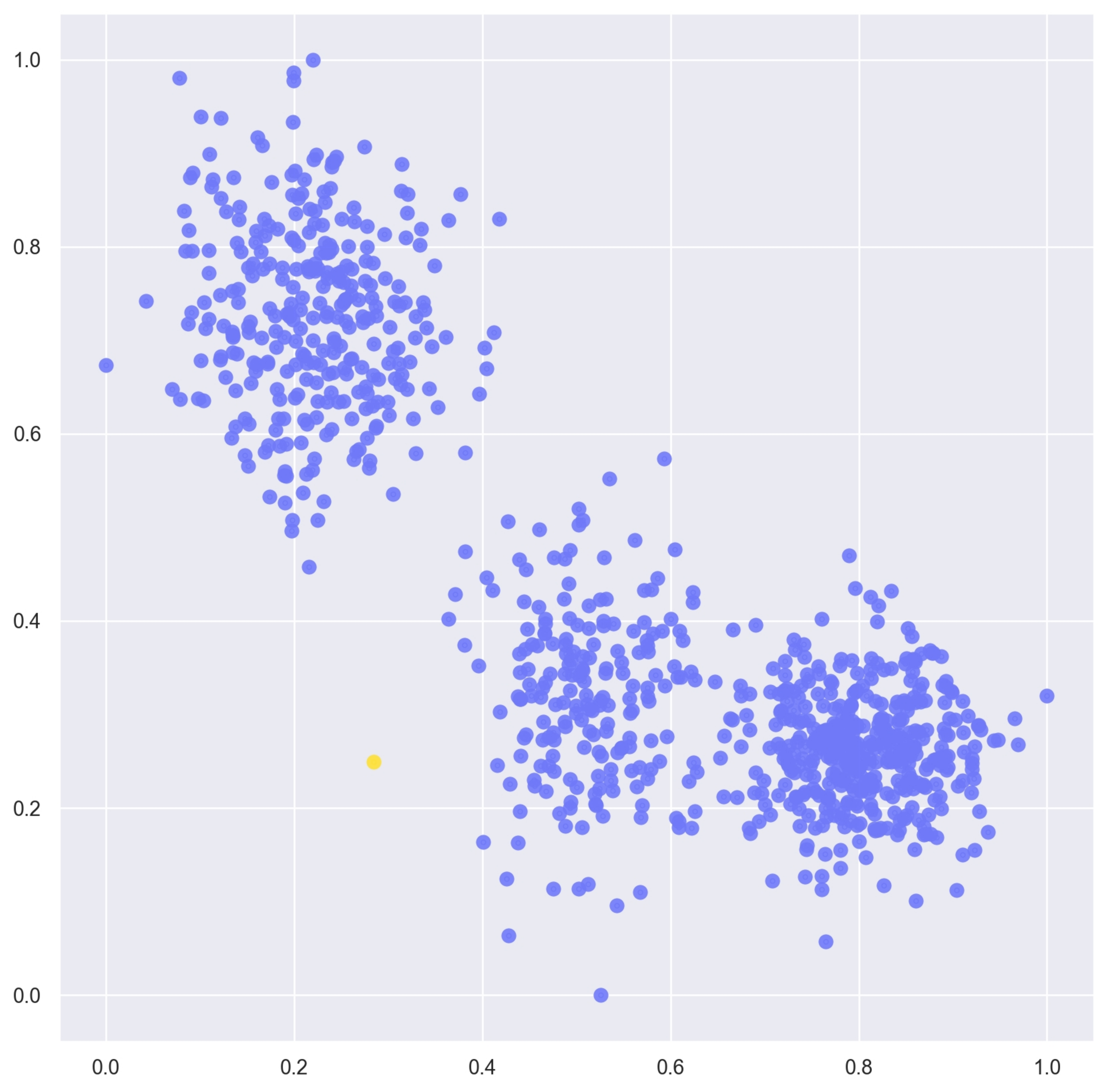}%
		\label{fig_4_1}}
	\subfloat[]{\includegraphics[width=1.65in]{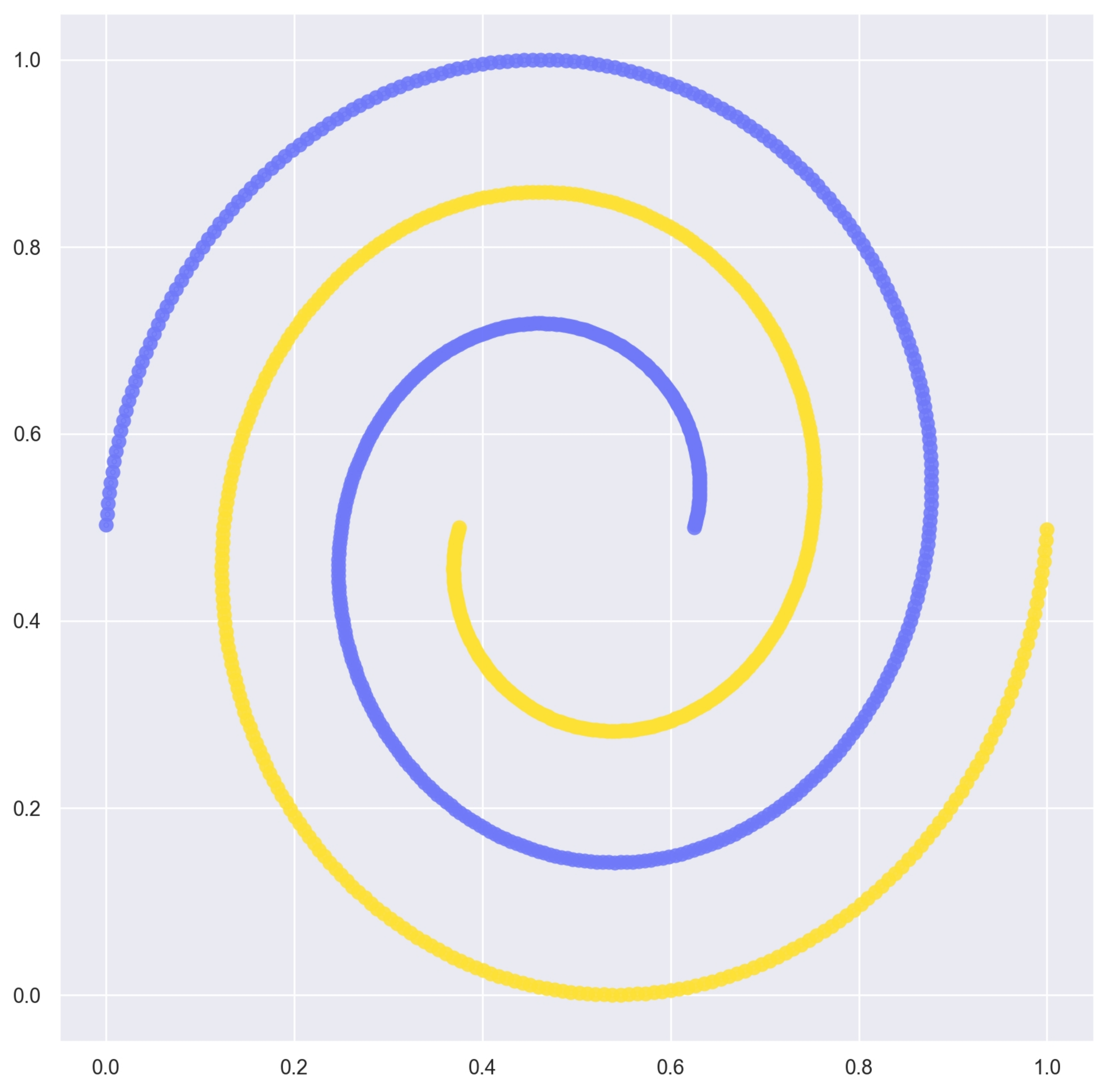}%
		\label{fig_4_2}}
	\subfloat[]{\includegraphics[width=1.65in]{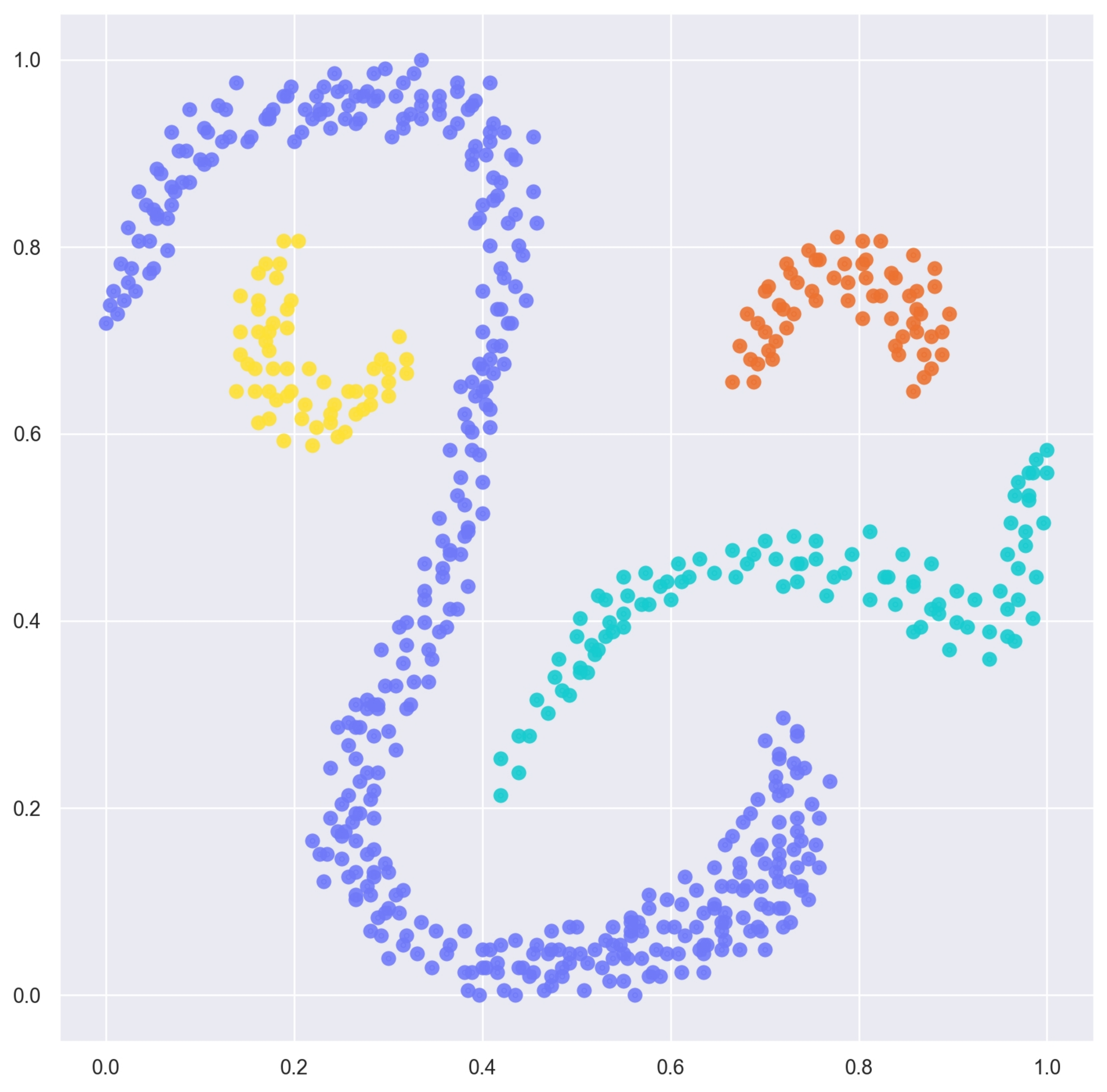}%
		\label{fig_4_3}}
	
	\subfloat[]{\includegraphics[width=1.65in]{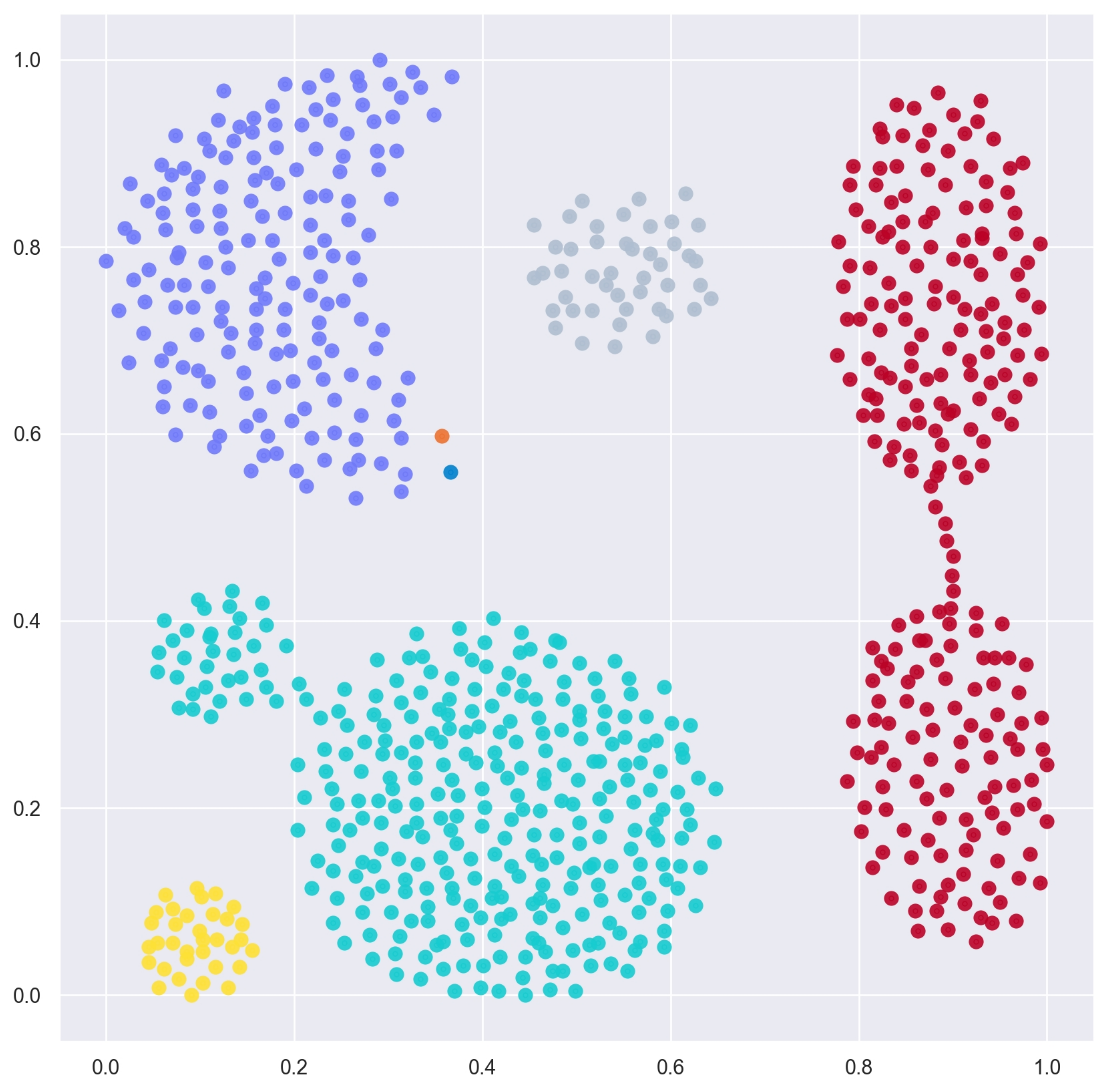}%
		\label{fig_4_4}}
	\subfloat[]{\includegraphics[width=1.65in]{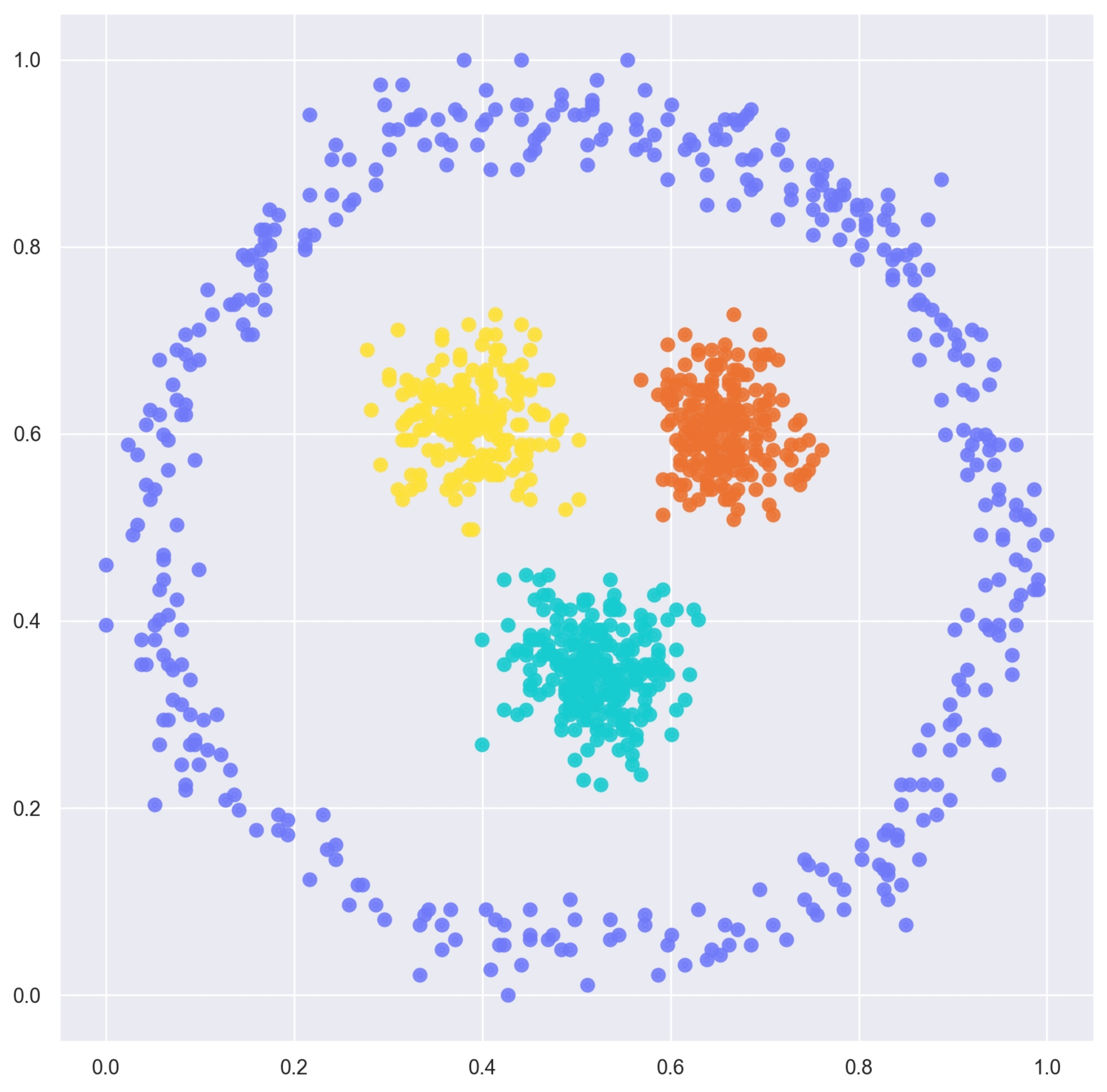}%
		\label{fig_4_5}}
	\subfloat[]{\includegraphics[width=1.65in]{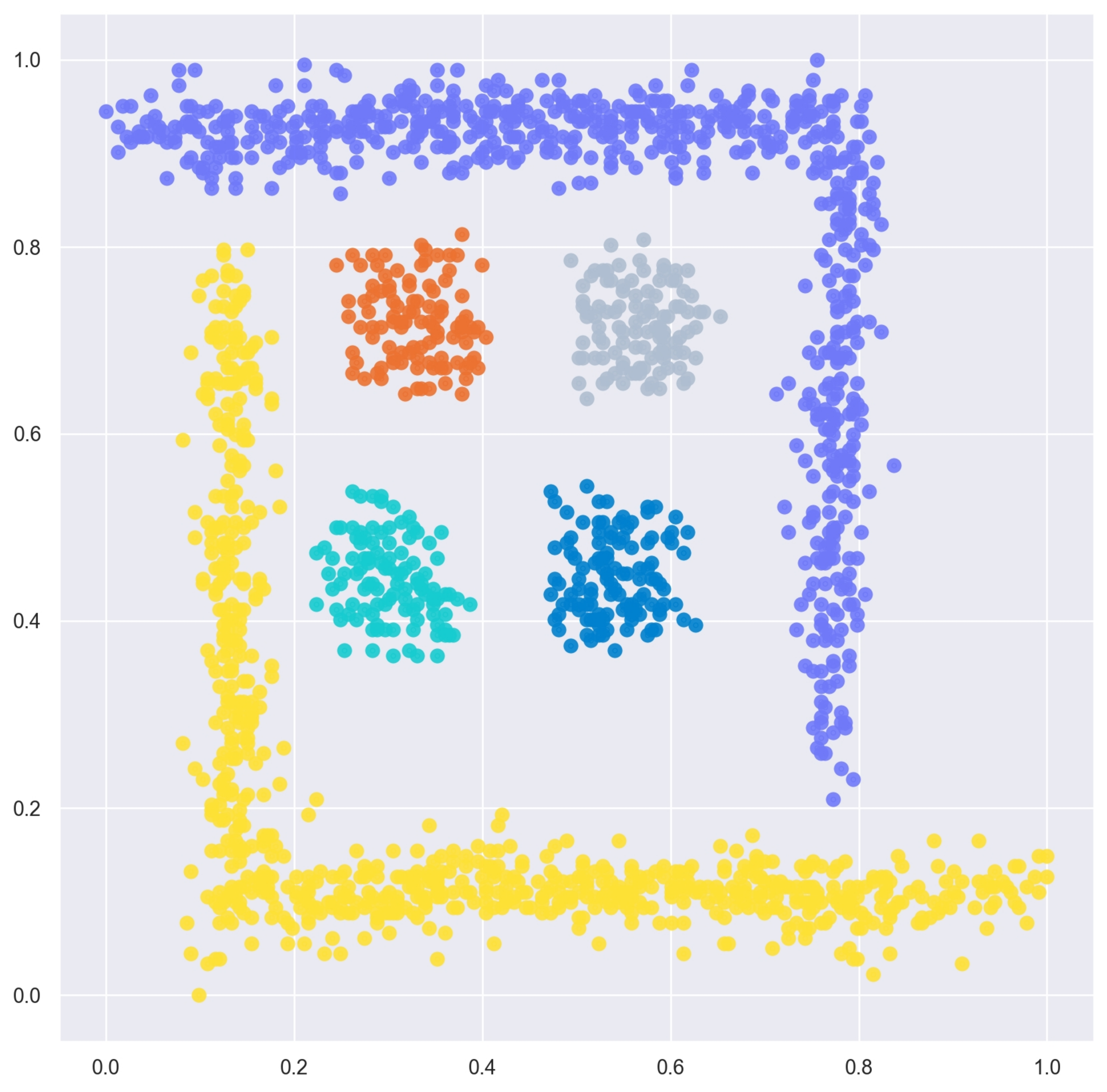}%
		\label{fig__6}}
	\caption{The clustering results of Normal\_MST on synthetic data set without noise.}
	\label{fig_4}
    \end{figure*}
	\begin{figure*}[!h]
	\centering
	\subfloat[]{\includegraphics[width=1.65in]{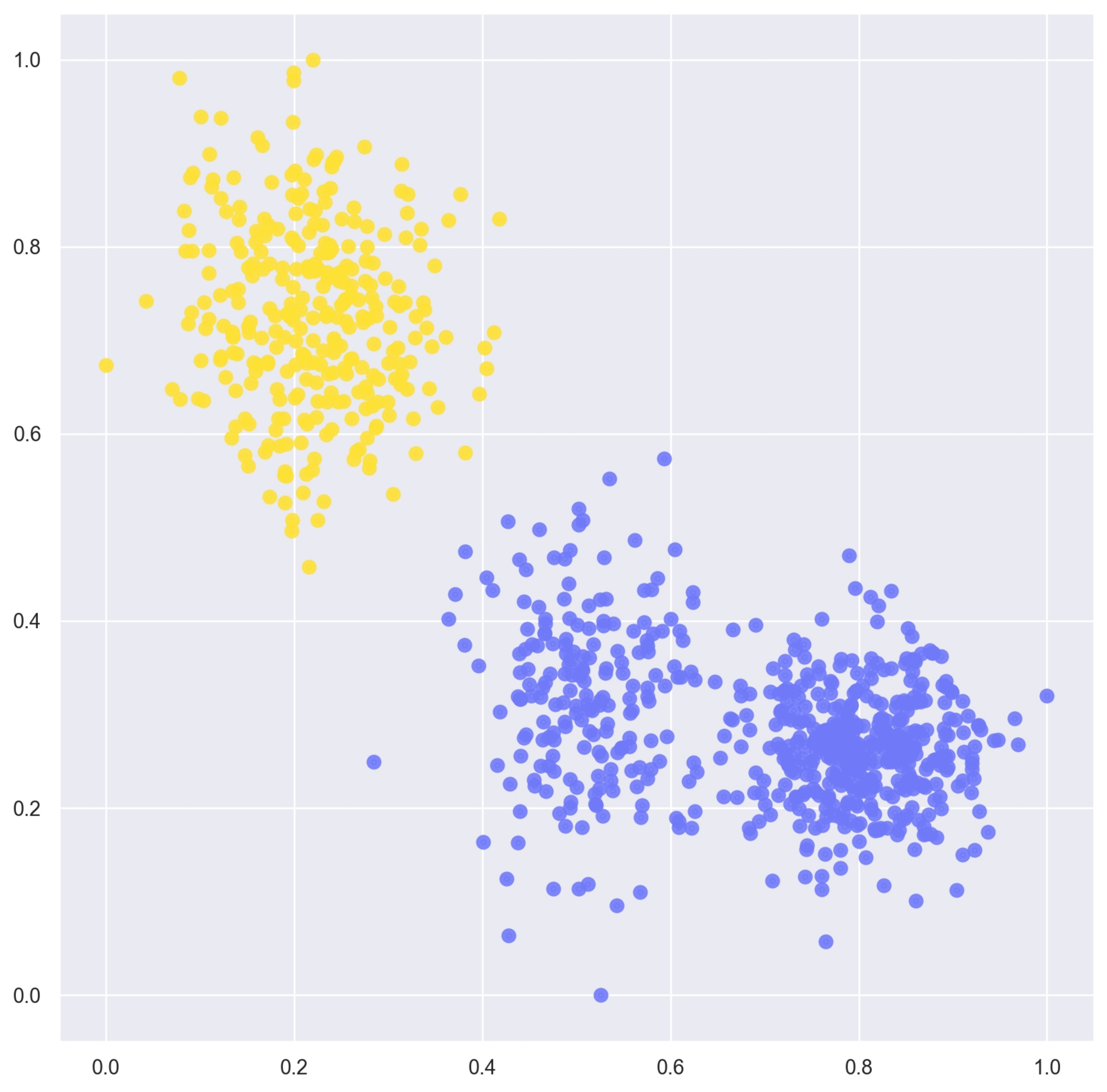}%
		\label{fig_5_1}}
	\subfloat[]{\includegraphics[width=1.65in]{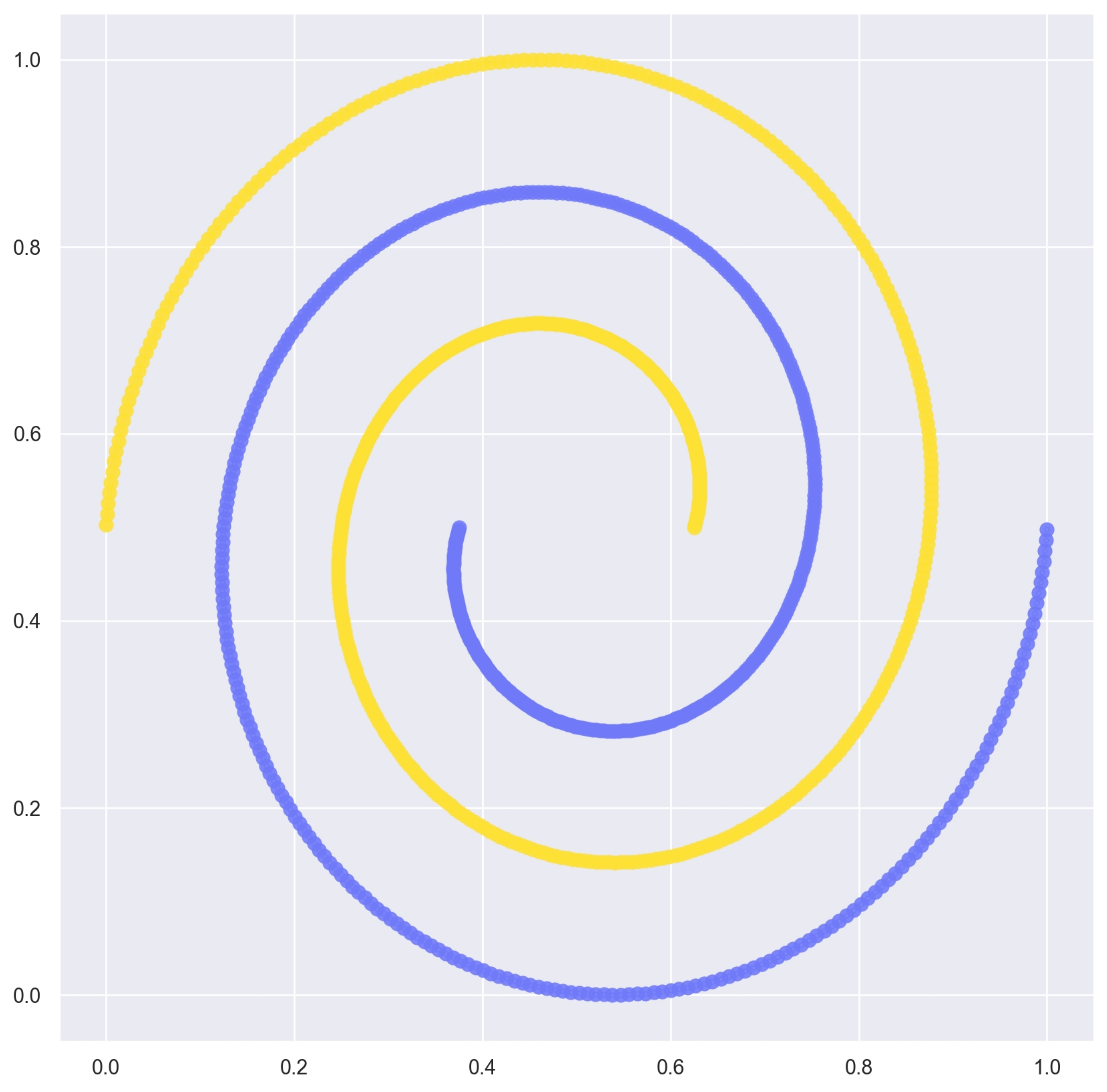}%
		\label{fig_5_2}}
	\subfloat[]{\includegraphics[width=1.65in]{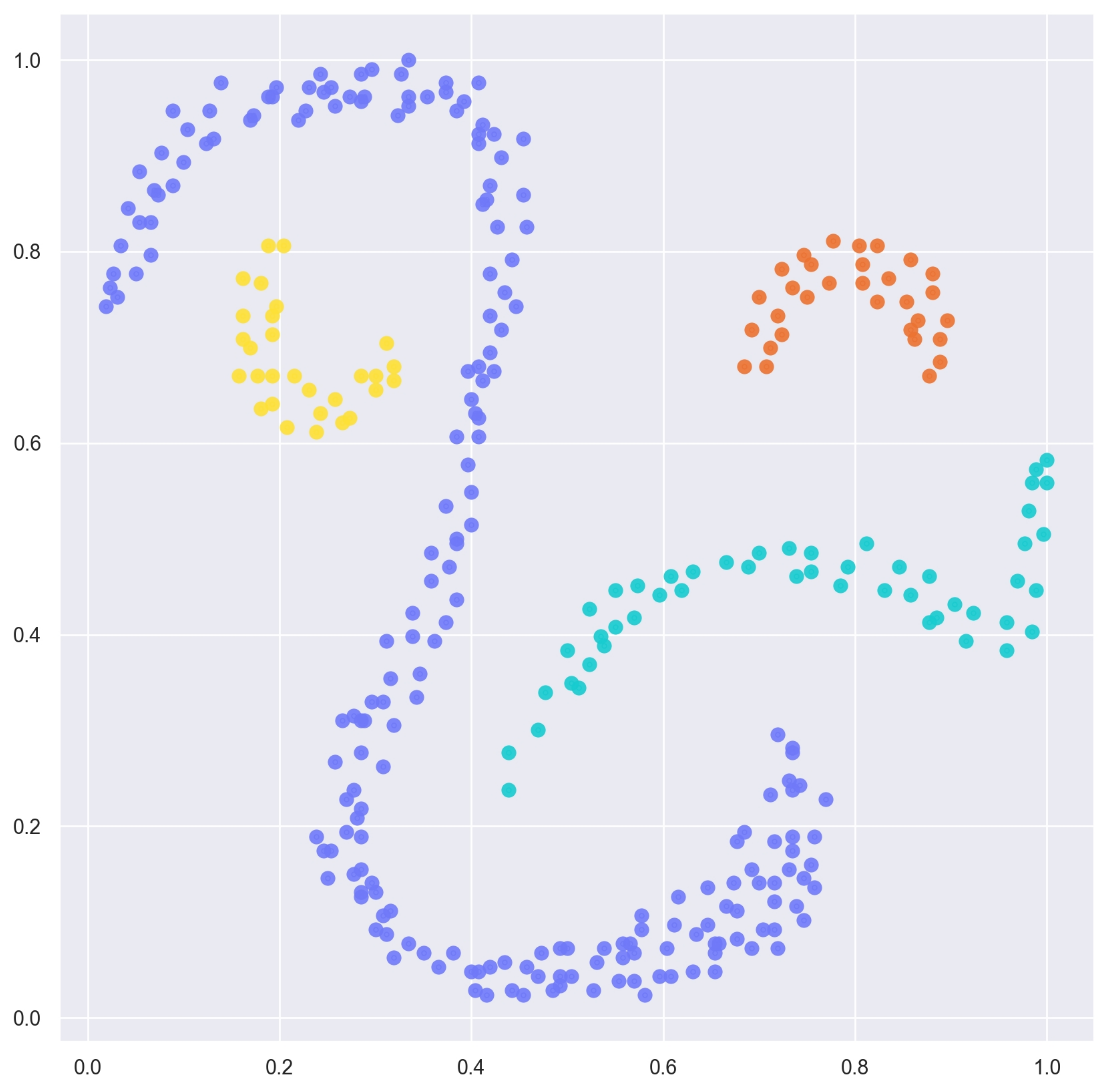}%
		\label{fig_5_3}}
	
	\subfloat[]{\includegraphics[width=1.65in]{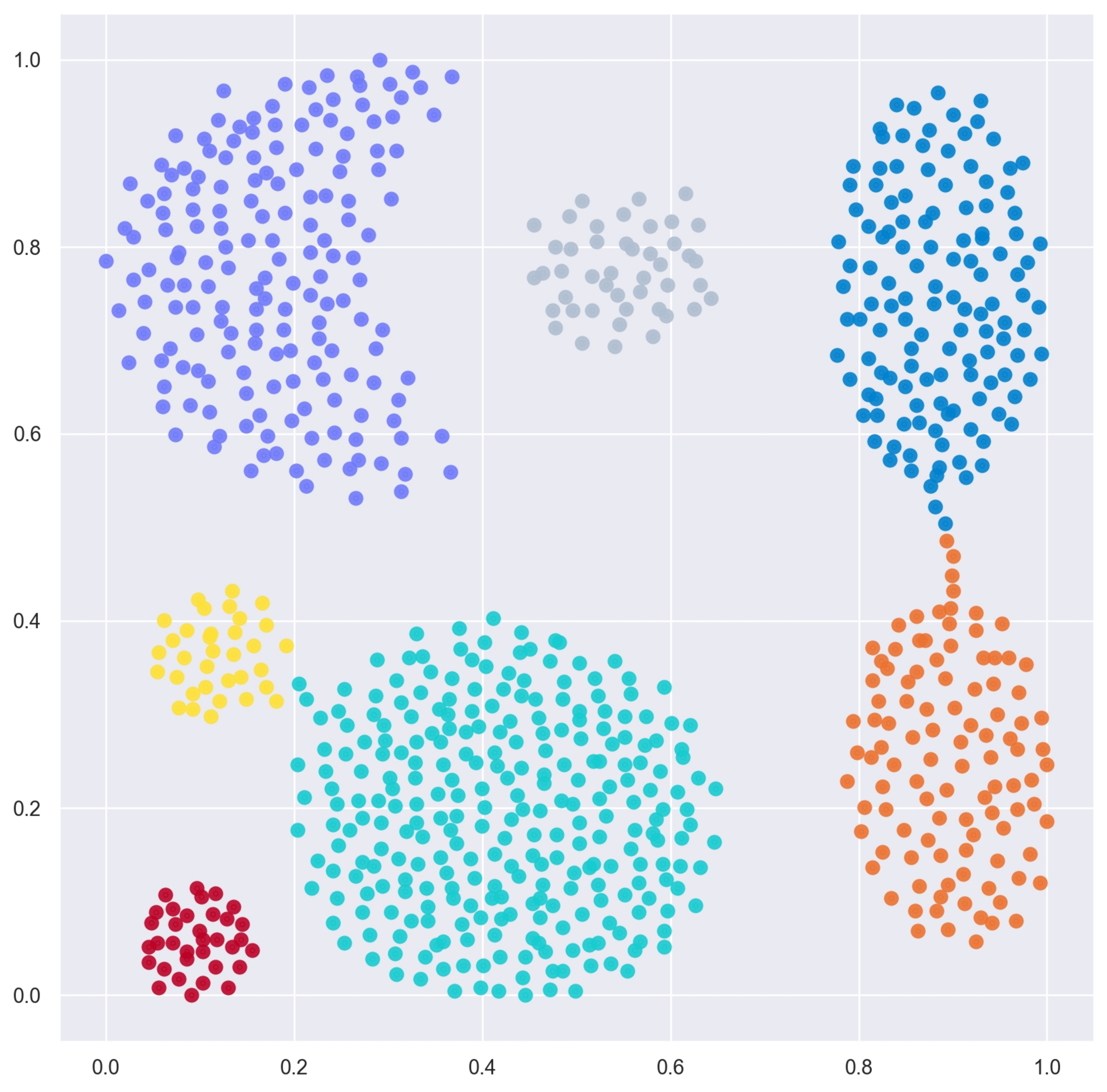}%
		\label{fig_5_4}}
	\subfloat[]{\includegraphics[width=1.65in]{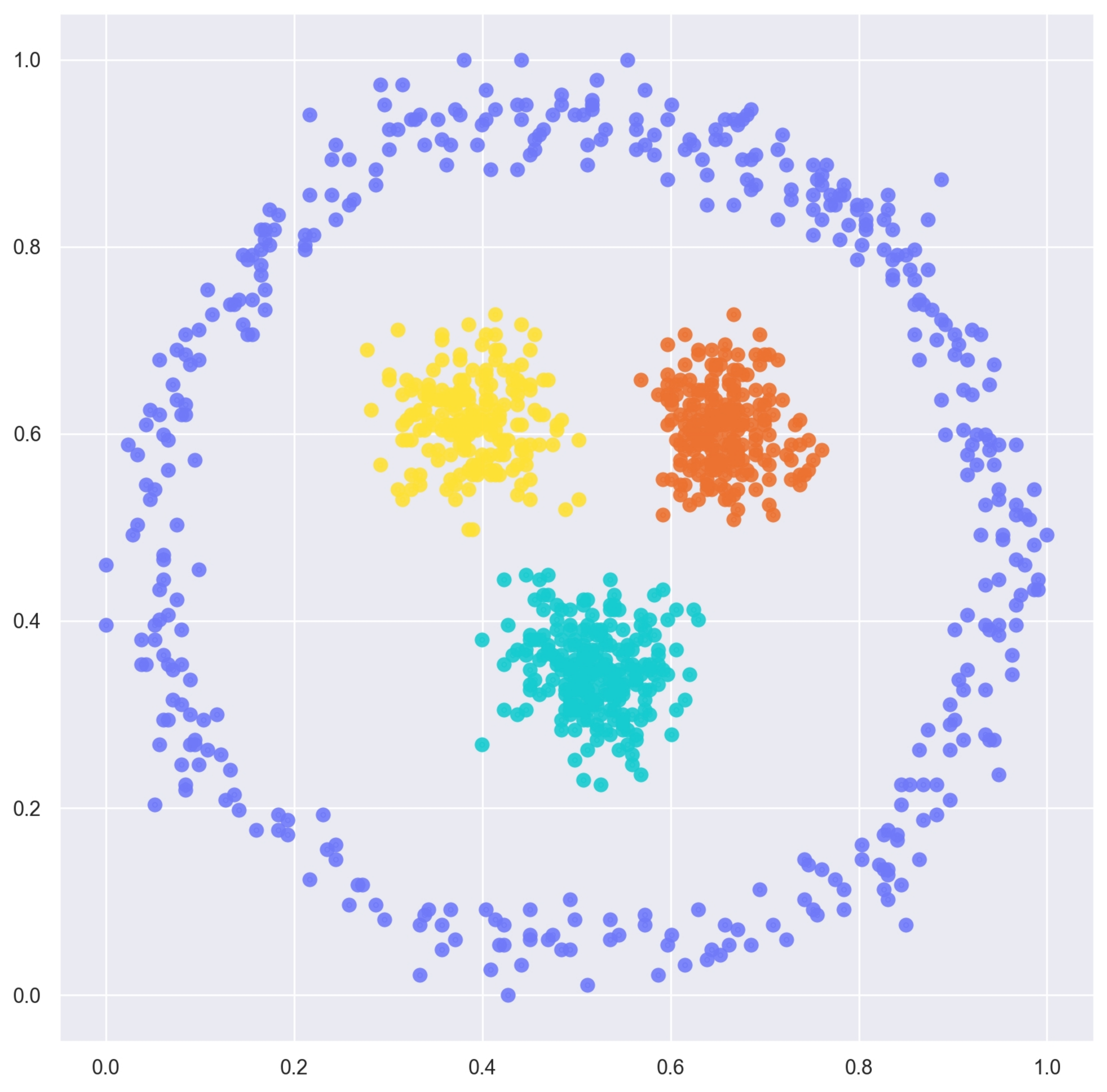}%
		\label{fig_5_5}}
	\subfloat[]{\includegraphics[width=1.65in]{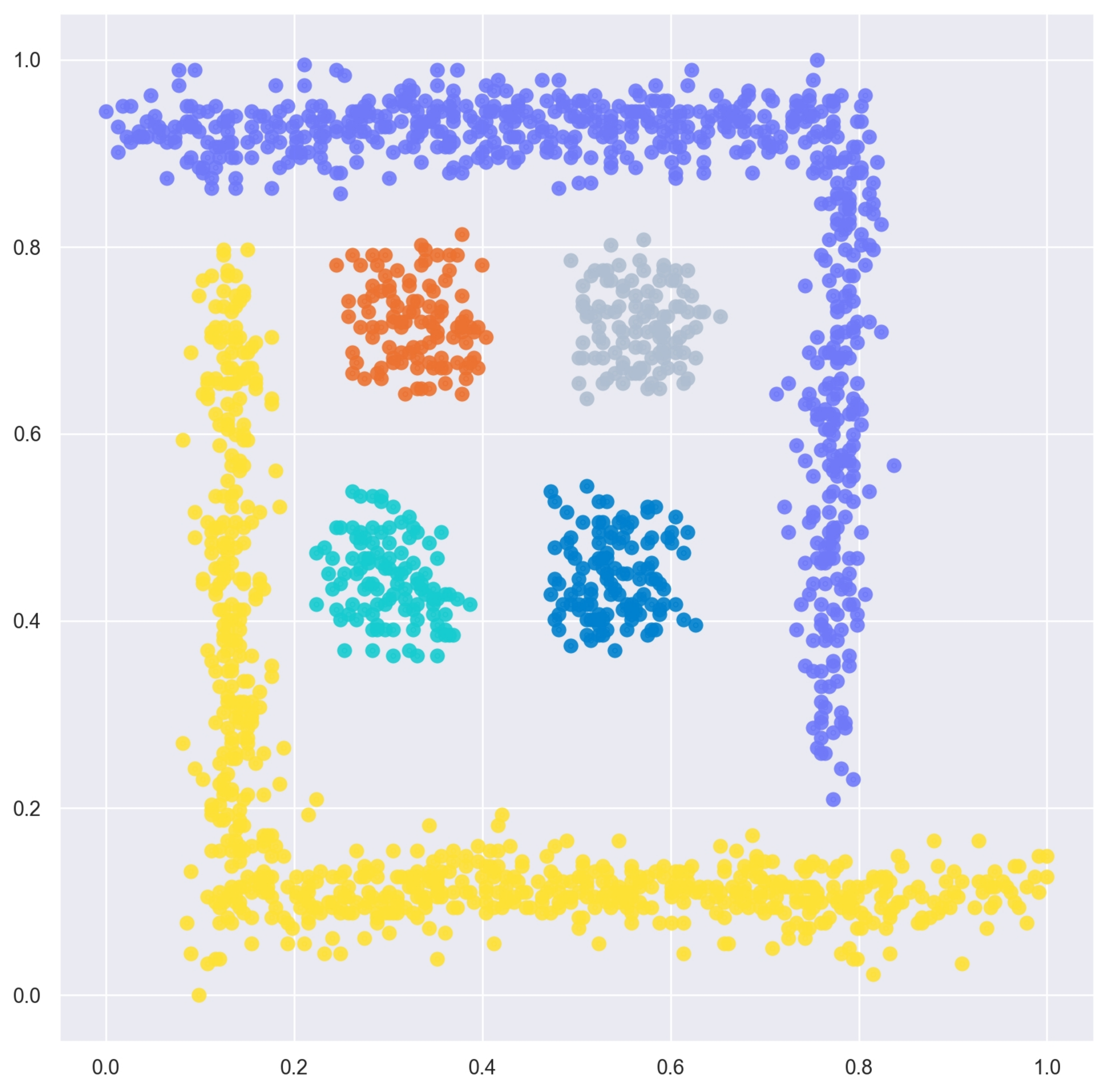}%
		\label{fig_5_6}}
	\caption{The clustering results of LDP\_MST on synthetic data set without noise.}
	\label{fig_5}
   \end{figure*}
	\begin{figure*}[!h]
	\centering
	\subfloat[]{\includegraphics[width=1.65in]{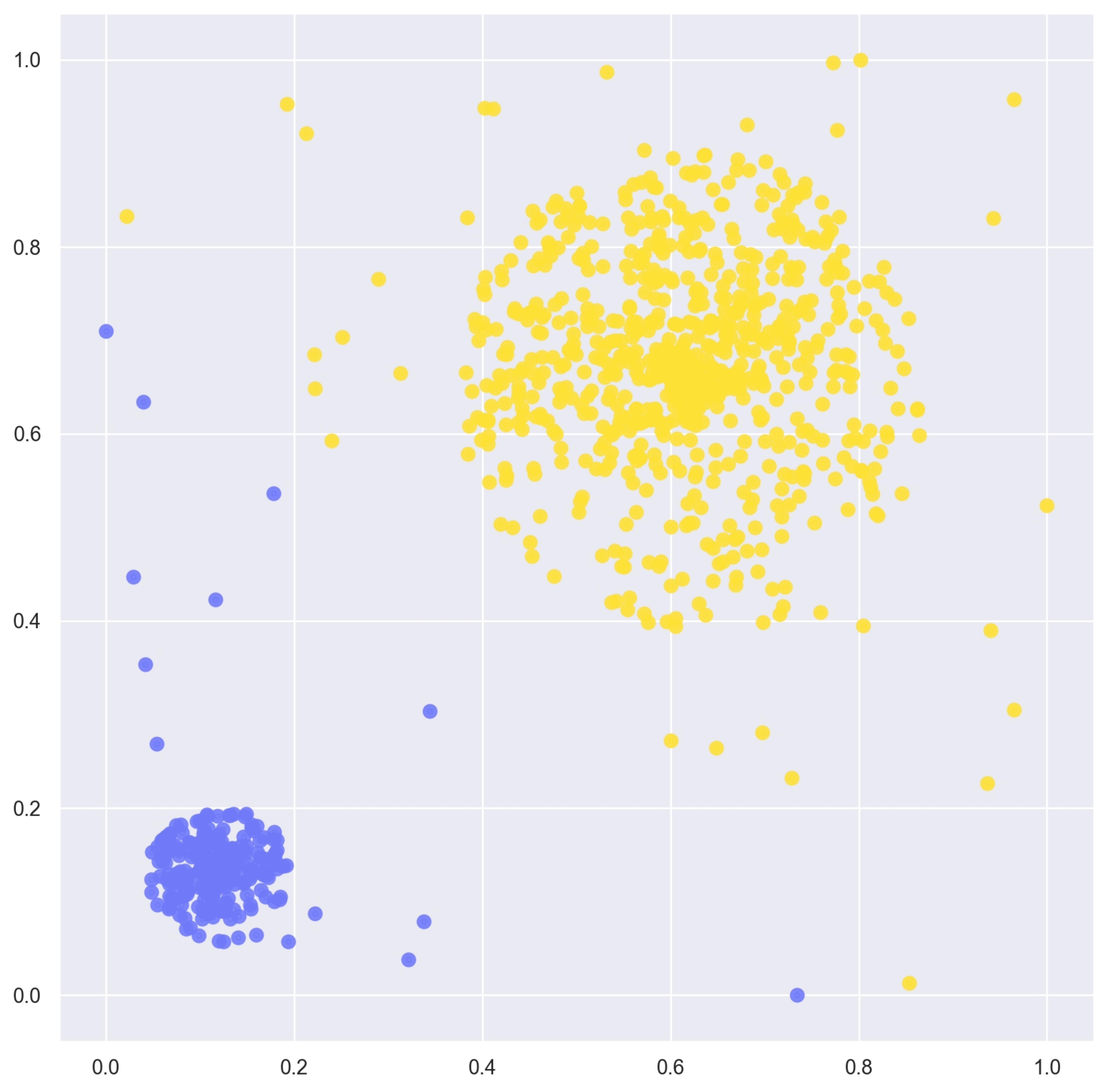}%
		\label{fig_6_1}}
	\subfloat[]{\includegraphics[width=1.65in]{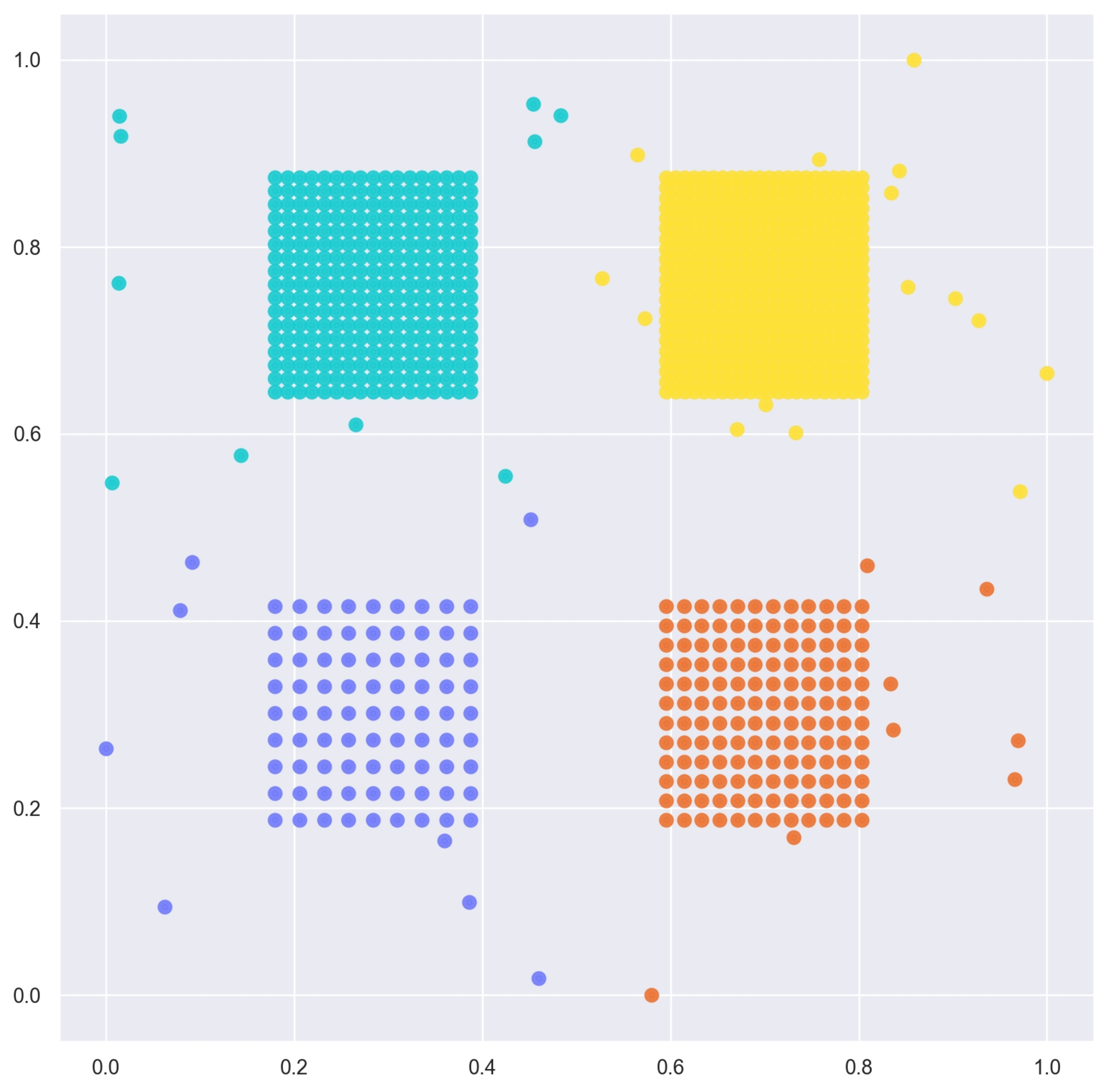}%
		\label{fig_6_2}}
	\subfloat[]{\includegraphics[width=1.65in]{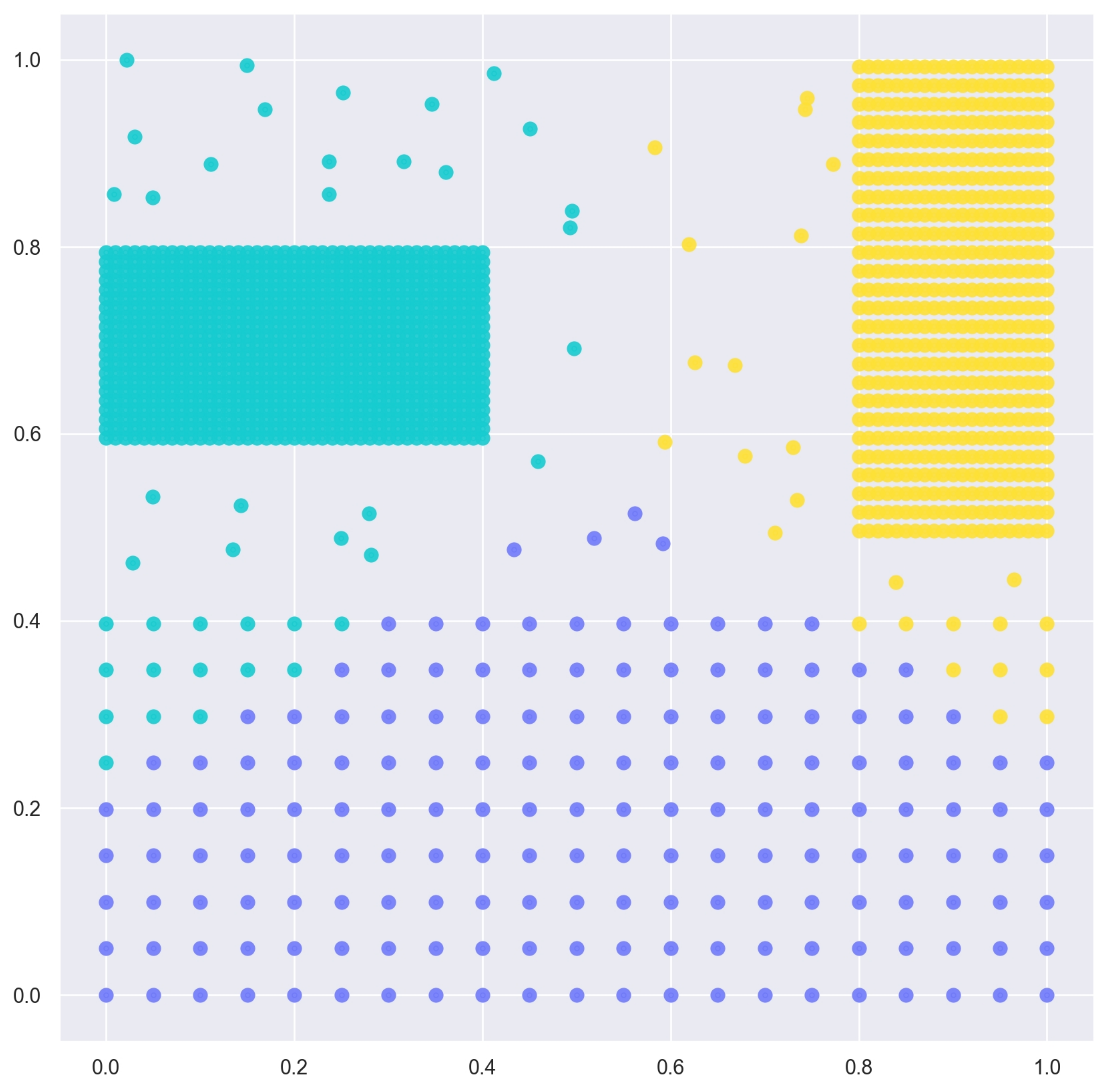}%
		\label{fig_6_3}}
	\subfloat[]{\includegraphics[width=1.65in]{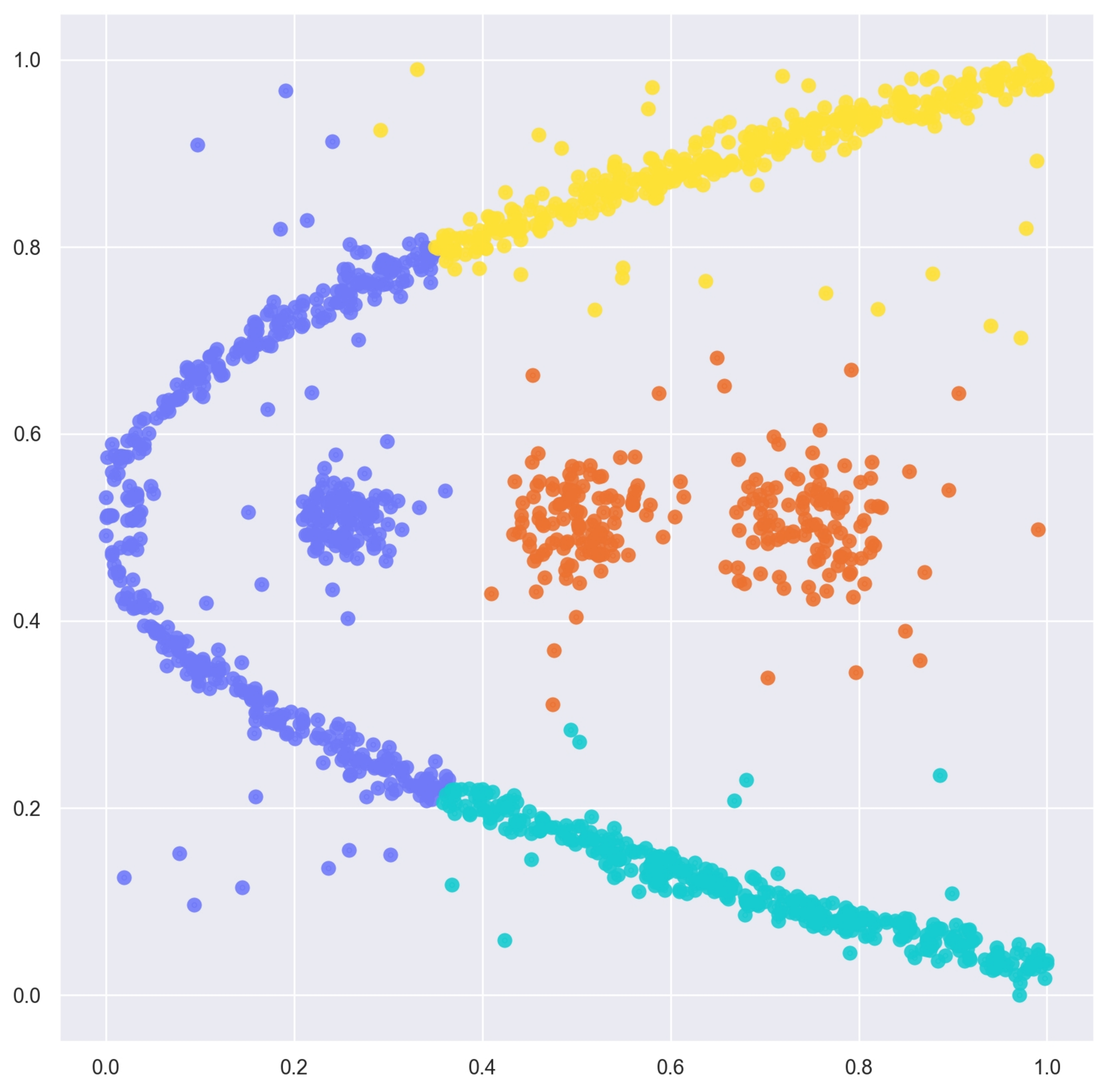}%
		\label{fig_6_4}}
	\caption{The clustering results of K-means on synthetic data set with noise.}
	\label{fig_5}
   \end{figure*}
	\begin{figure*}[!h]
	\centering
	\subfloat[]{\includegraphics[width=1.65in]{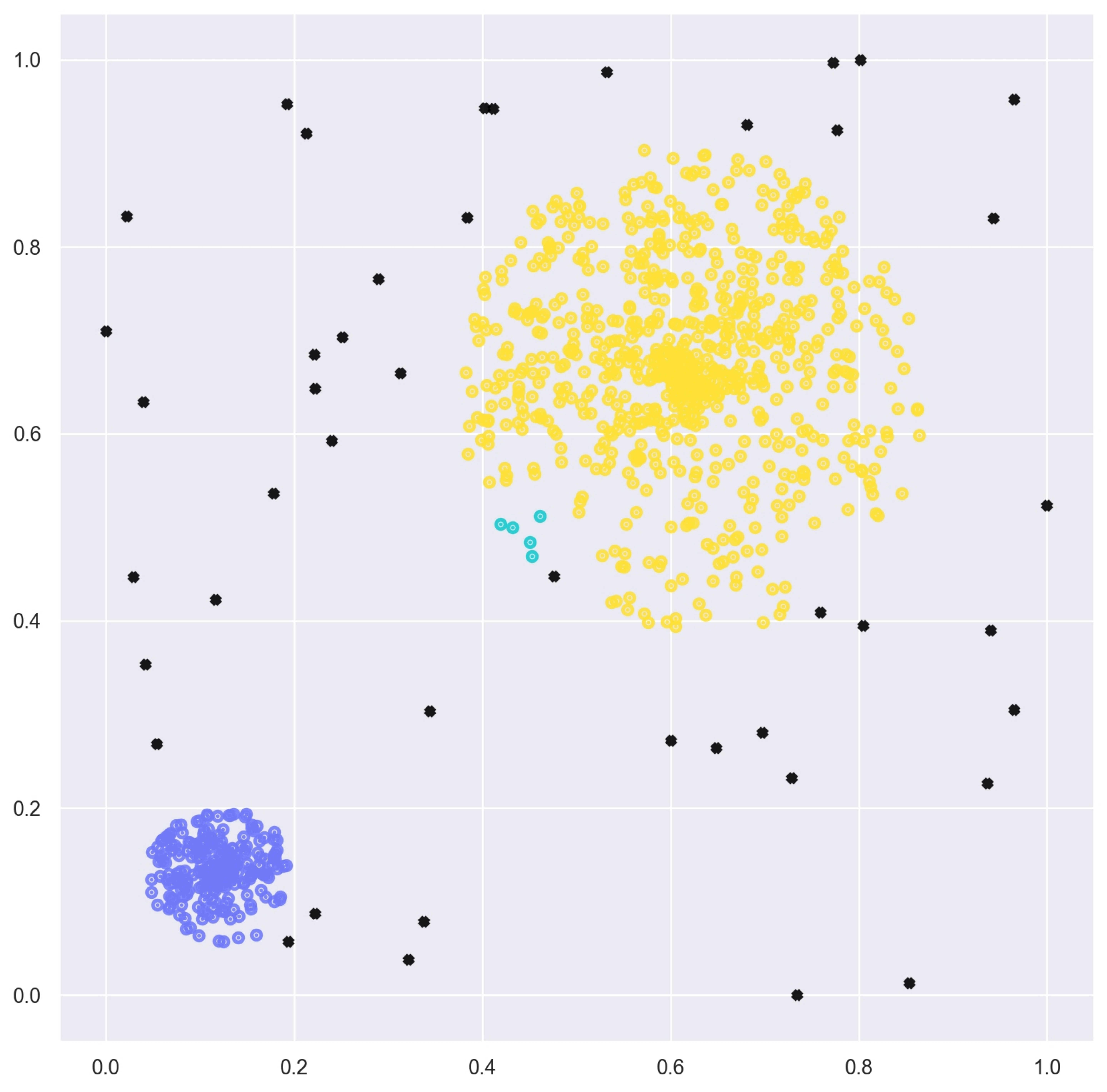}%
		\label{fig_7_1}}
	\subfloat[]{\includegraphics[width=1.65in]{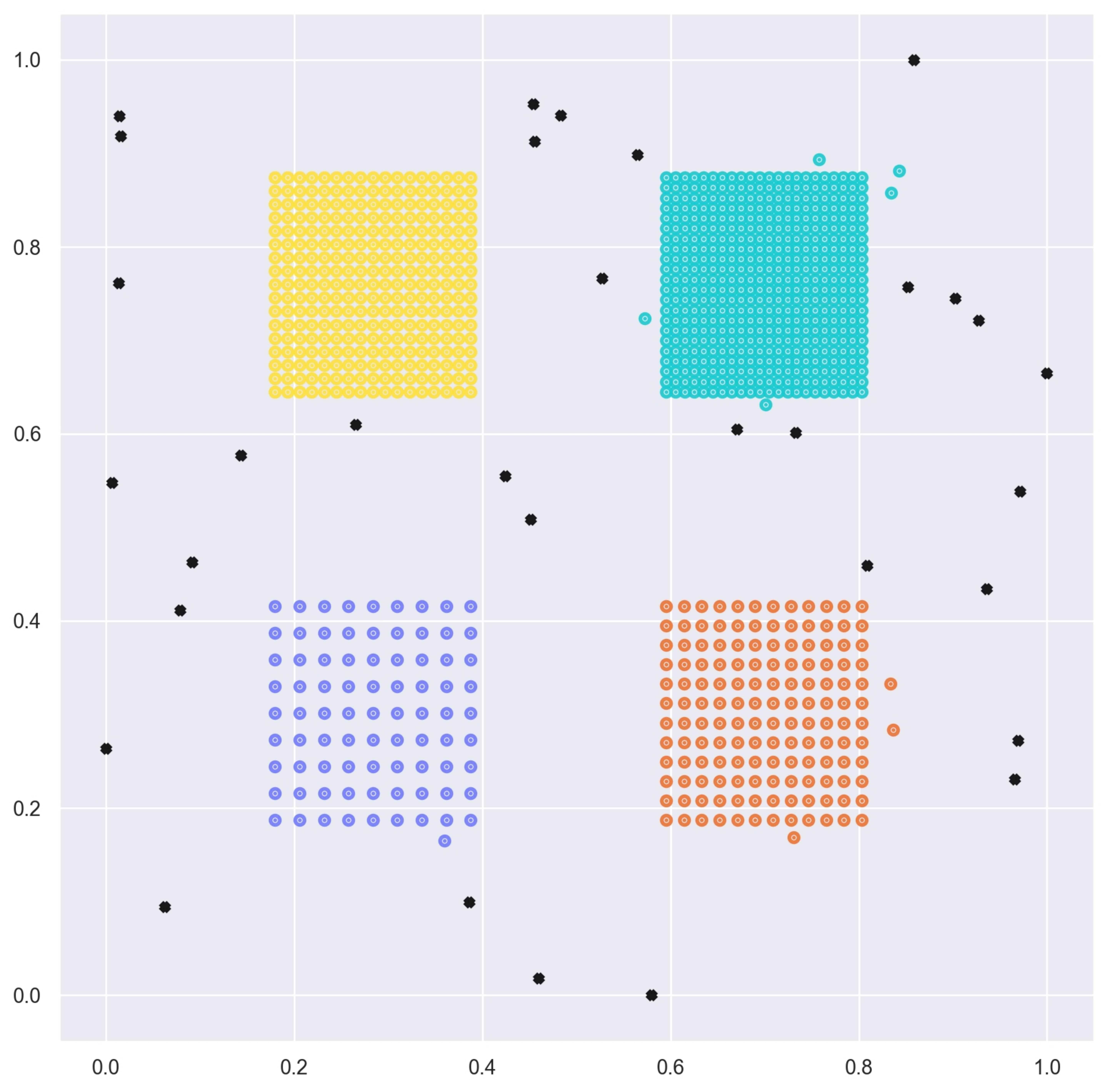}%
		\label{fig_7_2}}
	\subfloat[]{\includegraphics[width=1.65in]{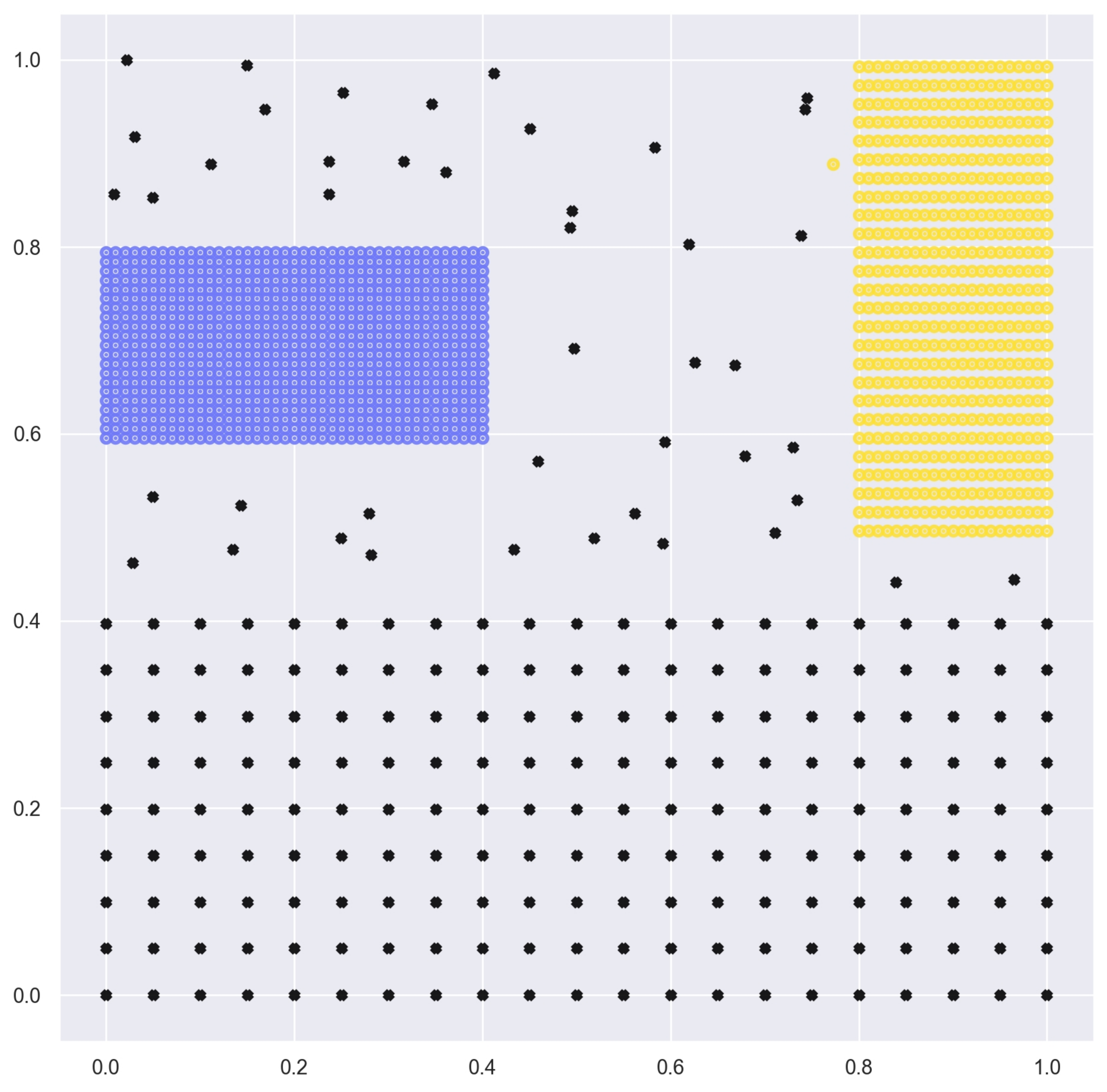}%
		\label{fig_7_3}}
	\subfloat[]{\includegraphics[width=1.65in]{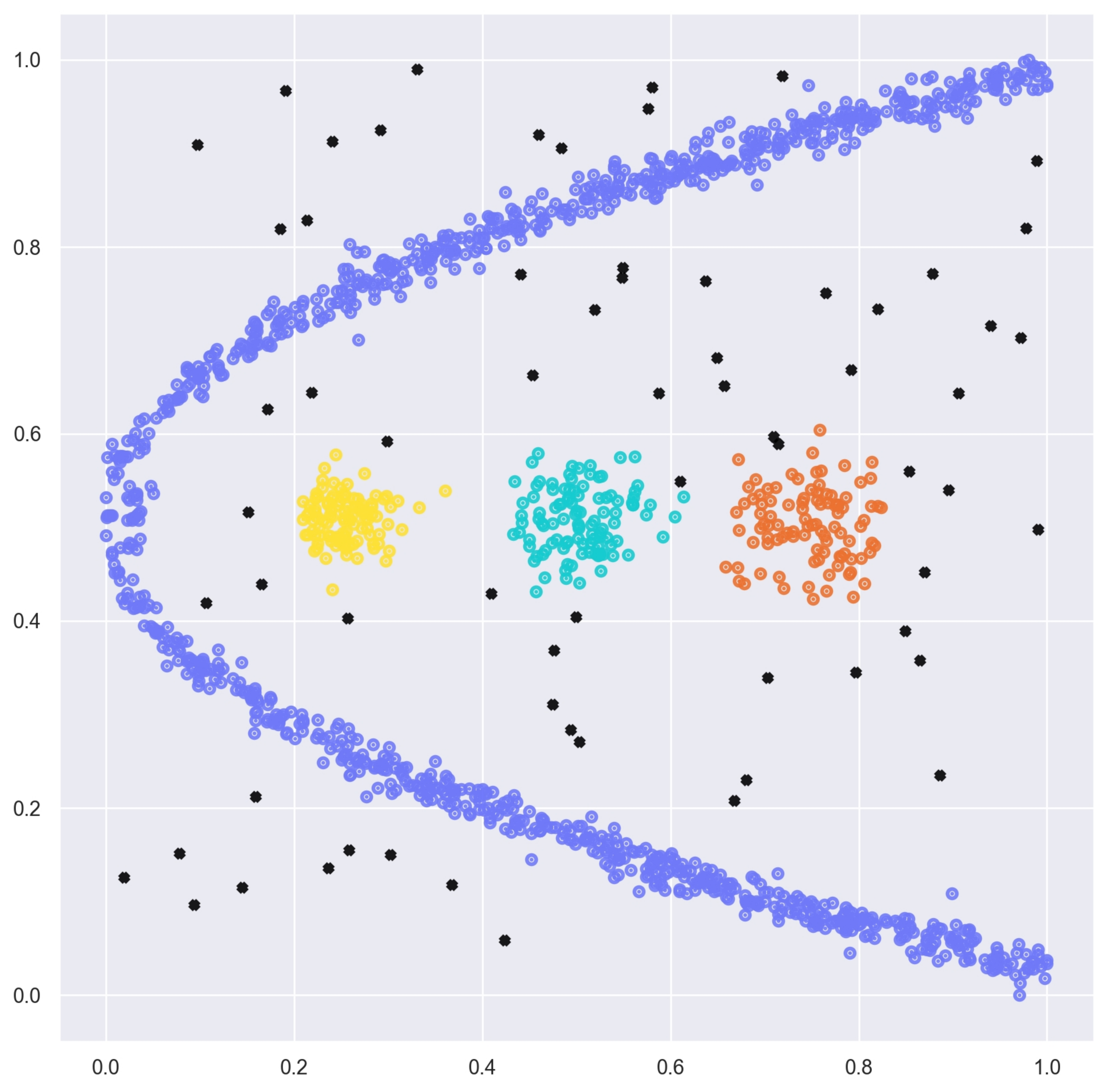}%
		\label{fig_7_4}}
	\caption{The clustering results of DBSCAN on synthetic data set with noise.}
	\label{fig_8}
\end{figure*}
	\begin{figure*}[!h]
	\centering
	\subfloat[]{\includegraphics[width=1.65in]{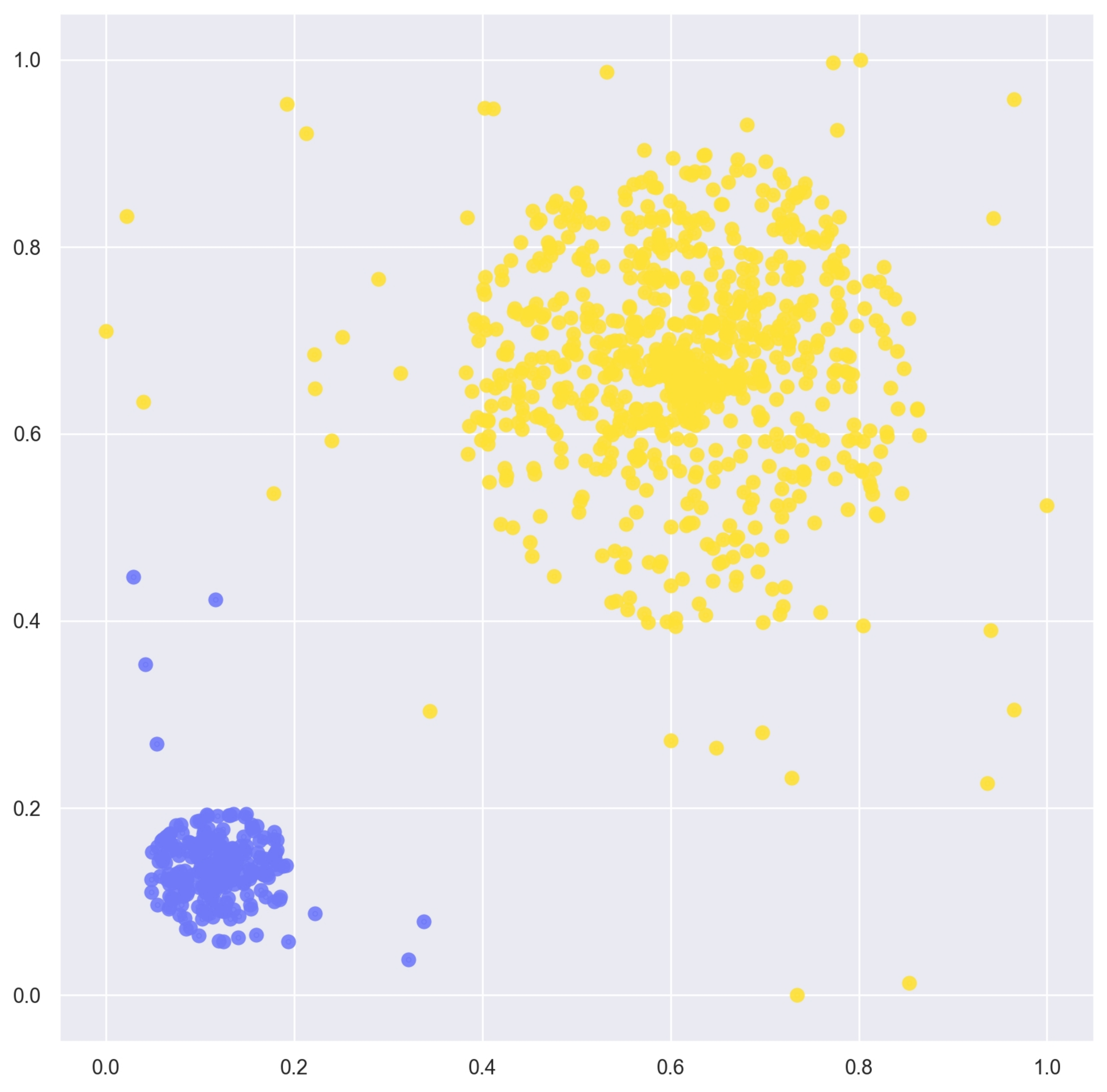}%
		\label{fig_8_1}}
	\subfloat[]{\includegraphics[width=1.65in]{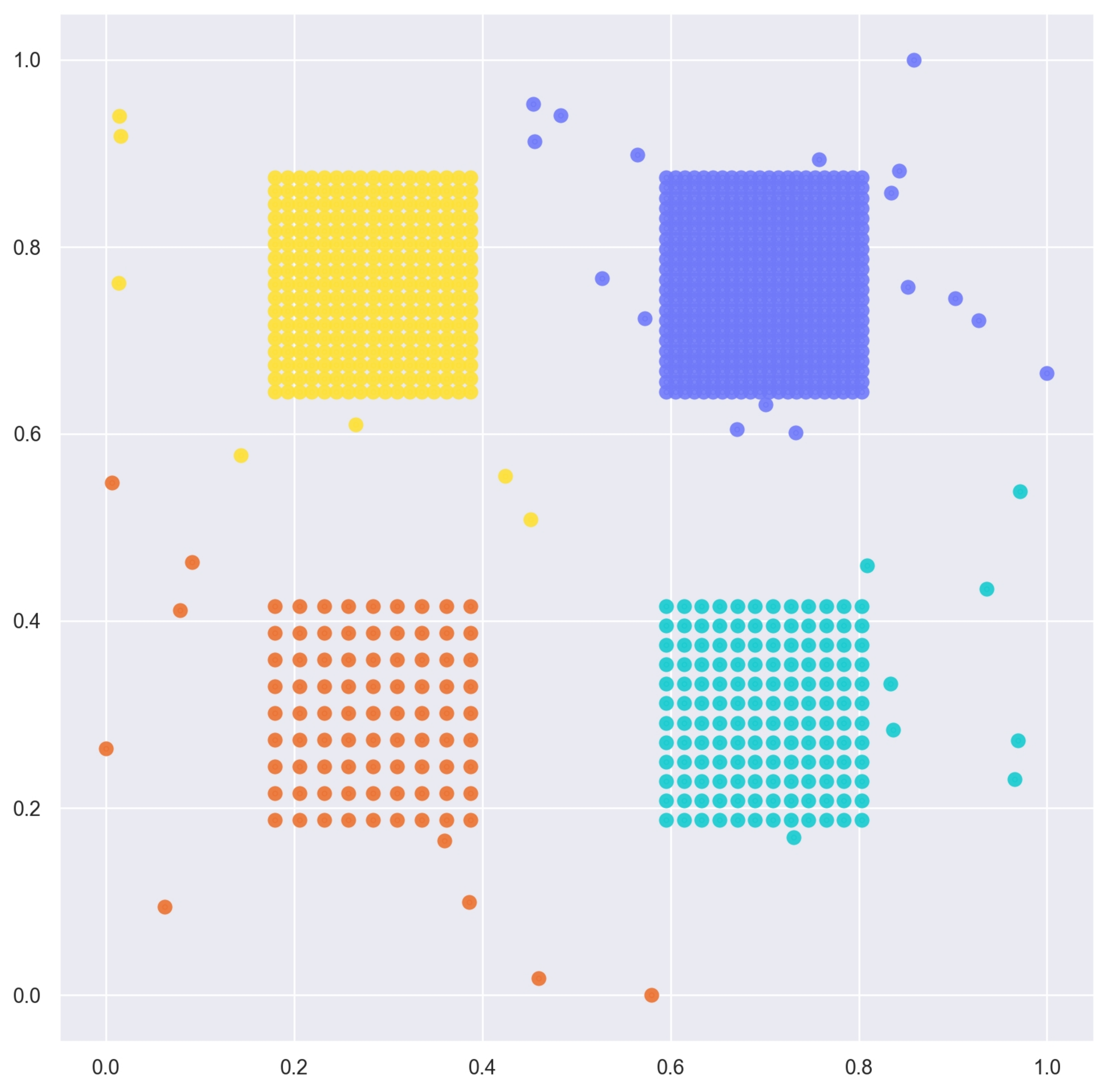}%
		\label{fig_8_2}}
	\subfloat[]{\includegraphics[width=1.65in]{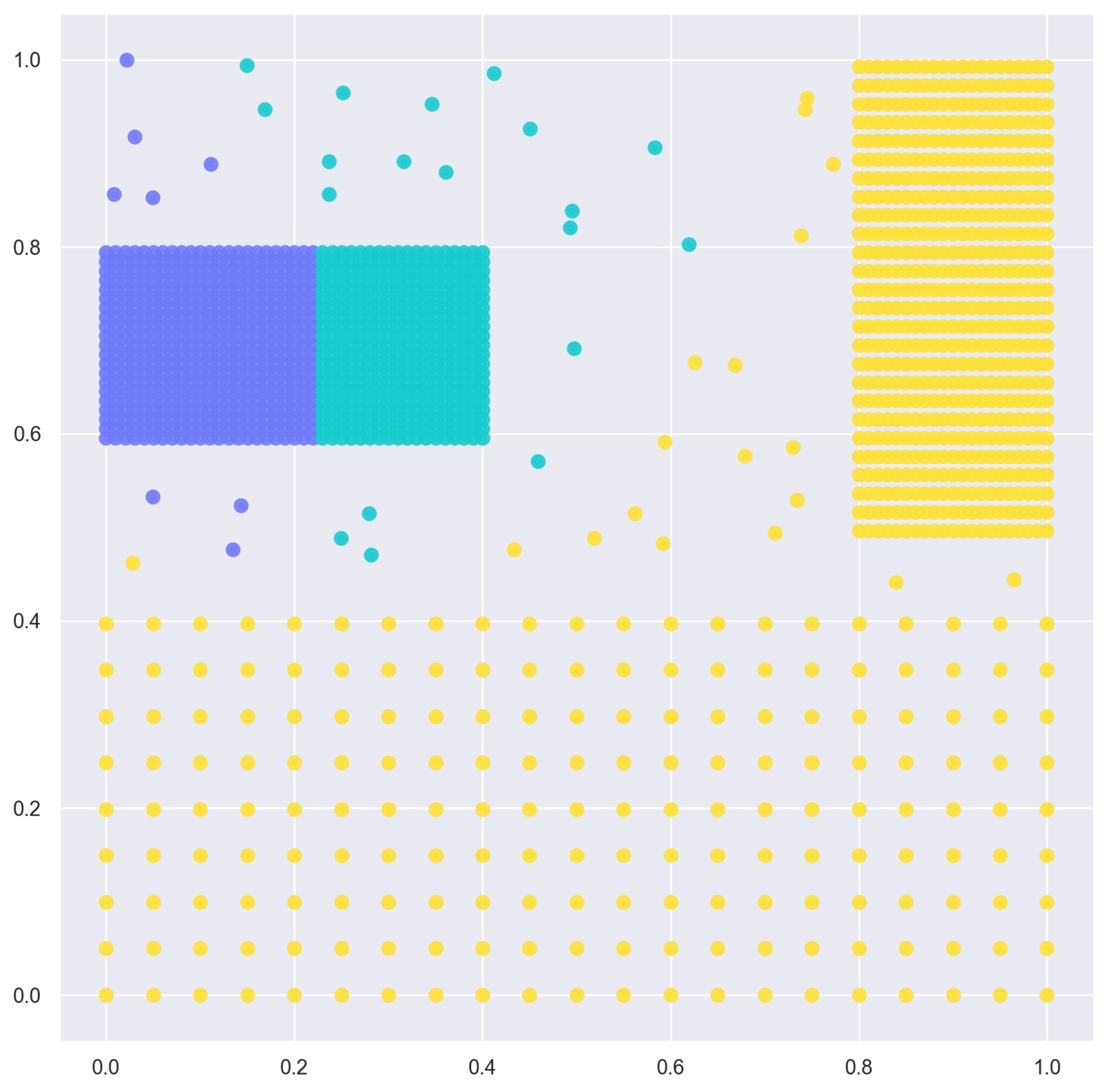}%
		\label{fig_8_3}}
	\subfloat[]{\includegraphics[width=1.65in]{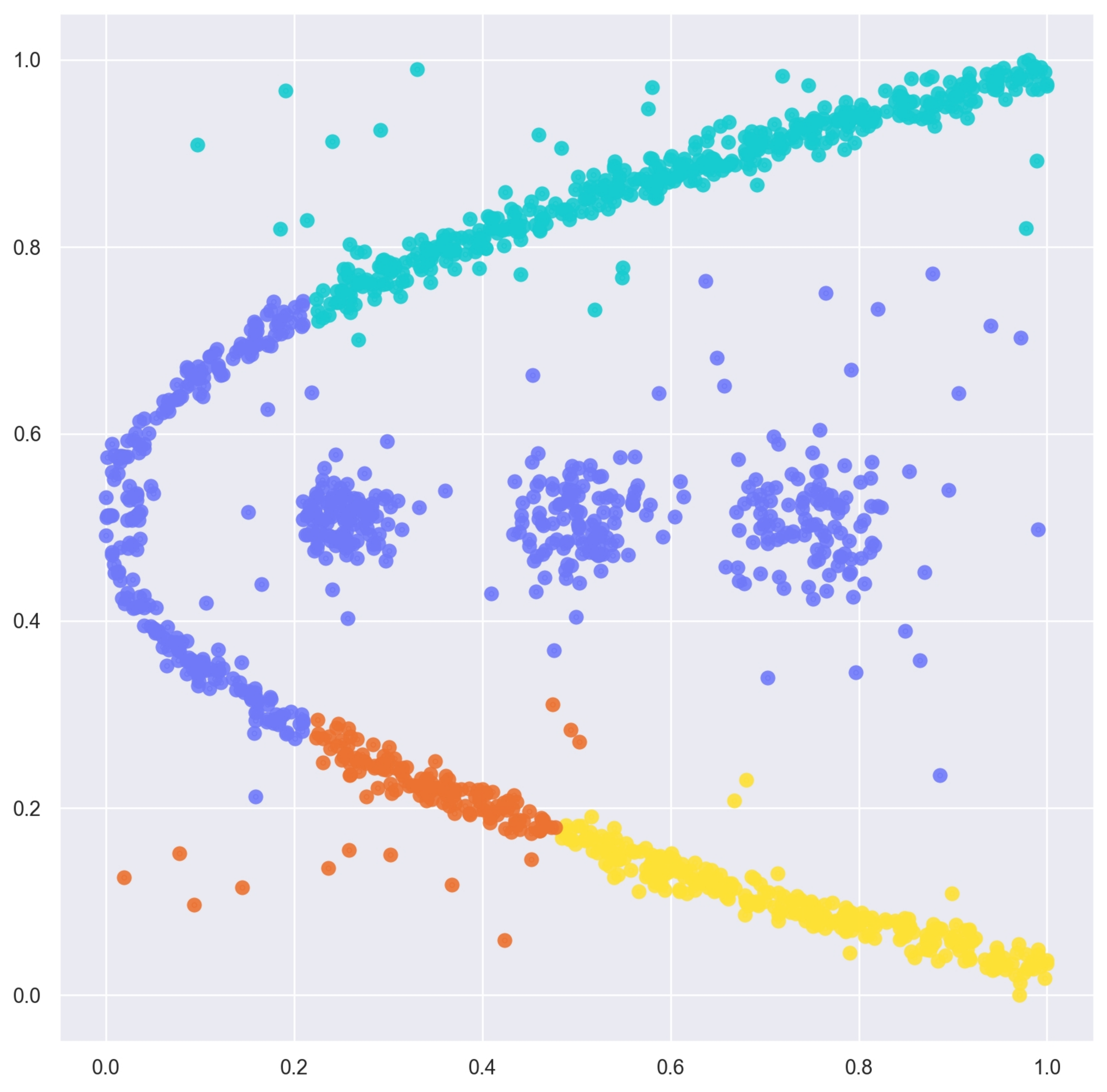}%
		\label{fig_8_4}}
	\caption{The clustering results of DP on synthetic data set with noise.}
	\label{fig_8}
\end{figure*}
	\begin{figure*}[!h]
	\centering
	\subfloat[]{\includegraphics[width=1.65in]{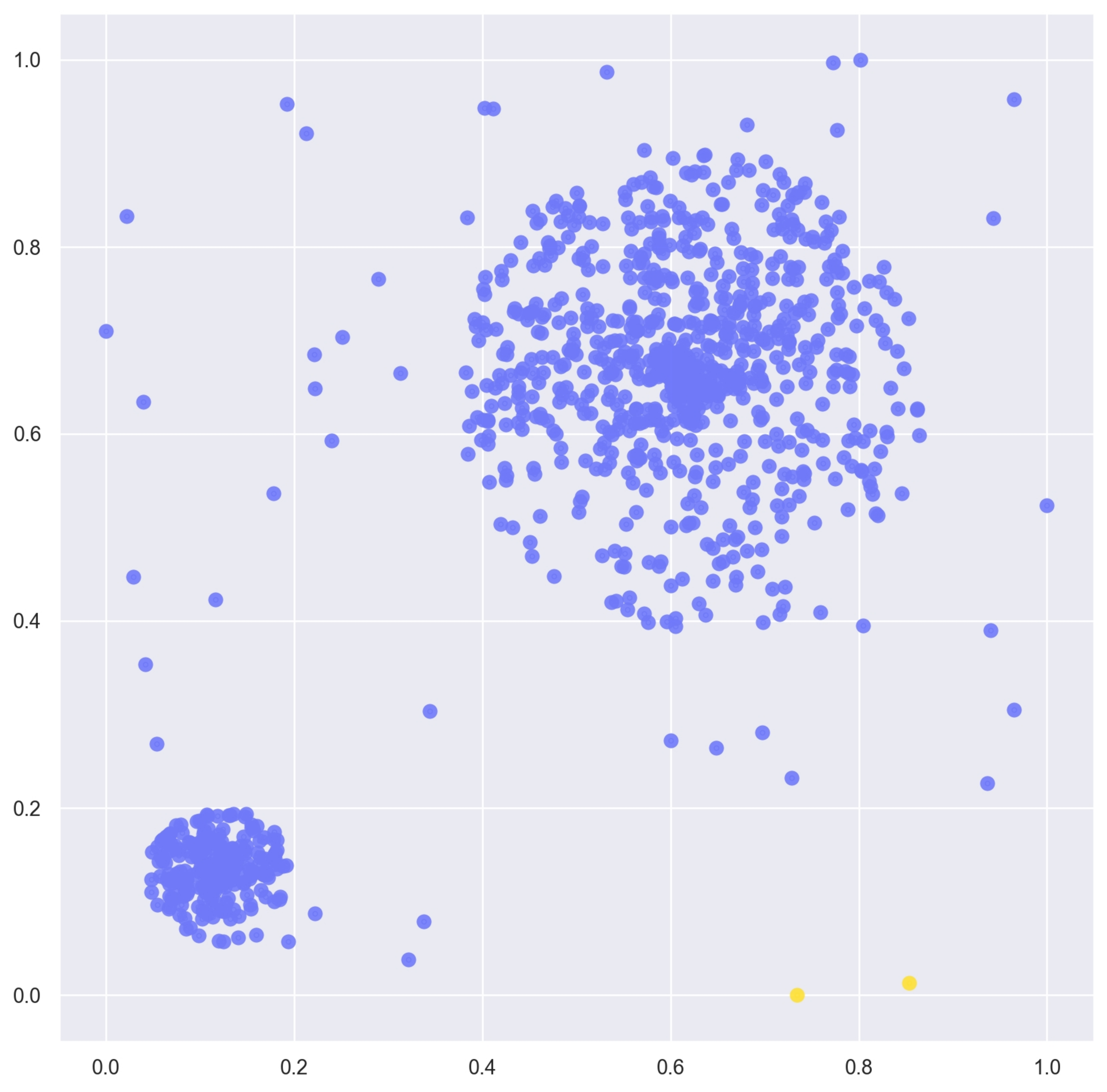}%
		\label{fig_9_1}}
	\subfloat[]{\includegraphics[width=1.65in]{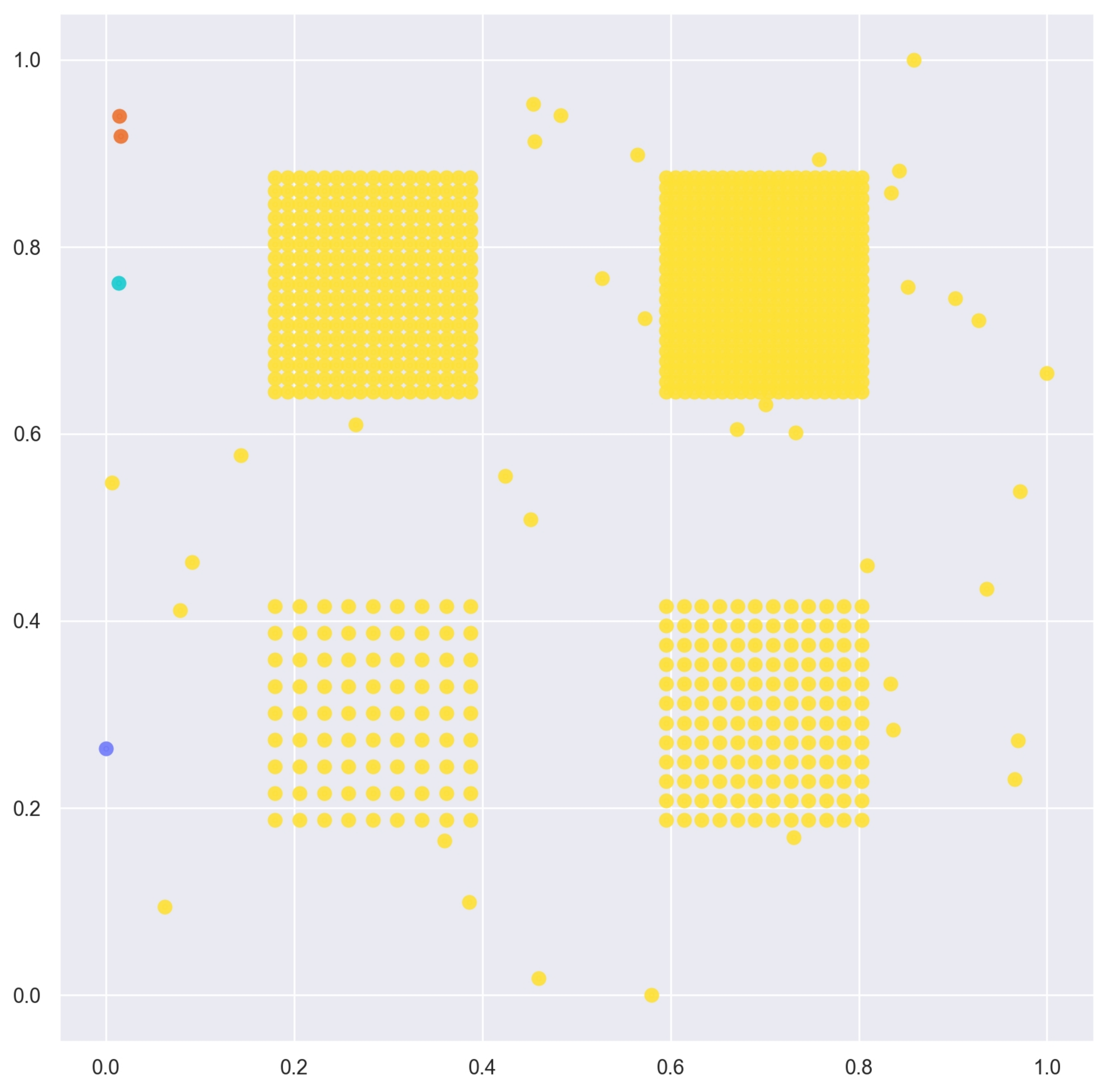}%
		\label{fig_9_2}}
	\subfloat[]{\includegraphics[width=1.65in]{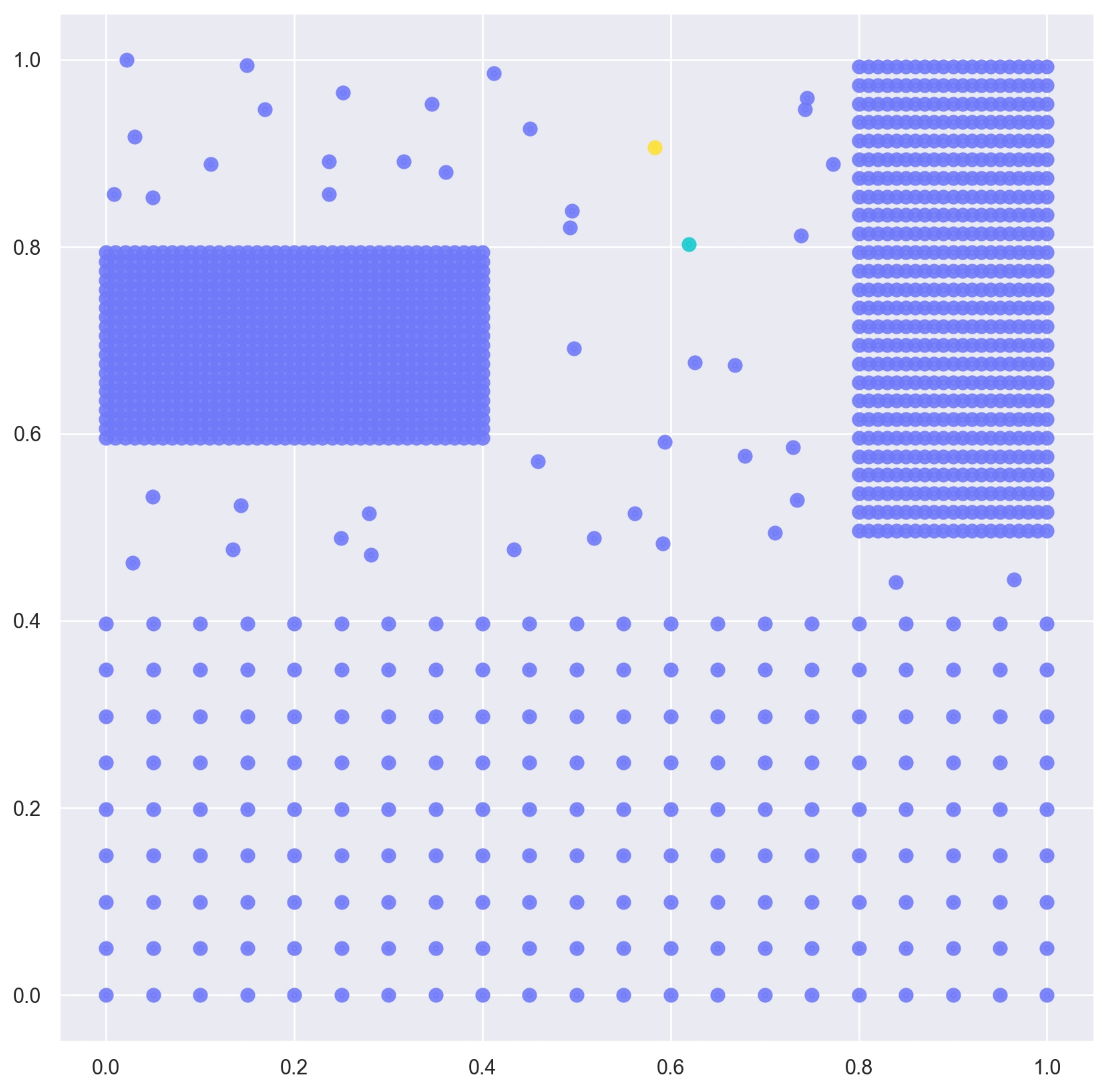}%
		\label{fig_9_3}}
	\subfloat[]{\includegraphics[width=1.65in]{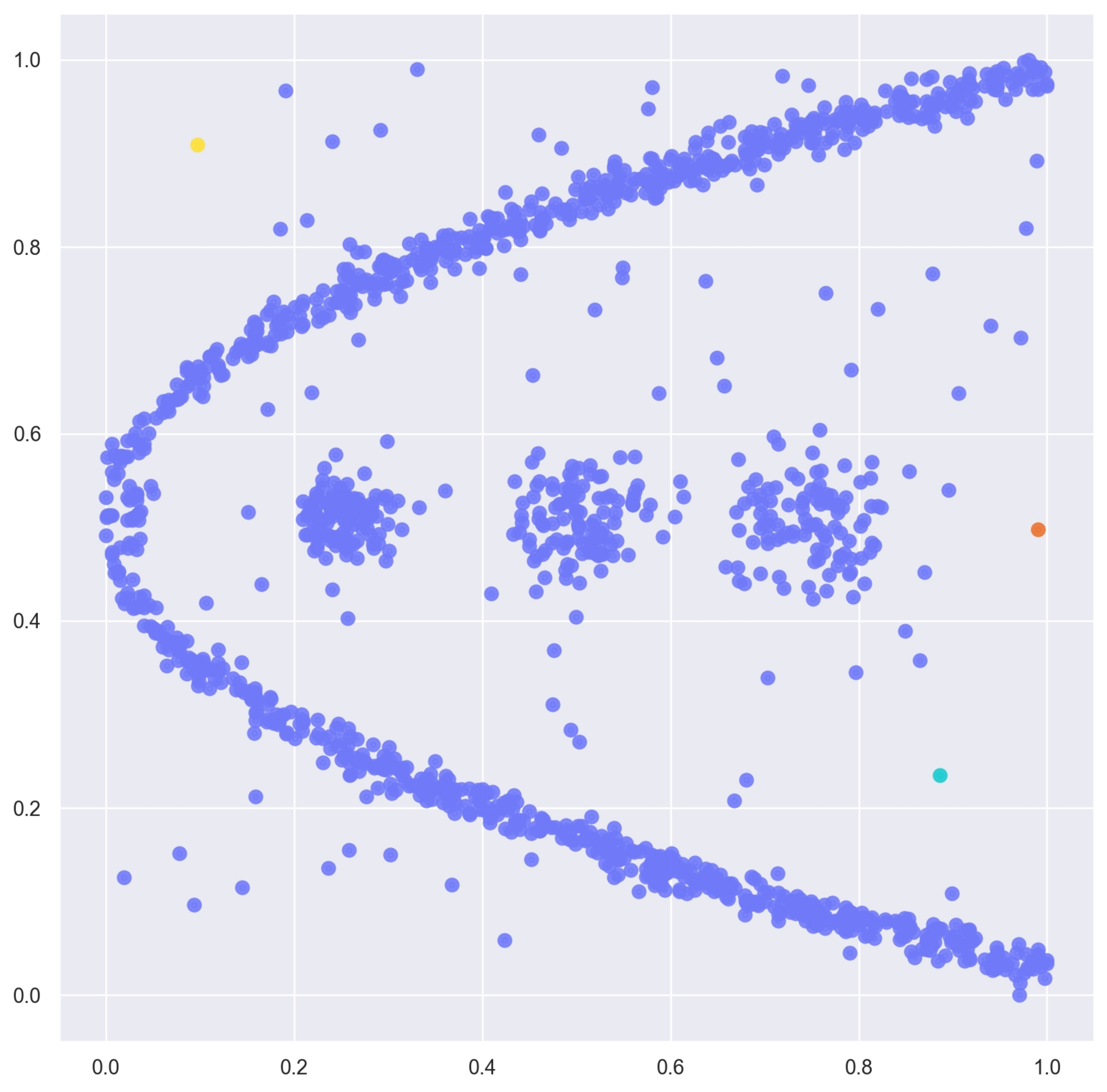}%
		\label{fig_9_4}}

	\caption{The clustering results of Normal\_MST on synthetic data set with noise.}
	\label{fig_9}
\end{figure*}
	\begin{figure*}[!h]
	\centering
	\subfloat[]{\includegraphics[width=1.65in]{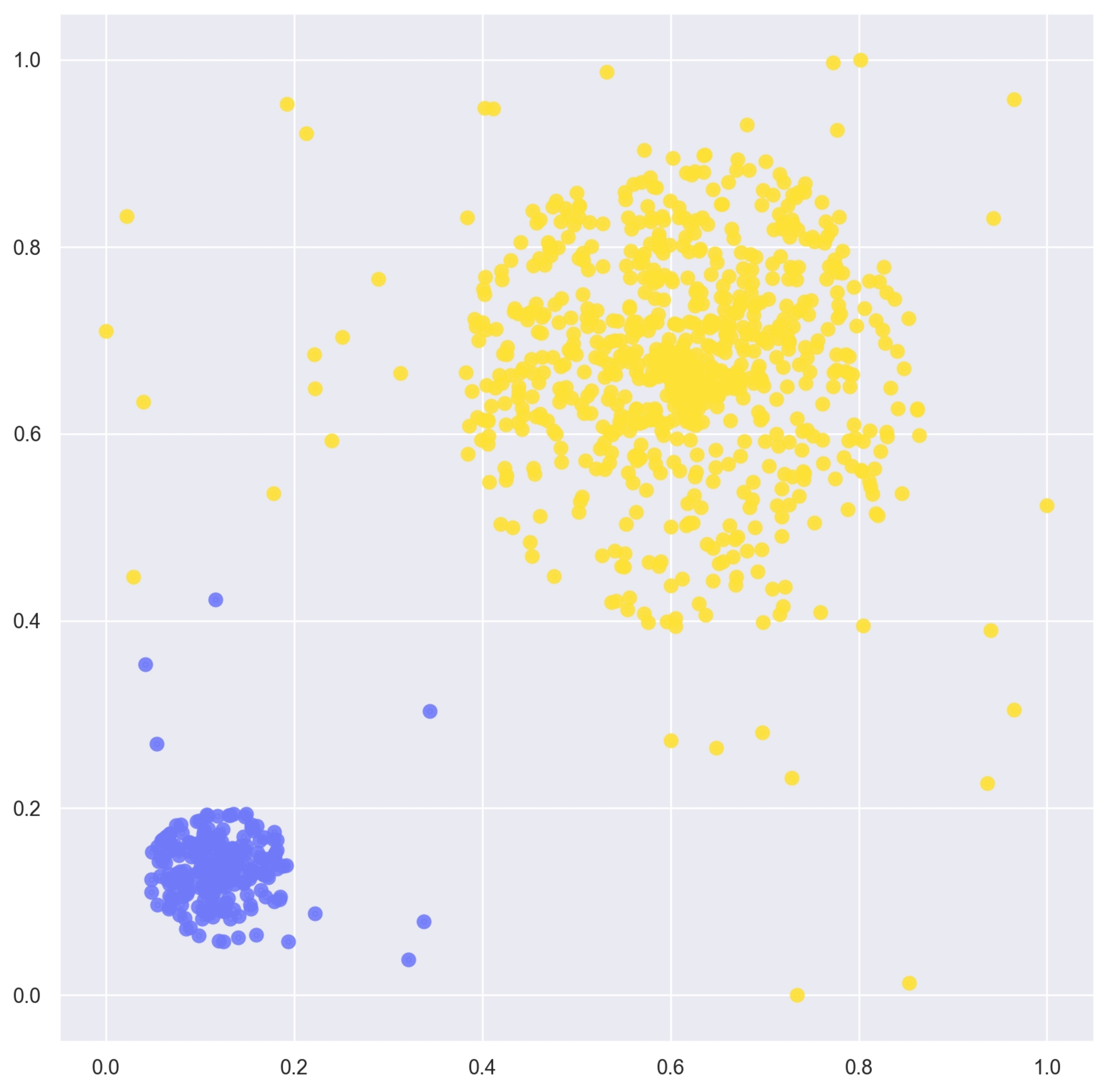}%
		\label{fig_10_1}}
	\subfloat[]{\includegraphics[width=1.65in]{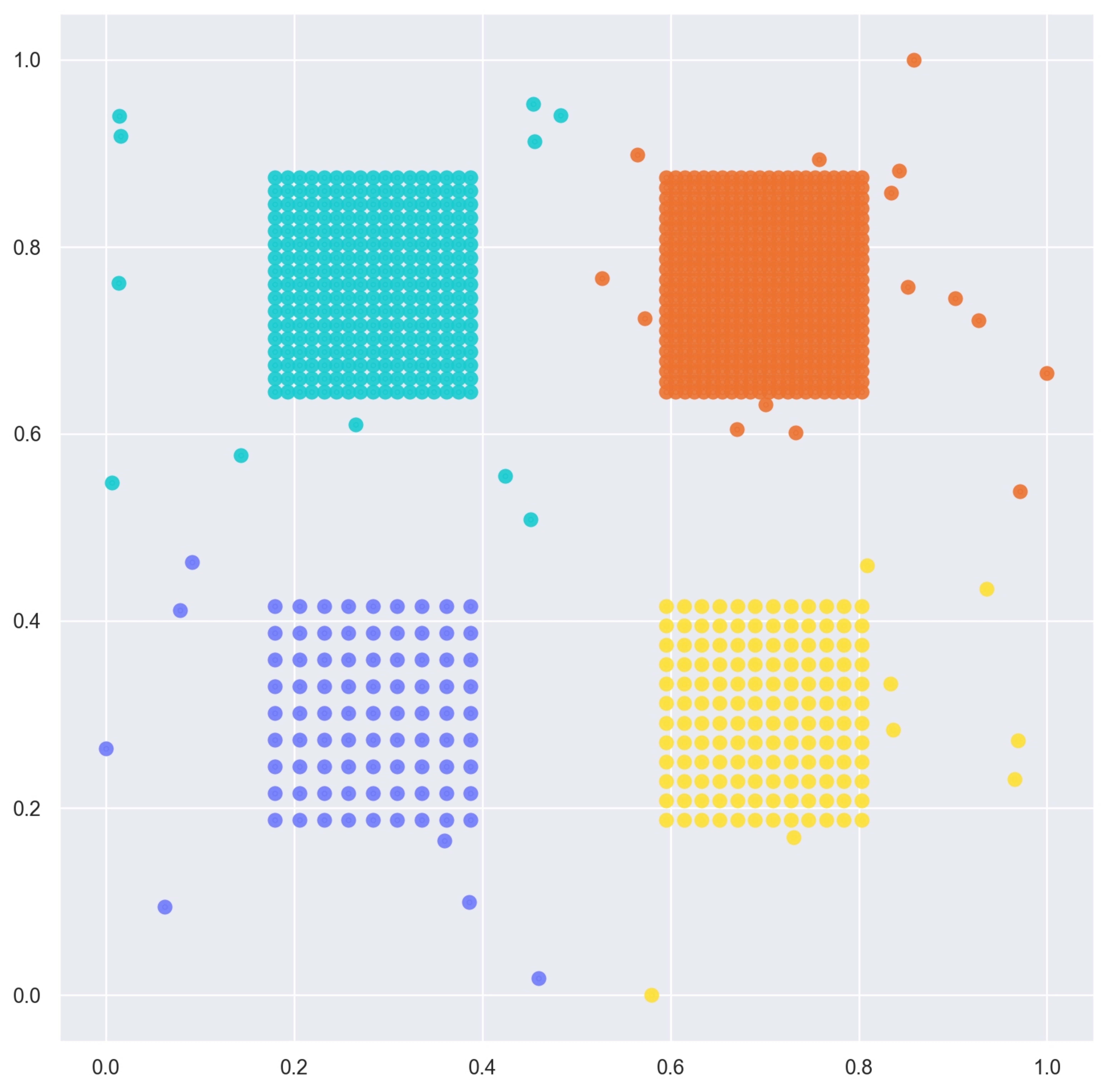}%
		\label{fig_10_2}}
	\subfloat[]{\includegraphics[width=1.65in]{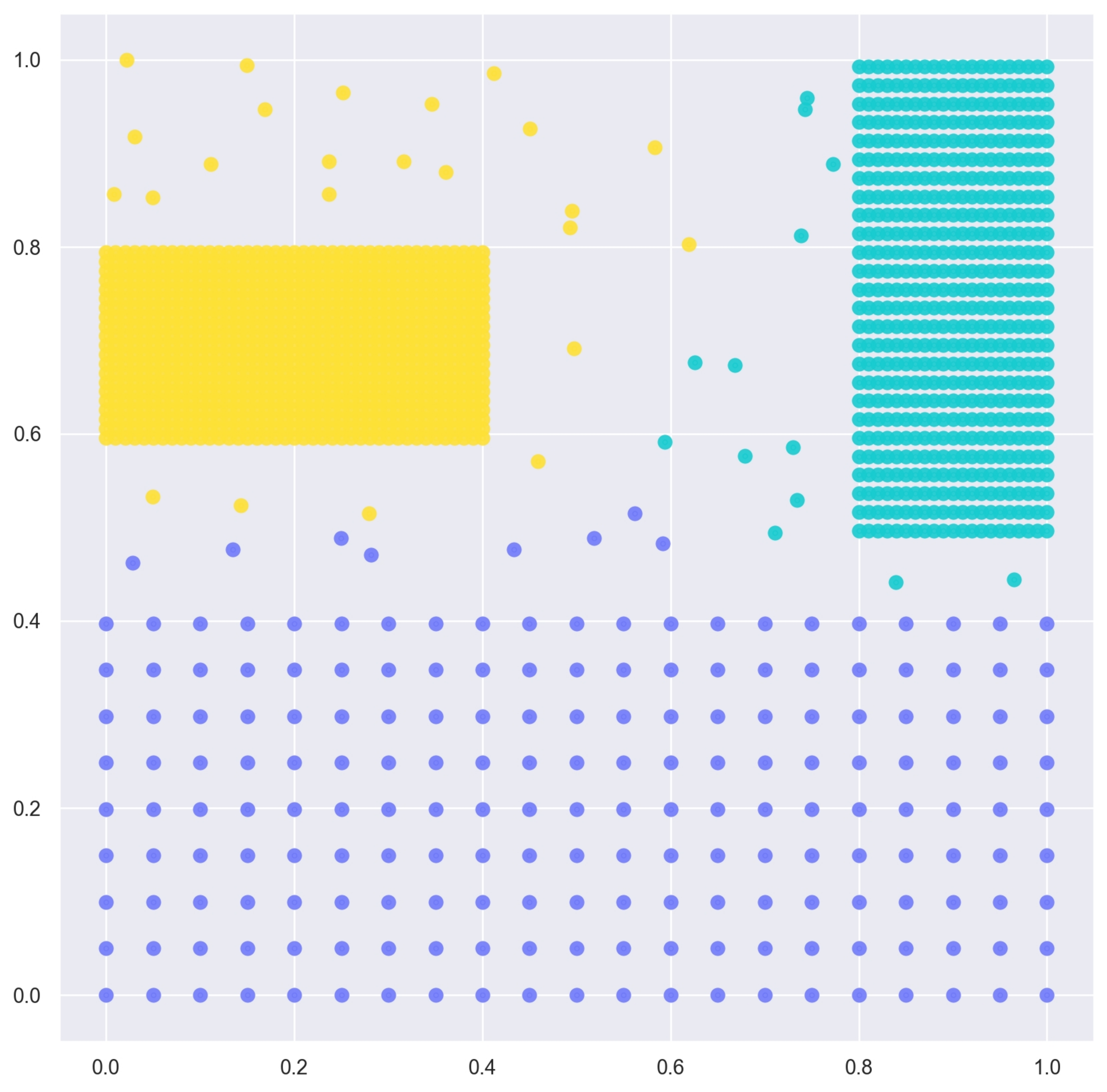}%
		\label{fig_10_3}}
	\subfloat[]{\includegraphics[width=1.65in]{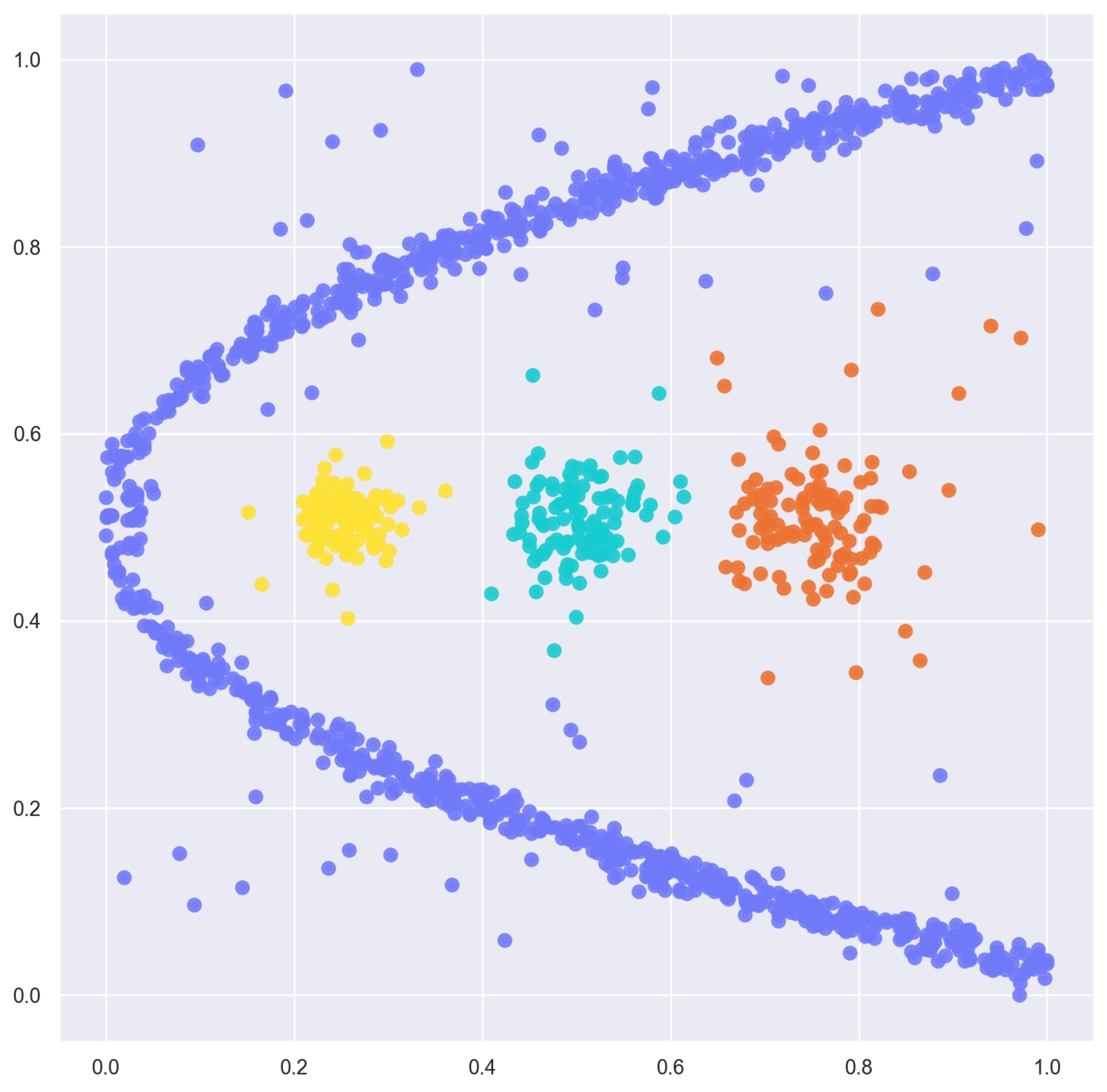}%
		\label{fig_10_4}}

	\caption{The clustering results of LDP\_MST on synthetic data set with noise.}
	\label{fig_10}
\end{figure*}